\documentclass{article}

\usepackage[main, final]{neurips_2026}
\usepackage[utf8]{inputenc} % allow utf-8 input
\usepackage[T1]{fontenc}    % use 8-bit T1 fonts
\usepackage{hyperref}       % hyperlinks
\usepackage{url}            % simple URL typesetting
\usepackage{booktabs}       % professional-quality tables
\usepackage{amsfonts}       % blackboard math symbols
\usepackage{nicefrac}       % compact symbols for 1/2, etc.
\usepackage{microtype}      % microtypography
\usepackage{xcolor}         % colors
\usepackage{amsmath}
\DeclareRobustCommand{\argmax}{\mathop{\mathrm{arg\,max}}\limits}
\usepackage{booktabs,multirow,colortbl,array}
\newcolumntype{C}{>{\centering\arraybackslash}p{1cm}}
\usepackage{graphicx}
\usepackage{amssymb}
\usepackage{tcolorbox}
\tcbuselibrary{skins}
\usepackage[table]{xcolor}
\usepackage{bm}                     % for \bm bold math
\usepackage{tcolorbox}              % for tinted gain boxes (optional but recommended)============================================================
 
\definecolor{clsblue}{HTML}{1F4E79}      % deep blue for CLS (the hero)
\definecolor{clsbg}{HTML}{E8F0F9}        % soft blue background for CLS row
\definecolor{s2sgray}{HTML}{6B6B6B}      % neutral gray for S2S
\definecolor{s2sbg}{HTML}{F2F2F2}        % light gray background for S2S row
\definecolor{basegray}{HTML}{9A9A9A}     % lighter gray for Base
\definecolor{gainpos}{HTML}{1B7A3E}      % green for positive gain
\definecolor{gainneg}{HTML}{B23A3A}      % red for regression
\definecolor{gainposbg}{HTML}{D4EDDA}    % light green tint for gain background
\definecolor{gainnegbg}{HTML}{F8D7DA}    % light red tint for drop background
\definecolor{rulegray}{HTML}{D0D0D0}

\definecolor{clsblue}{HTML}{1F4E79}
\definecolor{clsbg}{HTML}{E8F0F9}
\definecolor{s2sgray}{HTML}{6B6B6B}
\definecolor{s2sbg}{HTML}{F2F2F2}
 
\newcommand{\drop}[1]{\,\textcolor{gainneg}{\small\textbf{$\bm{\downarrow}$\,#1}}}
\definecolor{gainneutral}{HTML}{6B6B6B}
\newcommand{\tie}[1]{\,\textcolor{gainneutral}{\small\textbf{$\bm{=}$\,#1}}}

\definecolor{clsblue}{HTML}{1F4E79}
\definecolor{clsbg}{HTML}{E8F0F9}
\definecolor{s2sgray}{HTML}{6B6B6B}
\definecolor{s2sbg}{HTML}{F2F2F2}
\definecolor{basegray}{HTML}{9A9A9A}
\definecolor{gainpos}{HTML}{1B7A3E}
\definecolor{gainneg}{HTML}{B23A3A}
\definecolor{gainposbg}{HTML}{D4EDDA}
\definecolor{gainnegbg}{HTML}{F8D7DA}
\definecolor{gainneutralbg}{HTML}{ECECEC}
\definecolor{gainneutral}{HTML}{6B6B6B}
\definecolor{rulegray}{HTML}{B0B0B0}

\newcommand{\toofar}{\textcolor{gainneg}{\small\textbf{$\times$ too early}}}
\newcommand{\borderline}{\textcolor{gainneutral}{\small\textbf{$\sim$ borderline}}}
\newcommand{\justright}{\textcolor{gainpos}{\small\textbf{$\checkmark$ prescribed}}}

\definecolor{clsblue}{HTML}{1F4E79}
\definecolor{clsbg}{HTML}{E8F0F9}
\definecolor{s2sgray}{HTML}{6B6B6B}
\definecolor{s2sbg}{HTML}{F2F2F2}
\definecolor{basegray}{HTML}{9A9A9A}
\definecolor{gainpos}{HTML}{1B7A3E}
\definecolor{gainneg}{HTML}{B23A3A}
\definecolor{gainposbg}{HTML}{D4EDDA}
\definecolor{gainnegbg}{HTML}{F8D7DA}
\definecolor{gainneutralbg}{HTML}{ECECEC}
\definecolor{gainneutral}{HTML}{6B6B6B}
\definecolor{rulegray}{HTML}{B0B0B0}
\definecolor{groupbg}{HTML}{FAFAFA}
\title{How Fine-Tuning Objectives Shape Layer-Wise Information in LLM Hallucination Detection}

\author{%
  Kaiyang Wan\textsuperscript{$\spadesuit\,\clubsuit$}\quad
  Forrest Sheng Bao\textsuperscript{$\clubsuit$}\quad
  Amin Ahmad\textsuperscript{$\clubsuit$}\quad
  Yuxia Wang\textsuperscript{$\spadesuit$} \\
  \textsuperscript{$\spadesuit$}INSAIT, Sofia University ``St. Kliment Ohridski'', Sofia, Bulgaria \\
  \textsuperscript{$\clubsuit$}PAIR Systems, Inc., Mountain View, California \\
  \texttt{\{kaiyang.wan,\,yuxia.wang\}@insait.ai}\qquad
  \texttt{\{forrest,\,amin.ahmad\}@pairsys.ai}
}

\begin{document}

\maketitle

\begin{abstract}
Under matched conditions, classifier-head fine-tuning (\textsc{cls}) paradigm consistently outperforms the LLM-as-Judge sequence-to-sequence (\textsc{s2s}) on binary hallucination detection. 
We trace this gap to the representations induced by the two objectives. 
Static analysis shows that \textsc{cls} learns a task-specific discriminative direction, pushing hidden states into a clear two-cluster geometry with stronger class separation and linearly decodable label information. 
By contrast, \textsc{s2s} relies on the pretrained language-modeling head to make decisions. This minimizes the pressure to reorganize representations, leaving the hidden states much closer to the pretrained backbone.
Our analysis also reveals that the \textsc{cls} signal saturates before the final layer, after which upper layers form a plateau of statistically interchangeable and largely redundant representations.
However, endpoint redundancy alone does not show whether these layers are dispensable during training. 
Further training-dynamics analysis reveals 1) a lower pretrained substrate, 2) a narrow upper-middle \textbf{InfoWindow} where task information emerges, and 3) a later-synchronizing upper plateau that mirrors this signal.
This mechanism motivates us to remove layers above the InfoWindow. 
Truncating 31\% of the network leaves accuracy unchanged within seed noise on Qwen3-4B and 8B and cuts latency by $1.4\times$. 
The same depth fraction, fixed before any analysis of the target models, transfers to Llama-3.2-3B and Mistral-7B-v0.3, and the layer structure behind it reappears there and on QNLI and SST-2.
We hope this helps guide efficient task-specific LLM design.
\end{abstract}

\section{Introduction}
\label{sec:intro}

% Hallucination detection~\citep{manakul2023selfcheckgpt,belyi2025luna,orgadllms} is central to LLM-based~\citep{yang2025qwen3,grattafiori2024llama,jiang2023mistral7b} RAG~\citep{lewis2020retrieval,gao2023retrieval} and agentic systems~\citep{madaan2023self,wan2025cognitive}, where downstream modules and users need reliable, efficient verification of model outputs. 
% We study hallucination detectors fine-tuned from pretrained LLM backbones on labeled data, comparing two common paradigms. 
% \textbf{1)} LLM-as-Judge sequence-to-sequence fine-tuning (\textsc{s2s})~\citep{gu2024survey,zheng2023judging} generates a verbal verdict such as \texttt{Yes} or \texttt{No}. 
% \textbf{2)} Classifier-head fine-tuning (\textsc{cls})~\citep{houlsby2019parameter,tang2024minicheck} attaches a task-specific linear head for binary prediction.

Hallucination detection is to identify whether a model produces unsupported content relative to its input or factually false claims contradicting world knowledge~\citep{manakul2023selfcheckgpt,wang-etal-2024-factuality}.
It is central to the built-in RAG in LLMs~\citep{lewis2020retrieval,gao2023retrieval} and agentic systems~\citep{madaan2023self,wan2025cognitive,wan2026fano}, where downstream modules and users demand reliable, efficient verification of model outputs~\citep{belyi2025luna,orgadllms}. 
% ~\citep{yang2025qwen3,grattafiori2024llama,jiang2023mistral7b}
We study hallucination detectors fine-tuned using labeled data based on pretrained LLM backbones, comparing two paradigms: 
\textbf{1)} LLM-as-Judge sequence-to-sequence fine-tuning (\textsc{s2s}) generates a verbal verdict such as \texttt{Yes} or \texttt{No}~\citep{gu2024survey,zheng2023judging}; and
\textbf{2)} Classifier-head fine-tuning (\textsc{cls}) attaches a task-specific linear head for binary prediction~\citep{houlsby2019parameter,tang2024minicheck}.

% At first sight, \textsc{s2s} appears to possess the stronger inductive prior. 
% \textsc{s2s} preserves the pretrained language-modeling head, operates over a full-vocabulary output space, and benefits from representations shaped by large-scale next-token prediction~\citep{radford2019language,brown2020language}. 
% \textsc{cls}, in contrast, replaces this generative readout with a randomly initialized binary projection~\citep{wolf2020transformers}. 
% This comparison would suggest that \textsc{s2s} should provide the stronger detector. 
% Yet under matched fine-tuning conditions, \textsc{cls} consistently outperforms \textsc{s2s} across model families and evaluation splits. 
% This empirical reversal motivates our central question: why does a seemingly weaker discriminative head yield stronger hallucination detectors?

At first sight, \textsc{s2s} appears to benefit from a stronger inductive prior. 
\textsc{s2s} preserves the pretrained LLM head and optimizes the same objective. It directly exploits representations shaped by large-scale next-token prediction~\citep{radford2019language,brown2020language}. 
\textsc{cls}, in contrast, replaces the generative readout with a randomly initialized binary projection, introducing new parameters and an objective departing from LLM's pretraining objective~\citep{wolf2020transformers}, which weakens the transfer of pretrained knowledge.
This comparison would suggest that \textsc{s2s} should provide a stronger detector. 
Yet under matched fine-tuning conditions, \textsc{cls} outperforms \textsc{s2s} on every backbone we test, and the gap survives reseeding, likelihood scoring, and calibration. 
This empirical reversal motivates our central question: \textit{why does a ``weaker'' discriminative head yield stronger hallucination detectors?}

% Our analysis begins with a static view of the fine-tuned representation space.
% This analysis shows that the two objectives induce markedly different hidden-state geometries. 
In a static view of the fine-tuned representation space, we find that the two objectives induce markedly different hidden-state geometries.
Under \textsc{cls}, the model learns a task-specific discriminative direction and reorganizes hidden states into a linearly decodable two-cluster structure. 
Under \textsc{s2s}, the model instead mediates the binary decision through a pretrained full-vocabulary language-modeling head and leaves the hidden representation substantially closer to the pretrained backbone. 
We quantify this divergence with Fisher scores~\citep{fisher1936use}, which reveal substantially stronger class separation by \textsc{cls}. 
We further use layer-wise $\mathcal{V}$-information~\citep{xutheory} and linear probing~\citep{alain2016understanding,belinkov2022probing} to show that \textsc{cls} exposes considerably more linearly decodable label information.

The static analysis also shows that the \textsc{cls} signal saturates in the mid-to-upper layers, well before the final layer, forming a plateau of statistically interchangeable representations. 
This raises whether the plateau contains complementary task information or redundant encodings of the same binary signal. 
The answer implies different interventions: complementarity favors layer fusion, whereas redundancy favors upper-layer truncation or early exiting~\citep{xin2020deebert, fanreducing}. 
We test these possibilities with bootstrap layer-interchangeability tests~\citep{tibshirani1993introduction} and partial information decomposition (PID), which separates redundant, unique, and synergistic contributions across sources~\citep{williams2010nonnegative, bertschinger2014quantifying}. 
Both analyses support redundancy, suggesting that the plateau mainly preserves an already learned binary signal.

However, static redundancy alone cannot infer that the upper layers are dispensable during fine-tuning. 
% These layers could still be necessary for optimization if they form the task signal early and subsequently shape the representations below. 
These layers could still be necessary during optimization if they form task signals early and propagate gradients that influence how lower layers update.
We therefore examine the training trajectory to identify where the classification signal first emerges and how it propagates across depth. 
Our training-dynamics analysis reveals that task-relevant information grows only within a narrow upper-middle band of three to four layers. 
Below this band, label decodability remains close to the pretrained level. 
Within the band, label information accumulates rapidly, so we name it the \textbf{InfoWindow}. 
% Above it, layers align only later and mainly inherit the signal produced within the InfoWindow. 
Above it, layers become aligned with the task only at later stage of training and mainly inherit the signal produced within the InfoWindow.
Together with the endpoint redundancy analysis, this temporal ordering suggests that the upper plateau reflects a signal that has already formed below.

This mechanism yields a principled early-exit strategy. 
We retain the lower substrate, the InfoWindow, and a short stabilization margin, and remove the redundant layers above them. 
On Qwen3-4B and 8B, this truncation removes 31\% of the network, keeps accuracy within seed noise of full depth, and lowers latency by $1.4\times$. 
Since the static and dynamic analyses require substantial computation, we further evaluate whether the derived rule generalizes beyond the analyzed backbone. 
We fix the exit depths for Llama-3.2-3B and Mistral-7B-v0.3 as fractions of depth before running any analysis on them, and the transfer succeeds. 
Per-layer probing run afterwards finds the same substrate, InfoWindow, and plateau in both models, and again on QNLI and SST-2, suggesting that InfoWindow-guided truncation provides a practical route toward efficient task-specific LLM fine-tuning.

Our contributions are as follows: \textbf{i)} We measure the \textsc{cls}--\textsc{s2s} gap under matched conditions and locate it in the learned representations, where \textsc{cls} induces a stronger two-cluster geometry and higher label decodability that a fresh linear head cannot recover from \textsc{s2s} features (Sections~\ref{sec:setup} and~\ref{sec:objective-comparison}). \textbf{ii)} We show that the \textsc{cls} signal saturates into a statistically interchangeable and largely redundant upper plateau, as measured by bootstrap confidence intervals and PID (Section~\ref{sec:plateau}). \textbf{iii)} We identify an \textbf{InfoWindow} training dynamic, in which task information accumulates in a narrow upper-middle band and later synchronizes across the upper plateau (Section~\ref{sec:dynamic}). \textbf{iv)} We turn these analyses into an early-exit rule that removes 31\% of Qwen3 layers with accuracy unchanged within seed noise and $1.4\times$ lower latency, and transfers as a depth prior to Llama-3.2-3B and Mistral-7B-v0.3, where probing confirms the structure the prior assumes (Section~\ref{sec:early-exit}, Appendix~\ref{app:robustness}).

\section{Preliminaries}
\label{sec:prelim}
 
\textbf{Problem Formulation.} 
We formulate binary hallucination detection as sequence classification. 
Each example is a context--claim pair $x=(c,s)$ with label $Y \in \{0,1\}$, where $Y=1$ indicates that context $c$ supports claim $s$, and $Y=0$ indicates otherwise. 
A detector maps $x$ to a prediction $\hat{Y}$. 
We fine-tune a pretrained decoder-only LLM backbone $f_\theta$ with $L$ transformer blocks, and denote by $H_\ell(x)\in\mathbb{R}^D$ the hidden state of the final input token after block $\ell$.
 
\textbf{Fine-Tuning Paradigms.}
We compare two head--objective pairs under matched training conditions, isolating the fine-tuning paradigm as the only experimental variable. 

The \colorbox{s2sbg}{\textcolor{s2sgray}{\textbf{LLM-as-Judge paradigm}} (\textsc{s2s})} keeps the pretrained language-modeling head $W_{\mathrm{LM}}\in\mathbb{R}^{|V|\times D}$ and casts detection as verdict generation. 
Given input $x$, the model generates a verbalized label sequence $y_{1:T}$ corresponding to \texttt{yes} or \texttt{no} and minimizes
\begin{equation}
    \mathcal{L}_{\textsc{s2s}}(\theta)
    =
    -\sum_{t=1}^{T}
    \log
    \mathrm{softmax}\!\left(W_{\mathrm{LM}} h_t\right)_{y_t}.
    \label{eq:s2s}
\end{equation}

The \colorbox{clsbg}{\textcolor{clsblue}{\textbf{classifier-head paradigm}} (\textsc{cls})} replaces the generative head with a newly initialized two-way linear head $W_{\textsc{cls}}\in\mathbb{R}^{2\times D}$ on the final hidden state $H_L(x)$ and minimizes
\begin{equation}
    \mathcal{L}_{\textsc{cls}}(\theta,W_{\textsc{cls}})
    =
    -\log
    \mathrm{softmax}\!\left(W_{\textsc{cls}} H_L(x)\right)_{Y}.
    \label{eq:cls}
\end{equation}

Thus, we ask how replacing a pretrained full-vocabulary generative readout with a task-specific binary discriminative readout reshapes the hidden representations used for hallucination detection.

\section{Empirical Gap Between \textsc{cls} and \textsc{s2s}}
\label{sec:setup}

\begin{table}[ht]
\centering
\caption{\textbf{Hallucination detection under matched fine-tuning conditions.}
Macro-F$_1$ (\%) is reported as mean $\pm$ standard deviation over three fine-tuning seeds (42, 43, 44); \textsc{base} is evaluated once.
ID, OOD, and All are unweighted averages over the corresponding corpora, computed separately for each seed.
$\Delta$ reports paired \textsc{cls}$-$\textsc{s2s} differences.}
\label{tab:main}
\setlength{\tabcolsep}{3.5pt}
\renewcommand{\arraystretch}{1.15}
\resizebox{\linewidth}{!}{%
\begin{tabular}{@{} c !{\color{rulegray}\vrule} l ccc ccc !{\color{rulegray}\vrule} ccc @{}}
\toprule
\textbf{Backbone} & \textbf{Method}
& \multicolumn{3}{c}{\textbf{In-distribution}}
& \multicolumn{3}{c}{\textbf{Out-of-distribution}}
& \multicolumn{3}{c}{\textbf{Average}} \\
\cmidrule(lr){3-5} \cmidrule(lr){6-8} \cmidrule(lr){9-11}
& & RAGTruth & ClaimVerify & FactCheck & AggreFact & FaithBench & FavaBench & ID & OOD & All \\
\midrule
\multirow{4}{*}{\textit{Qwen3-4B}}
& \textcolor{basegray}{Base} & \textcolor{basegray}{55.7} & \textcolor{basegray}{70.5} & \textcolor{basegray}{71.0} & \textcolor{basegray}{80.9} & \textcolor{basegray}{24.3} & \textcolor{basegray}{69.4} & \textcolor{basegray}{65.7} & \textcolor{basegray}{58.2} & \textcolor{basegray}{62.0} \\
& \cellcolor{s2sbg}\textcolor{s2sgray}{S2S LoRA} & \cellcolor{s2sbg}81.7\,{\scriptsize$\pm$1.1} & \cellcolor{s2sbg}79.3\,{\scriptsize$\pm$0.5} & \cellcolor{s2sbg}72.5\,{\scriptsize$\pm$3.0} & \cellcolor{s2sbg}77.9\,{\scriptsize$\pm$1.1} & \cellcolor{s2sbg}27.3\,{\scriptsize$\pm$1.5} & \cellcolor{s2sbg}45.2\,{\scriptsize$\pm$4.2} & \cellcolor{s2sbg}77.8\,{\scriptsize$\pm$1.0} & \cellcolor{s2sbg}50.1\,{\scriptsize$\pm$2.1} & \cellcolor{s2sbg}64.0\,{\scriptsize$\pm$0.9} \\
& \cellcolor{clsbg}\textcolor{clsblue}{\textbf{CLS LoRA}} & \cellcolor{clsbg}86.8\,{\scriptsize$\pm$0.1} & \cellcolor{clsbg}79.4\,{\scriptsize$\pm$1.0} & \cellcolor{clsbg}76.9\,{\scriptsize$\pm$2.1} & \cellcolor{clsbg}83.1\,{\scriptsize$\pm$0.3} & \cellcolor{clsbg}32.7\,{\scriptsize$\pm$0.3} & \cellcolor{clsbg}96.7\,{\scriptsize$\pm$0.5} & \cellcolor{clsbg}81.0\,{\scriptsize$\pm$1.0} & \cellcolor{clsbg}70.8\,{\scriptsize$\pm$0.4} & \cellcolor{clsbg}75.9\,{\scriptsize$\pm$0.5} \\
& $\Delta$ (\textsc{cls}$-$\textsc{s2s}) & \textcolor{gainpos}{+5.1\,{\scriptsize$\pm$1.1}} & \textcolor{gainpos}{+0.1\,{\scriptsize$\pm$1.5}} & \textcolor{gainpos}{+4.4\,{\scriptsize$\pm$2.0}} & \textcolor{gainpos}{+5.2\,{\scriptsize$\pm$1.1}} & \textcolor{gainpos}{+5.4\,{\scriptsize$\pm$1.6}} & \textcolor{gainpos}{+51.5\,{\scriptsize$\pm$4.0}} & \textcolor{gainpos}{+3.2\,{\scriptsize$\pm$0.7}} & \textcolor{gainpos}{+20.7\,{\scriptsize$\pm$2.0}} & \textcolor{gainpos}{+11.9\,{\scriptsize$\pm$1.2}} \\
\midrule
\multirow{4}{*}{\textit{Qwen3-8B}}
& \textcolor{basegray}{Base} & \textcolor{basegray}{50.6} & \textcolor{basegray}{69.4} & \textcolor{basegray}{72.2} & \textcolor{basegray}{80.0} & \textcolor{basegray}{21.8} & \textcolor{basegray}{86.6} & \textcolor{basegray}{64.1} & \textcolor{basegray}{62.8} & \textcolor{basegray}{63.4} \\
& \cellcolor{s2sbg}\textcolor{s2sgray}{S2S LoRA} & \cellcolor{s2sbg}83.1\,{\scriptsize$\pm$0.2} & \cellcolor{s2sbg}78.3\,{\scriptsize$\pm$0.4} & \cellcolor{s2sbg}73.7\,{\scriptsize$\pm$1.4} & \cellcolor{s2sbg}79.7\,{\scriptsize$\pm$1.6} & \cellcolor{s2sbg}32.5\,{\scriptsize$\pm$0.4} & \cellcolor{s2sbg}81.2\,{\scriptsize$\pm$7.6} & \cellcolor{s2sbg}78.4\,{\scriptsize$\pm$0.5} & \cellcolor{s2sbg}64.5\,{\scriptsize$\pm$3.0} & \cellcolor{s2sbg}71.4\,{\scriptsize$\pm$1.3} \\
& \cellcolor{clsbg}\textcolor{clsblue}{\textbf{CLS LoRA}} & \cellcolor{clsbg}86.8\,{\scriptsize$\pm$0.3} & \cellcolor{clsbg}80.0\,{\scriptsize$\pm$0.6} & \cellcolor{clsbg}77.1\,{\scriptsize$\pm$1.5} & \cellcolor{clsbg}83.7\,{\scriptsize$\pm$0.8} & \cellcolor{clsbg}34.2\,{\scriptsize$\pm$4.4} & \cellcolor{clsbg}94.1\,{\scriptsize$\pm$2.5} & \cellcolor{clsbg}81.3\,{\scriptsize$\pm$0.7} & \cellcolor{clsbg}70.7\,{\scriptsize$\pm$2.4} & \cellcolor{clsbg}76.0\,{\scriptsize$\pm$0.9} \\
& $\Delta$ (\textsc{cls}$-$\textsc{s2s}) & \textcolor{gainpos}{+3.7\,{\scriptsize$\pm$0.4}} & \textcolor{gainpos}{+1.7\,{\scriptsize$\pm$0.7}} & \textcolor{gainpos}{+3.4\,{\scriptsize$\pm$2.2}} & \textcolor{gainpos}{+3.9\,{\scriptsize$\pm$1.3}} & \textcolor{gainpos}{+1.7\,{\scriptsize$\pm$4.1}} & \textcolor{gainpos}{+12.9\,{\scriptsize$\pm$5.7}} & \textcolor{gainpos}{+2.9\,{\scriptsize$\pm$0.7}} & \textcolor{gainpos}{+6.2\,{\scriptsize$\pm$2.4}} & \textcolor{gainpos}{+4.6\,{\scriptsize$\pm$0.9}} \\
\midrule
\multirow{4}{*}{\textit{Llama-3.2-3B}}
& \textcolor{basegray}{Base} & \textcolor{basegray}{41.4} & \textcolor{basegray}{36.0} & \textcolor{basegray}{44.1} & \textcolor{basegray}{29.1} & \textcolor{basegray}{43.5} & \textcolor{basegray}{43.4} & \textcolor{basegray}{40.5} & \textcolor{basegray}{38.7} & \textcolor{basegray}{39.6} \\
& \cellcolor{s2sbg}\textcolor{s2sgray}{S2S LoRA} & \cellcolor{s2sbg}78.3\,{\scriptsize$\pm$0.4} & \cellcolor{s2sbg}77.5\,{\scriptsize$\pm$0.9} & \cellcolor{s2sbg}70.5\,{\scriptsize$\pm$1.2} & \cellcolor{s2sbg}78.6\,{\scriptsize$\pm$1.3} & \cellcolor{s2sbg}28.6\,{\scriptsize$\pm$2.1} & \cellcolor{s2sbg}40.0\,{\scriptsize$\pm$7.0} & \cellcolor{s2sbg}75.4\,{\scriptsize$\pm$0.8} & \cellcolor{s2sbg}49.1\,{\scriptsize$\pm$2.8} & \cellcolor{s2sbg}62.3\,{\scriptsize$\pm$1.6} \\
& \cellcolor{clsbg}\textcolor{clsblue}{\textbf{CLS LoRA}} & \cellcolor{clsbg}84.6\,{\scriptsize$\pm$0.7} & \cellcolor{clsbg}79.6\,{\scriptsize$\pm$0.5} & \cellcolor{clsbg}73.2\,{\scriptsize$\pm$2.3} & \cellcolor{clsbg}77.6\,{\scriptsize$\pm$1.9} & \cellcolor{clsbg}29.8\,{\scriptsize$\pm$1.6} & \cellcolor{clsbg}49.6\,{\scriptsize$\pm$18.8} & \cellcolor{clsbg}79.1\,{\scriptsize$\pm$0.8} & \cellcolor{clsbg}52.3\,{\scriptsize$\pm$7.4} & \cellcolor{clsbg}65.7\,{\scriptsize$\pm$3.3} \\
& $\Delta$ (\textsc{cls}$-$\textsc{s2s}) & \textcolor{gainpos}{+6.3\,{\scriptsize$\pm$0.3}} & \textcolor{gainpos}{+2.1\,{\scriptsize$\pm$1.3}} & \textcolor{gainpos}{+2.8\,{\scriptsize$\pm$2.2}} & \textcolor{gainneg}{-1.0\,{\scriptsize$\pm$2.1}} & \textcolor{gainpos}{+1.2\,{\scriptsize$\pm$3.7}} & \textcolor{gainpos}{+9.5\,{\scriptsize$\pm$25.4}} & \textcolor{gainpos}{+3.7\,{\scriptsize$\pm$1.0}} & \textcolor{gainpos}{+3.2\,{\scriptsize$\pm$10.1}} & \textcolor{gainpos}{+3.5\,{\scriptsize$\pm$4.8}} \\
\midrule
\multirow{4}{*}{\textit{Mistral-7B-v0.3}}
& \textcolor{basegray}{Base} & \textcolor{basegray}{38.4} & \textcolor{basegray}{46.8} & \textcolor{basegray}{32.3} & \textcolor{basegray}{45.7} & \textcolor{basegray}{21.7} & \textcolor{basegray}{3.2} & \textcolor{basegray}{39.2} & \textcolor{basegray}{23.5} & \textcolor{basegray}{31.4} \\
& \cellcolor{s2sbg}\textcolor{s2sgray}{S2S LoRA} & \cellcolor{s2sbg}81.4\,{\scriptsize$\pm$0.6} & \cellcolor{s2sbg}79.5\,{\scriptsize$\pm$0.6} & \cellcolor{s2sbg}72.8\,{\scriptsize$\pm$0.6} & \cellcolor{s2sbg}79.6\,{\scriptsize$\pm$1.3} & \cellcolor{s2sbg}25.6\,{\scriptsize$\pm$0.6} & \cellcolor{s2sbg}63.0\,{\scriptsize$\pm$5.4} & \cellcolor{s2sbg}77.9\,{\scriptsize$\pm$0.2} & \cellcolor{s2sbg}56.1\,{\scriptsize$\pm$2.1} & \cellcolor{s2sbg}67.0\,{\scriptsize$\pm$1.0} \\
& \cellcolor{clsbg}\textcolor{clsblue}{\textbf{CLS LoRA}} & \cellcolor{clsbg}83.7\,{\scriptsize$\pm$2.8} & \cellcolor{clsbg}77.7\,{\scriptsize$\pm$0.5} & \cellcolor{clsbg}75.6\,{\scriptsize$\pm$0.8} & \cellcolor{clsbg}75.9\,{\scriptsize$\pm$6.2} & \cellcolor{clsbg}25.7\,{\scriptsize$\pm$4.1} & \cellcolor{clsbg}50.6\,{\scriptsize$\pm$24.3} & \cellcolor{clsbg}79.0\,{\scriptsize$\pm$1.3} & \cellcolor{clsbg}50.7\,{\scriptsize$\pm$11.5} & \cellcolor{clsbg}64.8\,{\scriptsize$\pm$6.3} \\
& $\Delta$ (\textsc{cls}$-$\textsc{s2s}) & \textcolor{gainpos}{+2.2\,{\scriptsize$\pm$2.5}} & \textcolor{gainneg}{-1.9\,{\scriptsize$\pm$0.9}} & \textcolor{gainpos}{+2.8\,{\scriptsize$\pm$1.3}} & \textcolor{gainneg}{-3.7\,{\scriptsize$\pm$5.8}} & \textcolor{gainpos}{+0.1\,{\scriptsize$\pm$4.7}} & \textcolor{gainneg}{-12.5\,{\scriptsize$\pm$26.0}} & \textcolor{gainpos}{+1.1\,{\scriptsize$\pm$1.1}} & \textcolor{gainneg}{-5.4\,{\scriptsize$\pm$12.1}} & \textcolor{gainneg}{-2.2\,{\scriptsize$\pm$6.6}} \\
\bottomrule
\end{tabular}}
\end{table}

\textbf{Experimental Setup.} 
We compare \textsc{cls} and \textsc{s2s} under matched conditions such as data and hardware.
We train rank-$32$ LoRA~\citep{hu2022lora} adapters with $\alpha=64$ on all projection matrices in every transformer block for three epochs with AdamW at learning rate $10^{-4}$, using $22{,}827$ training examples pooled from RAGTruth~\citep{niu2024ragtruth}, ClaimVerify~\citep{liu2023evaluating}, and FactCheck-GPT~\citep{wang2024factcheck}.
We report macro-F$_1$ on the eval-half of the corresponding in-distribution test splits and on the eval-half of three unseen out-of-distribution corpora, AggreFact~\citep{tang2023understanding}, FavaBench~\citep{mishrafine}, and FaithBench~\citep{bao2025faithbench}; the remaining probe-half is reserved for the analyses in Sections~\ref{sec:objective-comparison}--\ref{sec:plateau}.
Every backbone--paradigm cell in Table~\ref{tab:main} is trained with three fine-tuning seeds (42, 43, 44); we report the mean and standard deviation and compute the \textsc{cls}$-$\textsc{s2s} gap per seed before averaging.
We provide full experimental details in Appendix~\ref{app_setup}.

\textbf{Results.}
Table~\ref{tab:main} shows a consistent gap between the two paradigms.
On the in-distribution corpora, \textsc{cls} leads \textsc{s2s} on every backbone, by $3.2\pm0.7$, $2.9\pm0.7$, $3.7\pm1.0$, and $1.1\pm1.1$ macro-F$_1$ points over three fine-tuning seeds, and the gap is positive in 11 of the 12 individual runs.
The out-of-distribution gap is larger on average but far less stable.
FavaBench has a positive rate of 0.20, and at the fixed 0.5 threshold individual seeds of either paradigm can collapse toward one class there (Appendix~\ref{app:fava}).
A paired bootstrap that excludes FavaBench still gives a gap of $+2.2$ to $+3.6$ points with $p\le0.004$ on all four backbones (Table~\ref{tab:fava}).
Stronger \textsc{s2s} readouts do not close the gap, where likelihood-ratio scoring, verbalizer ensembles, and per-corpus threshold calibration applied to both paradigms leave \textsc{cls} ahead by $2.2$ to $10.4$ points, and so does a LoRA-adapted LM head (Appendix~\ref{app:strong-s2s}).
Throughout, \textsc{s2s} emits a single verdict token, so the head and loss are the only variables, and rationale-based judges fall outside this comparison (Section~\ref{sec:conclusion}).
This pattern suggests that the two objectives induce different internal representations.
We next test this hypothesis through layer-wise geometry, information, redundancy, and training-dynamics analyses.
We use endpoint diagnostics of class geometry and linearly decodable label information to trace the performance gap to objective-induced representation differences.

\section{Objective-Induced Representation Geometry}
\label{sec:objective-comparison}

\textbf{Setup.}
We analyze the same Qwen3-4B and Qwen3-8B variants from Section~\ref{sec:setup}: pretrained \textsc{base}, \textsc{s2s}-fine-tuned, and \textsc{cls}-fine-tuned. 
We feed byte-identical inputs to all variants, cache final-token residual-stream states at each layer, and evaluate probes on held-out splits that are disjoint from the benchmark evaluation splits. 
We provide the full details in Appendix~\ref{app:tools}.

\subsection{Geometric Class Separation}
\label{sec:geometry}

From a representation perspective, if fine-tuning shapes hidden states into a binary task representation, the two classes should become more separable in hidden space. 
This criterion is natural here because both paradigms produce a binary verdict through a linear logit contrast in the final hidden state, with $w_1-w_0$ for \textsc{cls} and $w_{\texttt{Yes}}-w_{\texttt{No}}$ for our single-token \textsc{s2s} verdicts. 
We therefore compare how strongly each objective aligns hidden states with a binary decision direction. We use a shared \textsc{cls}-defined projection for visualization and a direction-aligned Fisher score for quantification.

\textbf{Visualization.}
For each scale, we set $w^{\mathrm{disc}}\propto w_1-w_0$ as the decision-normal axis and choose $w^\perp$ as the leading residual direction orthogonal to it. 
We then project \textsc{base}, \textsc{s2s}, and \textsc{cls} onto the same axes with implementation details in Appendix~\ref{app:visual}.
Figure~\ref{fig:geometry} shows that \textsc{cls} yields the clearest separation along $w^{\mathrm{disc}}$ while \textsc{base} and \textsc{s2s} remain more mixed. 
This shared-axis comparison suggests that the binary geometry is induced most strongly by the \textsc{cls} objective.

\vspace{-3mm}
\begin{figure}[th]
  \centering
  \includegraphics[width=\linewidth]{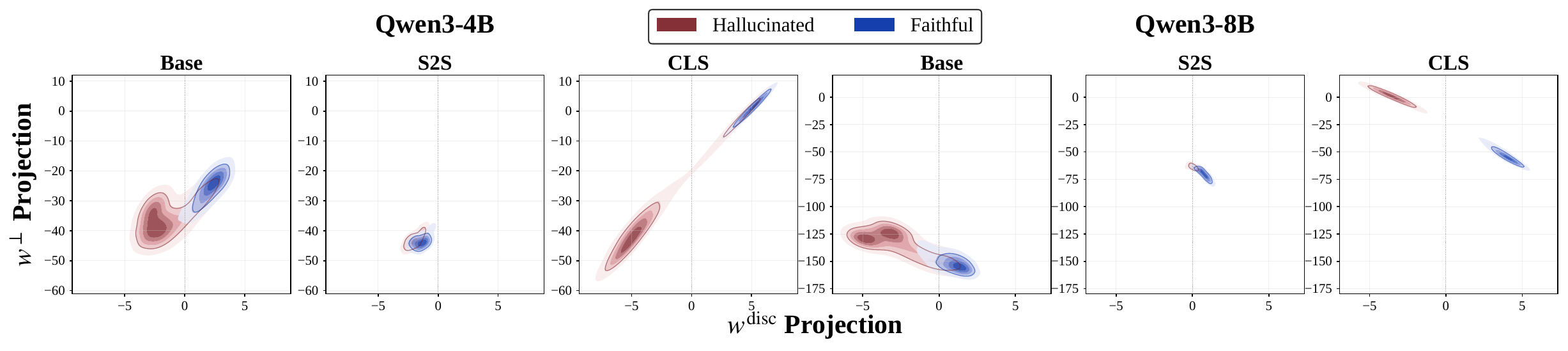}
  \vspace{-7.3mm}
  \caption{\textbf{Last-layer hidden-state geometry across paradigms and scales.}
  All variants at a fixed scale share the same \textsc{cls}-defined projection. 
  \textsc{cls} shows the clearest separation along $w^{\mathrm{disc}}$ than \textsc{s2s}.}
  \label{fig:geometry}
\end{figure}

\vspace{-0.5cm}

\begin{figure}[h]
  \centering
  \includegraphics[width=0.95\linewidth]{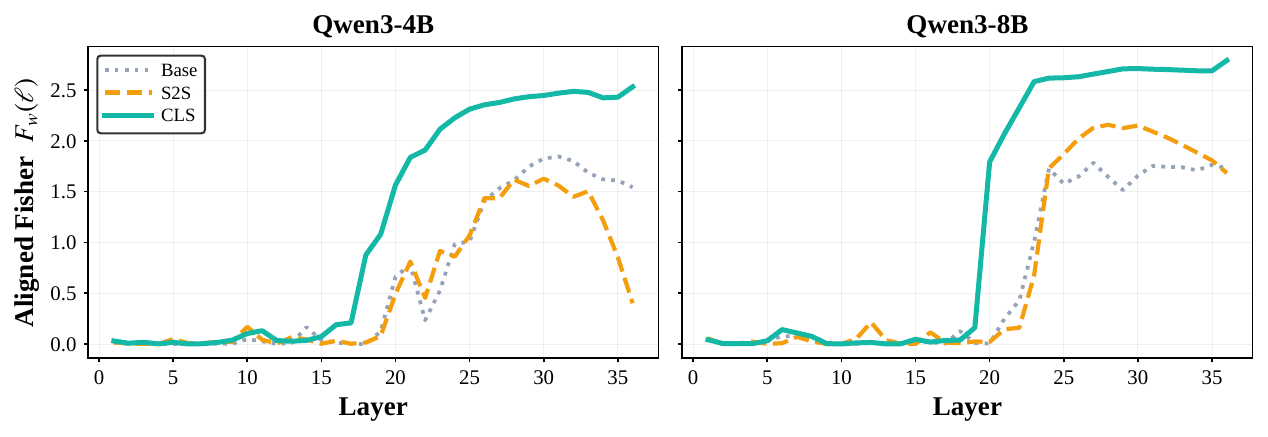}
  \vspace{-0.4cm}
  \caption{\textbf{Per-layer direction-aligned Fisher score.}
  \textsc{cls} shows an earlier and stronger rise in separation along the learned decision direction than \textsc{s2s}.}
  \label{fig:fisher_compare}
\end{figure}
\vspace{-0.1cm}

\textbf{Quantification.}
The shared projection provides a visual intuition of class separation. 
We next quantify this separation across layers by projecting each layer's hidden states onto the same \textsc{cls} decision direction $w^{\mathrm{disc}}$ and computing
\begin{equation}
F_w(\ell) \;=\; \frac{\bigl(w^{\mathrm{disc}}\!\cdot\!(\mu_1(\ell)-\mu_0(\ell))\bigr)^2}
                      {w^{\mathrm{disc}\,\top}\bigl(\Sigma_0(\ell)+\Sigma_1(\ell)\bigr)w^{\mathrm{disc}}},
\end{equation}
where $\mu_y(\ell)$ and $\Sigma_y(\ell)$ denote the class-conditional mean and covariance at layer $\ell$. 
The score $F_w$ extends the horizontal separation in Fig.~\ref{fig:geometry} to every layer by measuring decision-axis class separation normalized by within-class spread, with full details provided in Appendix~\ref{app:geometry}.
Figure~\ref{fig:fisher_compare} shows that \textsc{cls} rises earlier and reaches a higher aligned-separation plateau than \textsc{s2s} at both scales.

Together, the visualization and Fisher analysis show that \textsc{cls} more strongly shapes hidden states into a binary geometry than \textsc{s2s}. 
However, geometry is a structural diagnostic which does not measure how much label information a downstream readout can extract. 
We therefore next quantify linearly extractable label information across layers.

\subsection{V-Information and Linear-Probe Decodability}
\label{sec:vinfo}

Geometry captures class separation but does not directly measure how much label information a simple readout can extract. 
For every layer $\ell$, we train a two-way logistic probe 
$g_\ell:\mathbb{R}^D\rightarrow\Delta^2$ on the final-token hidden state 
$h_\ell(x)\in\mathbb{R}^D$ and evaluate it on the in-domain held-out split.
We report macro-F$_1$ for classification and $\mathcal{V}$-information~\citep{xutheory} for calibrated label information:
\begin{equation}
    I_{\mathcal{V}}(\ell)
    =
    H(Y)-\mathrm{CE}(g_{\ell}),
\end{equation}
where $H(Y)$ is the label entropy and $\mathrm{CE}(g_{\ell})$ is the temperature-scaled cross-entropy of the layer-$\ell$ probe. 
Thus, $I_{\mathcal{V}}(\ell)$ measures how much label uncertainty a linear readout removes from the hidden state. 
We provide the full probing details in Appendix~\ref{app:vinfo}.

\vspace{-0.2cm}

\begin{figure}[h]
  \centering
  \includegraphics[width=\linewidth]{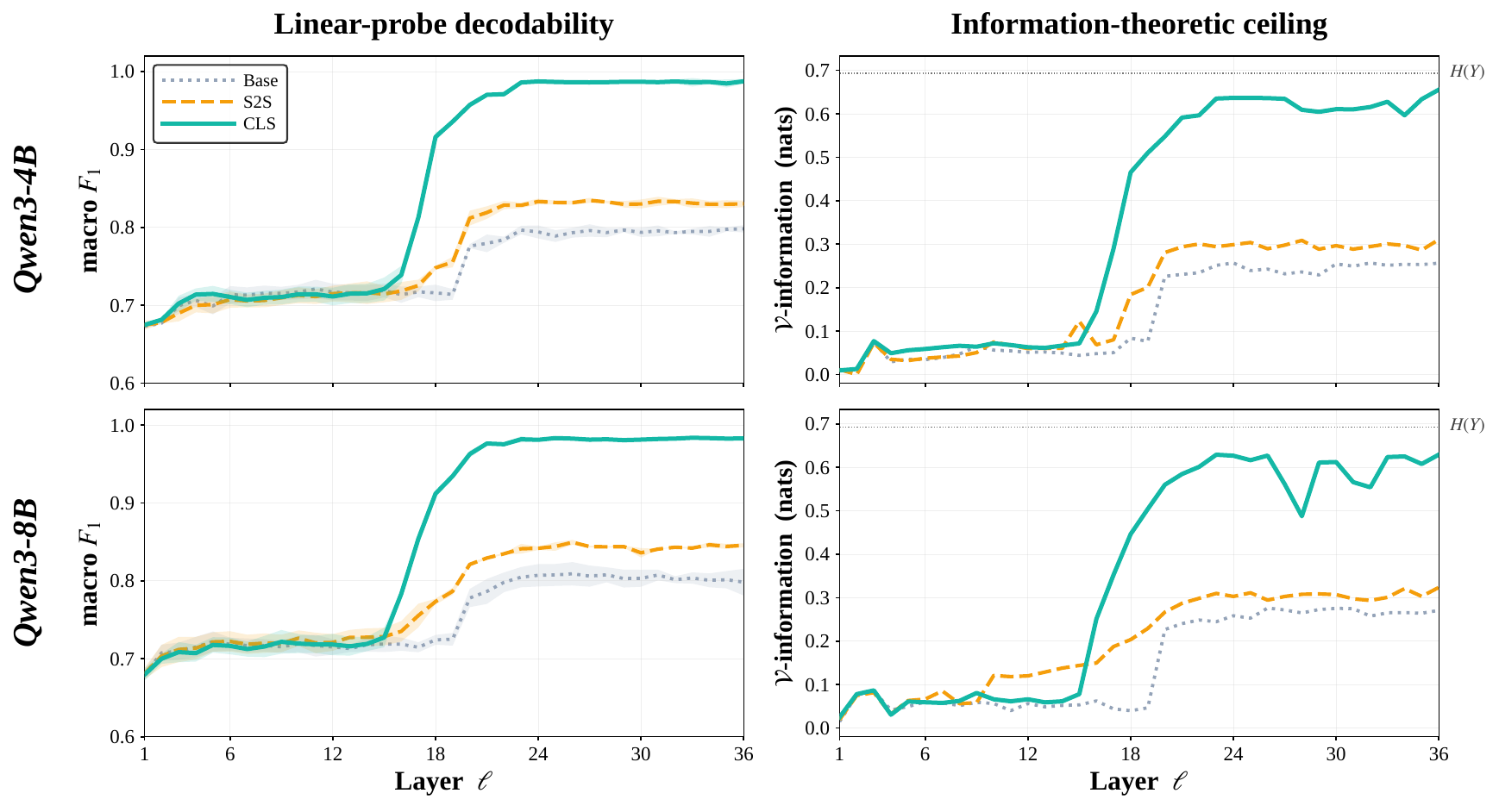}
  \vspace{-0.2cm}
  \caption{\textbf{Per-layer probe F$_1$ and $\mathcal{V}$-information.}
  \textsc{cls} rises sharply in the mid-to-upper layers and saturates near the label-entropy ceiling, whereas \textsc{s2s} improves modestly over \textsc{base}.}
  \label{fig:vinfo}
\end{figure}

Figure~\ref{fig:vinfo} shows that \textsc{cls} exposes more label information to a linear readout than \textsc{s2s}. 
Probe F$_1$ and $I_{\mathcal{V}}$ rise sharply in the mid-to-upper \textsc{cls} layers and saturate near the label-entropy ceiling, while \textsc{s2s} improves gradually over the pretrained backbone and plateaus lower. 
Together with the Fisher analysis, this shows that \textsc{cls} aligns hidden states with a binary decision direction and makes the signal more linearly decodable, whereas \textsc{s2s} remains closer to the pretrained representation.
The same ordering holds when both heads are discarded and one fresh linear readout is fit on frozen features: the \textsc{s2s} representation trails \textsc{cls} by $0.13$ to $0.15$ F$_1$ on all four backbones (Appendix~\ref{app:cross-model}).

The layer-wise curves also reveal a second, more actionable pattern: \textit{\textbf{\textsc{cls} reaches near-final label decodability well before the final layer}}. 
Once the signal becomes decodable, the key question is what the remaining layers add. 
If they contribute complementary evidence, the plateau is informationally heterogeneous and may benefit from layer fusion. 
If they mainly preserve the same binary signal, the plateau is homogeneous and may be safely truncated. 
Therefore in next section, we analyze the information structure of the saturated \textsc{cls} representations.

\section{The Upper Plateau in \textsc{cls} Representations}
\label{sec:plateau}

The early saturation in Section~\ref{sec:vinfo} raises a new question: \textit{\textbf{what do the remaining upper layers add after the task signal is already decodable?}} 
In this section, we address this question in three steps.

\subsection{Upper-Layer Saturation and Plateau Interchangeability}
\label{sec:saturation}

The $\mathcal V$-information curves of Sec.~\ref{sec:vinfo} visually saturate well before the final layer.
We now formalize that observation as a statistical claim: \textit{\textbf{from which layer onward does a single linear readout match the final-layer readout?}}

For each candidate layer $\ell$, we train a logistic probe on the layer-$\ell$ hidden states and evaluate its macro-F$_1$ on the in-domain held-out split.
Let $L_{\max}$ denote the final layer.
We form the per-layer F$_1$ gap to the final layer,
\[
\Delta_{\mathrm{final}}(\ell)
\;=\;
F_1(\ell) - F_1(L_{\max}),
\]
and compute its 95\% confidence interval (CI) by paired bootstrap with $1{,}000$ replicates on the same test samples, which preserves sample-level correlation between the two probes' predictions.
We define the saturation layer as the smallest layer whose F$_1$ gap to the final layer is statistically indistinguishable from zero,
\[
L^\star
\;=\;
\min\bigl\{\, \ell \,:\, 0 \in \widehat{\mathrm{CI}}_{0.95}\bigl(\Delta_{\mathrm{final}}(\ell)\bigr)\,\bigr\}.
\]
The candidate range is the band where fine-tuning measurably raises linear-probe decodability over the pretrained backbone (L16 onward).
We provide the full protocol in Appendix~\ref{app:saturation}.

\begin{table}[h]
\centering
\caption{\textbf{Bootstrap plateau test for upper-layer saturation in \textsc{cls}.}
$L^\star$ is the lowest layer whose probe F$_1$ is statistically indistinguishable from the final-layer under a 1000-sample bootstrap 95\% CI. A confidence interval containing zero indicates readout equivalence to the final layer.}
\label{tab:lstar}
\setlength{\tabcolsep}{8pt}
\renewcommand{\arraystretch}{1.25}
\resizebox{0.6\linewidth}{!}{%
\begin{tabular}{@{} l c c c c @{}}
\toprule
\textbf{Backbone}
& $\bm{L^\star}$
& $\bm{L^\star / L_{\max}}$
& $\bm{\Delta_{\mathrm{final}}}$ \textit{\small [95\% CI]}
& \textbf{Equiv.} \\
\midrule
\textit{Qwen3-4B}
& \textbf{L22}
& \textbf{61\%}
& $\bm{+0.001}$\, \scriptsize $[-0.007,\,+0.008]$
& \textcolor{gainpos}{\textbf{$\checkmark$}} \\
\textit{Qwen3-8B}
& \textbf{L23}
& \textbf{64\%}
& $\bm{-0.002}$\, \scriptsize $[-0.009,\,+0.005]$
& \textcolor{gainpos}{\textbf{$\checkmark$}} \\
\bottomrule
\end{tabular}}
\end{table}

Table~\ref{tab:lstar} shows that \textsc{cls} reaches final-layer-equivalent readout performance well before the final layer.
$L^\star$ lies at L22 for Qwen3-4B and L23 for Qwen3-8B, corresponding to 61--64\% of the backbone, and both 95\% CIs for $\Delta_{\mathrm{final}}$ comfortably contain zero.
The upper \textsc{cls} layers therefore reach a readout-equivalent saturation point at $L^\star$ for the synchronous-saturation band.

However, readout equivalence does not imply information redundancy.
Two layers can produce the same probe F$_1$ while encoding different task-relevant evidence, and that distinction has opposite engineering implications: heterogeneous upper layers favor fusion, homogeneous upper layers favor truncation.
Therefore, we next test which case actually holds.

\subsection{Joint Probing and Partial Information Decomposition}
\label{sec:pid}

The plateau test in Section~\ref{sec:saturation} shows that earlier upper \textsc{cls} layers can match the final layer as single-layer readouts. 
However, readout equivalence does not specify the information structure of the plateau. 
Layers with the same probe F$_1$ may encode similar task information, or they may encode different task-relevant evidence that a joint readout could exploit. 
This distinction determines how the plateau should be used: homogeneous information would support truncation, whereas heterogeneous information would support fusion. 
We therefore ask: \textit{\textbf{is the upper plateau informationally homogeneous or heterogeneous?}} We study this question at two levels.

First, we test the information structure of the plateau with joint probing and partial information decomposition (PID)~\citep{williams2010nonnegative}. 
For each candidate layer $\ell$ and a plateau layer $p$, we train a fusion probe on $(H_\ell,H_p)$ and two marginal probes on $H_\ell$ and $H_p$ alone. 
We then decompose the joint $\mathcal{V}$-information into redundant, unique, and synergistic components:
\[
I_{\mathcal V}\bigl(H_\ell, H_p; Y\bigr)
\;=\;
R \,+\, UI_{\ell} \,+\, UI_{p} \,+\, \mathcal{S},
\]
where $R$ is shared information, $UI_{\ell}$ and $UI_p$ are layer-unique contributions, and $\mathcal{S}$ is synergy. 
We implement this decomposition with a framework-free min-MI bound and report the estimated atoms $\widehat R$, $\widehat{UI}_{\ell}$, $\widehat{UI}_{p}$, and $\widehat{\mathcal{S}}$; Appendix~\ref{app:pid} gives the estimator and compatibility conditions. 
A heterogeneous plateau should yield non-trivial unique or synergistic information, whereas a homogeneous plateau should concentrate most task information in the redundant component. 
To give the strongest chance for detecting heterogeneity, we report the decomposition at the synergy-optimal candidate layer,
\[
\ell^\star = \arg\max_{\ell}\, \widehat{\mathcal{S}}(\ell).
\]
If any layer carries information complementary to the plateau, this choice should make it most visible.

Second, we test the corresponding deployment implication. 
If plateau layers provide complementary evidence, a trained fusion readout should outperform a single-layer readout under the same data and optimization budget. 
We therefore compare a fusion probe against the corresponding single-layer probe on the six-corpus held-out split, covering both in-domain and out-of-domain datasets. 
We report fusion-minus-single-layer macro-F$_1$ with a 10{,}000-sample paired bootstrap stratified by dataset. 
Appendix~\ref{app:pid} gives the full implementation details.

\begin{table}[t]
\centering
\caption{\textbf{PID-style and fusion tests support upper-plateau homogeneity in \textsc{cls}.}
At the synergy-optimal layer $\ell^\star$, the candidate layer contributes zero unique information once a plateau layer is available, and the synergy budget remains small. 
Fair OOD $\Delta F_1$ reports fusion minus single-layer readout, with 10{,}000-sample paired-bootstrap 95\% CI stratified by dataset. 
Neither \textsc{cls} backbone shows a positive fusion gain, supporting truncation rather than layer fusion.}
\label{tab:pid}
\setlength{\tabcolsep}{6pt}
\renewcommand{\arraystretch}{1.25}
\resizebox{0.6\linewidth}{!}{%
\begin{tabular}{@{} l c c c c @{}}
\toprule
\textbf{Backbone}
& $\bm{\ell^\star}$
& $\bm{\widehat{UI}_{\ell}}$
& $\bm{\widehat{\mathcal{S}}}$ \textit{\small (nats)}
& \textbf{Fair OOD $\bm{\Delta F_1}$} \textit{\small [95\% CI, pp]} \\
\midrule
\textit{Qwen3-4B}
& \textbf{L18}
& \textcolor{gainpos}{$\bm{0}$}
& \textbf{0.053}
& \tie{-0.2\,pp}\,{\scriptsize\textcolor{basegray}{$[-0.6,\,+0.2]^{\,\mathrm{ns}}$}} \\
\textit{Qwen3-8B}
& \textbf{L12}
& \textcolor{gainpos}{$\bm{0}$}
& \textbf{0.049}
& \drop{0.9\,pp}\,{\scriptsize\textcolor{basegray}{$[-1.7,\,-0.0]$}} \\
\bottomrule
\end{tabular}}

\vspace{-0.5cm}
\end{table}

Table~\ref{tab:pid} supports the homogeneous-plateau interpretation. 
For both Qwen3 scales, the synergy-optimal layer contributes zero unique information once a plateau layer is available, and the synergy remains below $0.06$ nats, less than $9\%$ of the binary label entropy $H(Y)=\ln 2$. 
Repeating the decomposition with different plateau layers gives the same conclusion within seed noise. The fusion test reaches the same conclusion at the deployment level. 
Fusion changes macro-F$_1$ by $-0.2$ pp on Qwen3-4B, with a CI containing zero, and by $-0.9$ pp on Qwen3-8B. 
Thus, plateau layers behave less like complementary sources of evidence, which favors truncation over fusion.

However, static endpoint redundancy does not imply that upper layers are unnecessary during training. 
They could still help the lower or middle layers form the task signal before becoming redundant at the endpoint. 
We therefore next examine the training trajectory.

\subsection{Training Dynamics of the Plateau}
\label{sec:dynamic}

The endpoint analyses show that the upper \textsc{cls} layers contain largely redundant label information. 
However, static redundancy alone does not show how this plateau forms. 
The upper layers might still play an essential role during training if they create the task-aligned signal or sequentially refine complementary evidence across depth. 
We therefore turn to the training trajectory and ask \textit{\textbf{when different layers acquire linearly decodable label information}} during fine-tuning.

To describe how the task signal emerges, we track when each layer gains linearly decodable label information during fine-tuning. 
At each checkpoint $t$, we extract final-token hidden states at every layer, fit a fast linear probe, and compute $I_{\mathcal{V}}(\ell,t)$. 
This gives a layer-by-time trajectory of label information. We summarize each layer's trajectory with two quantities. 
The first is the total information gain,
\[
\Delta I_{\mathcal{V}}(\ell)
=
I_{\mathcal{V}}(\ell,T)-I_{\mathcal{V}}(\ell,0),
\]
which measures how much additional label information becomes linearly extractable during training. 
The second is the saturation time $t^{0.9}_{\ell}$, defined as the first checkpoint at which layer $\ell$ reaches $90\%$ of its final $I_{\mathcal{V}}$ value. 
Together, these quantities tell us whether a layer learns a task-aligned signal, and when that signal becomes decodable. We use these trajectories to compare when different layers acquire linearly decodable label information. 
We first inspect the layer-by-time structure of $I_{\mathcal{V}}(\ell,t)$ and then use the information-gain and saturation-time statistics to summarize the pattern. 
For reference, we use $\tau=0.10$ to mark non-trivial information gain and $\tau_{\mathrm{step}}=1000$ training steps to mark near-synchronous saturation. We provide the full details in Appendix~\ref{app:dynamics}. 

\begin{figure}[h]
  \centering
  \includegraphics[width=0.99\linewidth]{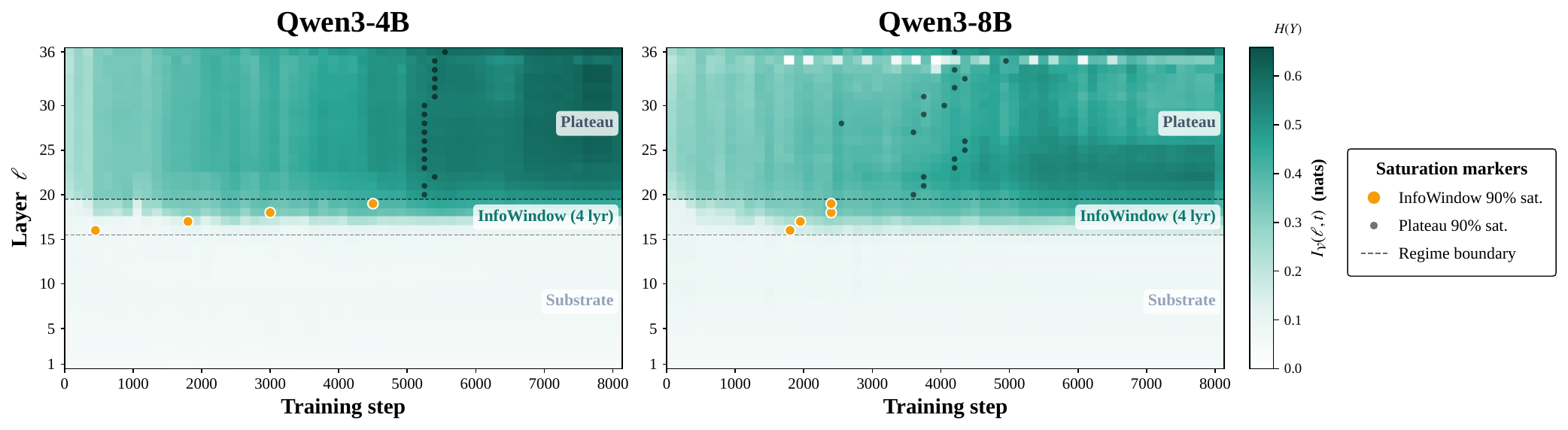}
  \vspace{-3mm}
  \caption{\textbf{Per-layer $\mathcal{V}$-information across the \textsc{cls} training trajectory.}
  The figure plots layers as rows and saved checkpoints as columns, with color indicating $I_{\mathcal{V}}(\ell,t)$ in nats. 
  The lower substrate (L1--L15) remains near its pretrained level, the InfoWindow (L16--L19) shows staggered task-information growth, and the upper plateau (L20--L36) synchronizes later within a tight training-step cluster.}
  \label{fig:iv-trajectory}
\end{figure}

Figure~\ref{fig:iv-trajectory} reveals three dynamic regimes. 
\textbf{1) Lower substrate.} Layers L1--L15 remain close to their pretrained level and gain little additional linearly decodable label information, suggesting that they mainly provide general features. 
\textbf{2) InfoWindow.} Layers L16--L19 show the first non-trivial growth in $I_{\mathcal{V}}$ and saturate in a staggered manner across depth, so we refer to this signal-forming band as the \textbf{InfoWindow}. 
Qwen3-4B exhibits an upward-moving signal through this band, while Qwen3-8B shows the same depth-staggered pattern in temporally compressed form.
\textbf{3) Synchronous upper plateau.} Layers L20--L36 reach high $\mathcal{V}$-information later and within a tight training-step window, forming a synchronous plateau. 
This temporal pattern suggests that the upper plateau aligns to an already decodable signal instead of constructing heterogeneous representation.

This temporal order connects the dynamic evidence with the endpoint redundancy analysis. 
At the endpoint, the upper plateau contains largely homogeneous label information; during training, the same plateau appears only after the InfoWindow has already formed a decodable binary signal. 
Thus, the upper layers look less like the source of task information and more like late redundant carriers of an already formed representation. 
The resulting partition motivates the early-exit rule in the next section, where we preserve the substrate and InfoWindow and test whether removing layers above the saturated plateau boundary yields a parameter-efficient detector.

\section{From Plateau Analysis to Early Exit}
\label{sec:early-exit}

In this section, we turn the InfoWindow into an early-exit intervention. 
On the analyzed Qwen3 backbones, the InfoWindow gives a principled exit rule. 
% On additional backbones, the resulting exit depth can serve as a depth-fractional prior for selecting useful analysis-free candidate exits.
For other backbones, the exit depth of Qwen3 can serve as a depth-fractional prior for selecting useful analysis-free candidate exits.

\paragraph{InfoWindow-Guided Exits on Qwen3.}
On Qwen3, the InfoWindow spans L16--L19 and the synchronous upper plateau begins at L20, while the readout-equivalence point $L^\star$ appears slightly later at L22/L23 for 4B/8B.
This gap suggests that although task information first emerges inside the InfoWindow, the representation needs a short stabilization margin before it reaches final-layer-level readout quality.
We therefore select three proof-of-concept exits aligned with these regimes: L15 cuts before the signal-forming window, L20 reads out at the first plateau layer, and L25 keeps the InfoWindow plus a five-layer stabilization margin.
Thus, L25 removes roughly 31\% of the forward-pass depth while preserving the signal-forming band and its post-InfoWindow buffer.
These three quantities are measured on the probe half alone, and the eval half enters only the final numbers in Table~\ref{tab:early-exit} (Appendix~\ref{app:early_exit}).

\paragraph{Analysis-Free Transfer via a Depth Prior.}
For Llama-3.2-3B ($L=28$) and Mistral-7B-v0.3 ($L=32$), we fix the exits before running any representation analysis on either model, so the transfer is an out-of-sample test of the depth prior. 
Instead, we transfer the Qwen3 exit as two pre-specified retained-depth ratios, with conservative $\ell_{\mathrm{exit}}=\lceil 0.70L\rceil$ and aggressive $\ell_{\mathrm{exit}}=\mathrm{round}(0.60L)$. 
We use the conservative rule to approximate Qwen3's retained ratio $25/36\approx0.69$, and we use the aggressive rule to test a deeper cut. 
These rules give Llama exits at L20/L17 and Mistral exits at L23/L19. 
We then test whether this \{conservative, aggressive\} depth prior identifies at least one competitive efficient model per backbone. 
We provide full training details in Appendix~\ref{app:early_exit}.

\begin{table}[t]
\centering
\caption{\textbf{Early-exit accuracy and retained depth.}
Macro-F$_1$ (\%) is reported as mean $\pm$ standard deviation over three fine-tuning seeds (42, 43, 44), except for L15 (seed 42 only).
``Saved'' is the fraction of transformer blocks removed; $\Delta$ is the six-corpus average difference from the full-depth model, paired by seed.
Exit depths are held fixed across runs.}
\label{tab:early-exit}
\setlength{\tabcolsep}{3.5pt}
\renewcommand{\arraystretch}{1.15}
\resizebox{\linewidth}{!}{%
\begin{tabular}{@{} c !{\color{rulegray}\vrule} l c c c ccc ccc !{\color{rulegray}\vrule} c c @{}}
\toprule
\textbf{Backbone} & \textbf{Exit position} & \textbf{Verdict} & \textbf{Layer} & \textbf{Saved}
& RAGTruth & ClaimVerify & FactCheck & AggreFact & FaithBench & FavaBench & \textbf{Avg.} & $\bm{\Delta}$ \\
\midrule
\multicolumn{13}{@{}l}{\cellcolor{groupbg}\textit{\textbf{Analyzed backbones}} \,\textcolor{basegray}{---\,exit chosen from the per-model InfoWindow analysis (Section~\ref{sec:dynamic})}} \\
\midrule
\multirow{4}{*}{\textit{Qwen3-4B}}
& \textcolor{basegray}{Full \textsc{cls} (baseline)} & \textcolor{basegray}{---} & \textcolor{basegray}{36} & \textcolor{basegray}{---} & \textcolor{basegray}{86.8\,{\scriptsize$\pm$0.1}} & \textcolor{basegray}{79.4\,{\scriptsize$\pm$1.0}} & \textcolor{basegray}{76.9\,{\scriptsize$\pm$2.1}} & \textcolor{basegray}{83.1\,{\scriptsize$\pm$0.3}} & \textcolor{basegray}{32.7\,{\scriptsize$\pm$0.3}} & \textcolor{basegray}{96.7\,{\scriptsize$\pm$0.5}} & \textcolor{basegray}{75.9\,{\scriptsize$\pm$0.5}} & \textcolor{basegray}{---} \\
& Before InfoWindow & \toofar & 15 & 58\% & 84.9 & 79.1 & 74.5 & 75.7 & 28.0 & 52.9 & 65.9 & \textcolor{gainneg}{-9.9\,{\scriptsize($n{=}1$)}} \\
& First plateau layer & \borderline & 20 & 44\% & 85.0\,{\scriptsize$\pm$1.3} & 80.1\,{\scriptsize$\pm$1.5} & 76.8\,{\scriptsize$\pm$1.2} & 79.2\,{\scriptsize$\pm$2.4} & 27.7\,{\scriptsize$\pm$4.5} & 79.2\,{\scriptsize$\pm$18.8} & 71.4\,{\scriptsize$\pm$4.3} & \textcolor{gainneg}{-4.5\,{\scriptsize$\pm$4.8}} \\
& \textbf{InfoWindow + margin} & \justright & 25 & 31\% & 86.4\,{\scriptsize$\pm$0.9} & 80.2\,{\scriptsize$\pm$0.4} & 76.7\,{\scriptsize$\pm$0.4} & 82.3\,{\scriptsize$\pm$1.5} & 32.0\,{\scriptsize$\pm$2.6} & 94.6\,{\scriptsize$\pm$2.9} & 75.4\,{\scriptsize$\pm$1.1} & \textcolor{basegray}{-0.5\,{\scriptsize$\pm$1.4}} \\
\midrule
\multirow{4}{*}{\textit{Qwen3-8B}}
& \textcolor{basegray}{Full \textsc{cls} (baseline)} & \textcolor{basegray}{---} & \textcolor{basegray}{36} & \textcolor{basegray}{---} & \textcolor{basegray}{86.8\,{\scriptsize$\pm$0.3}} & \textcolor{basegray}{80.0\,{\scriptsize$\pm$0.6}} & \textcolor{basegray}{77.1\,{\scriptsize$\pm$1.5}} & \textcolor{basegray}{83.7\,{\scriptsize$\pm$0.8}} & \textcolor{basegray}{34.2\,{\scriptsize$\pm$4.4}} & \textcolor{basegray}{94.1\,{\scriptsize$\pm$2.5}} & \textcolor{basegray}{76.0\,{\scriptsize$\pm$0.9}} & \textcolor{basegray}{---} \\
& Before InfoWindow & \toofar & 15 & 58\% & 83.3 & 79.9 & 72.5 & 75.1 & 28.7 & 37.5 & 62.8 & \textcolor{gainneg}{-14.4\,{\scriptsize($n{=}1$)}} \\
& First plateau layer & \borderline & 20 & 44\% & 85.9\,{\scriptsize$\pm$0.2} & 80.3\,{\scriptsize$\pm$0.5} & 74.9\,{\scriptsize$\pm$0.4} & 82.9\,{\scriptsize$\pm$0.3} & 30.6\,{\scriptsize$\pm$2.0} & 95.7\,{\scriptsize$\pm$0.9} & 75.1\,{\scriptsize$\pm$0.6} & \textcolor{basegray}{-0.9\,{\scriptsize$\pm$1.2}} \\
& \textbf{InfoWindow + margin} & \justright & 25 & 31\% & 86.8\,{\scriptsize$\pm$0.2} & 80.8\,{\scriptsize$\pm$0.2} & 75.1\,{\scriptsize$\pm$0.9} & 85.1\,{\scriptsize$\pm$0.1} & 35.8\,{\scriptsize$\pm$2.4} & 97.2\,{\scriptsize$\pm$0.6} & 76.8\,{\scriptsize$\pm$0.4} & \textcolor{gainpos}{+0.8\,{\scriptsize$\pm$1.1}} \\
\midrule
\multicolumn{13}{@{}l}{\cellcolor{groupbg}\textit{\textbf{Transfer backbones}} \,\textcolor{basegray}{---\,candidate exits from the depth prior ($\lceil 0.70L\rceil$ / $\mathrm{round}(0.60L)$); selected before per-model probing}} \\
\midrule
\multirow{3}{*}{\textit{Llama-3.2-3B}}
& \textcolor{basegray}{Full \textsc{cls} (baseline)} & \textcolor{basegray}{---} & \textcolor{basegray}{28} & \textcolor{basegray}{---} & \textcolor{basegray}{84.6\,{\scriptsize$\pm$0.7}} & \textcolor{basegray}{79.6\,{\scriptsize$\pm$0.5}} & \textcolor{basegray}{73.2\,{\scriptsize$\pm$2.3}} & \textcolor{basegray}{77.6\,{\scriptsize$\pm$1.9}} & \textcolor{basegray}{29.8\,{\scriptsize$\pm$1.6}} & \textcolor{basegray}{49.6\,{\scriptsize$\pm$18.8}} & \textcolor{basegray}{65.7\,{\scriptsize$\pm$3.3}} & \textcolor{basegray}{---} \\
& Conservative ($f{=}0.70$) & --- & 20 & 29\% & 85.5\,{\scriptsize$\pm$0.6} & 79.4\,{\scriptsize$\pm$0.4} & 74.2\,{\scriptsize$\pm$2.2} & 78.5\,{\scriptsize$\pm$0.5} & 32.2\,{\scriptsize$\pm$0.5} & 34.4\,{\scriptsize$\pm$5.9} & 64.0\,{\scriptsize$\pm$0.9} & \textcolor{gainneg}{-1.7\,{\scriptsize$\pm$3.8}} \\
& Aggressive ($f{=}0.60$) & --- & 17 & 39\% & 84.9\,{\scriptsize$\pm$0.2} & 81.3\,{\scriptsize$\pm$0.5} & 73.9\,{\scriptsize$\pm$1.4} & 78.9\,{\scriptsize$\pm$0.1} & 30.0\,{\scriptsize$\pm$0.4} & 55.7\,{\scriptsize$\pm$14.7} & 67.5\,{\scriptsize$\pm$2.5} & \textcolor{gainpos}{+1.7\,{\scriptsize$\pm$1.9}} \\
\midrule
\multirow{3}{*}{\textit{Mistral-7B-v0.3}}
& \textcolor{basegray}{Full \textsc{cls} (baseline)} & \textcolor{basegray}{---} & \textcolor{basegray}{32} & \textcolor{basegray}{---} & \textcolor{basegray}{83.7\,{\scriptsize$\pm$2.8}} & \textcolor{basegray}{77.7\,{\scriptsize$\pm$0.5}} & \textcolor{basegray}{75.6\,{\scriptsize$\pm$0.8}} & \textcolor{basegray}{75.9\,{\scriptsize$\pm$6.2}} & \textcolor{basegray}{25.7\,{\scriptsize$\pm$4.1}} & \textcolor{basegray}{50.6\,{\scriptsize$\pm$24.3}} & \textcolor{basegray}{64.8\,{\scriptsize$\pm$6.3}} & \textcolor{basegray}{---} \\
& Conservative ($f{=}0.70$) & --- & 23 & 28\% & 84.7\,{\scriptsize$\pm$0.1} & 78.2\,{\scriptsize$\pm$0.5} & 74.9\,{\scriptsize$\pm$1.0} & 81.0\,{\scriptsize$\pm$1.0} & 29.1\,{\scriptsize$\pm$1.2} & 74.9\,{\scriptsize$\pm$10.7} & 70.4\,{\scriptsize$\pm$1.8} & \textcolor{gainpos}{+5.6\,{\scriptsize$\pm$5.9}} \\
& Aggressive ($f{=}0.60$) & --- & 19 & 41\% & 81.6\,{\scriptsize$\pm$3.3} & 77.3\,{\scriptsize$\pm$3.2} & 75.0\,{\scriptsize$\pm$1.1} & 77.3\,{\scriptsize$\pm$4.3} & 28.0\,{\scriptsize$\pm$2.2} & 65.5\,{\scriptsize$\pm$14.2} & 67.4\,{\scriptsize$\pm$3.7} & \textcolor{gainpos}{+2.6\,{\scriptsize$\pm$2.7}} \\
\bottomrule
\end{tabular}}
\end{table}

\paragraph{Early-Exit Results.}
Table~\ref{tab:early-exit} reports macro-F$_1$ over three fine-tuning seeds.
The Qwen3 results follow the InfoWindow prediction: the L15 exits remove the signal-forming band and drop sharply, the L20 exits keep too little plateau margin and fluctuate across seeds ($-4.5\pm4.8$ on 4B), and the L25 exits preserve the InfoWindow plus a short margin and stay within seed noise of full depth ($-0.5\pm1.4$ and $+0.8\pm1.1$) with 31\% fewer layers.
The saving is realized in wall-clock time as well: with the blocks above the exit removed, per-example latency falls by $1.4\times$ at L25 and $1.8\times$ at L20 on both an H200 and an RTX A6000 (Appendix~\ref{app:latency}).
The transfer results show that the Qwen3-derived depth prior identifies one competitive efficient variant for each additional backbone, although single-seed cells move by several points, so we read the ratio as a prior over candidates rather than a prescription.
Together, these results show that the InfoWindow provides a principled early-exit rule for analyzed models and a lightweight depth prior for unanalyzed models.

\paragraph{Discussion and Analysis.}
After the transfer results, we ran the per-layer probing of Section~\ref{sec:vinfo} on both backbones, and then on Qwen3-4B fine-tuned on QNLI and SST-2 with the same recipe.
Both transfer backbones show the same substrate, InfoWindow, and plateau, with the \textsc{cls} saturation layer at $0.43$ and $0.56$ of depth.
All four pre-specified exits lie above it and all four stay within seed noise of full depth or above it, whereas the one Qwen3 exit below its saturation layer, L20, is the unstable one.
QNLI and SST-2 reproduce the structure with the saturation layer at $0.61$ and $0.50$ of depth, while the \textsc{cls} advantage shrinks to $0.02$ and $0.00$ F$_1$ because the pretrained backbone already decodes both tasks at $0.89$ to $0.94$.
Appendix~\ref{app:robustness} reports these probes together with the seed-level runs behind Tables~\ref{tab:main} and~\ref{tab:early-exit}, the stronger \textsc{s2s} readouts, the FavaBench diagnosis, and the measured latency.

\section{Related Work}
\label{sec:related}

\textbf{Hallucination detection.} Hallucination detection has been studied through factuality benchmarks, self-checking methods, LLM-as-judge protocols, and lightweight detectors over model states or outputs~\citep{manakul2023selfcheckgpt,niu2024ragtruth,wang-etal-2024-factuality,tang2024minicheck,belyi2025luna}. 
Prior work suggests that truthfulness or factuality signals can be exposed either from internal LLM representations or from model self-evaluations, including activation-based classifiers, unsupervised truth directions, activation interventions, and verbalized $P(\mathrm{True})$ estimates~\citep{azaria2023internal,burnsdiscovering,kadavath2022language,li2023inference}. 
We build on this evidence and study how such signals are shaped during \textsc{cls} and \textsc{s2s} fine-tuning.

\textbf{Layer-wise probing and representation analysis.} Layer-wise probing is widely used to study what information neural representations make accessible to simple readouts~\citep{alain2016understanding,belinkov2022probing,tenney2019bert,hewitt2019structural}. 
Information-theoretic variants further quantify usable label information and distinguish shared from complementary signals across representations~\citep{xutheory,williams2010nonnegative,bertschinger2014quantifying}.
We build on these tools to ask how two fine-tuning objectives shape the same backbone. 
We track Fisher separation, $\mathcal{V}$-information, and redundancy-related measurements across checkpoints, connecting endpoint layer-wise structure to the training dynamics that produced it.

\textbf{Early exit for transformers.} Early-exit methods reduce transformer inference cost by adding intermediate classifiers, dropping layers, or gating computation at inference time~\citep{xin2020deebert,liu2020fastbert,fanreducing,xin2021berxit,zhou2020bert,elhoushi2024layerskip,chen2024ee}. 
They make exit decisions adaptively for each input and choose exits or thresholds using validation performance or confidence criteria. 
We use layer-wise training dynamics to choose a single static cutoff for all inputs. 
The InfoWindow trajectory provides a representation-based criterion for choosing this cutoff, complementing confidence- or validation-based early-exit rules.

\section{Conclusion and Future Work}
\label{sec:conclusion}

We studied why classifier-head fine-tuning (\textsc{cls}) outperforms LLM-as-Judge sequence-to-sequence fine-tuning (\textsc{s2s}) on binary hallucination detection under matched conditions. 
Our analyses show that \textsc{cls} reshapes the backbone into a more discriminative representation, with clearer two-cluster geometry, stronger class separation, and more linearly decodable label information, whereas \textsc{s2s} leaves hidden states closer to the pretrained representation. 
The layer-wise results further show that the \textsc{cls} signal emerges in a narrow upper-middle \textbf{InfoWindow} and saturates before the final layer. 
The layers above this region form a statistically interchangeable and largely redundant plateau. 
This plateau structure motivates a simple static early-exit rule that removes layers above the InfoWindow. 
The resulting truncation removes 31\% of the network on Qwen3-4B and 8B with accuracy unchanged within seed noise and latency reduced by $1.4\times$, and the same depth fraction, fixed before any analysis of the target models, transfers to Llama-3.2-3B and Mistral-7B-v0.3. 
Probing those models afterwards finds the same substrate, InfoWindow, and plateau, and so does probing Qwen3-4B fine-tuned on QNLI and SST-2, which places the structure beyond the task and backbone family it was derived on. 
Together, these findings suggest that objective-induced representation structure can help explain detector accuracy and guide efficient task-specific LLM design.

Several directions remain open.
First, we restrict \textsc{s2s} to a single verdict token so that the objective is the only variable.
Judges that generate a rationale before the verdict add supervision and inference cost that our deployment setting does not allow, and comparing them with \textsc{cls} where that cost is acceptable is a natural next step.
Second, the layer structure is general, but the \textsc{cls} advantage is not.
It is large on hallucination detection and near zero on QNLI and SST-2 (Section~\ref{sec:early-exit}, Appendix~\ref{app:second-task}), and whether the exit rule itself transfers to such tasks remains untested.
Third, our early-exit experiments use proof-of-concept cutoff choices and one static exit for all inputs.
Future work could select exits from representation-level signals more systematically, combine InfoWindow-guided truncation with input-adaptive exits, distillation, or pruning, and design objectives that shape these representations more directly.
Fourth, we have traced how label information and redundancy are distributed across layers.
It remains open how they are distributed across features within a layer and which features carry the signal.

\bibliographystyle{plainnat}
\bibliography{references}

@article{hu2022lora,
  title={Lora: Low-rank adaptation of large language models.},
  author={Hu, Edward J and Shen, Yelong and Wallis, Phillip and Allen-Zhu, Zeyuan and Li, Yuanzhi and Wang, Shean and Wang, Liang and Chen, Weizhu and others},
  journal={Iclr},
  volume={1},
  number={2},
  pages={3},
  year={2022}
}

@inproceedings{niu2024ragtruth,
  title={Ragtruth: A hallucination corpus for developing trustworthy retrieval-augmented language models},
  author={Niu, Cheng and Wu, Yuanhao and Zhu, Juno and Xu, Siliang and Shum, KaShun and Zhong, Randy and Song, Juntong and Zhang, Tong},
  booktitle={Proceedings of the 62nd Annual Meeting of the Association for Computational Linguistics (Volume 1: Long Papers)},
  pages={10862--10878},
  year={2024}
}

@inproceedings{liu2023evaluating,
  title={Evaluating verifiability in generative search engines},
  author={Liu, Nelson F and Zhang, Tianyi and Liang, Percy},
  booktitle={Findings of the Association for Computational Linguistics: EMNLP 2023},
  pages={7001--7025},
  year={2023}
}

@inproceedings{wang2024factcheck,
  title={Factcheck-bench: Fine-grained evaluation benchmark for automatic fact-checkers},
  author={Wang, Yuxia and Reddy, Revanth Gangi and Mujahid, Zain Muhammad and Arora, Arnav and Rubashevskii, Aleksandr and Geng, Jiahui and Afzal, Osama Mohammed and Pan, Liangming and Borenstein, Nadav and Pillai, Aditya and others},
  booktitle={Findings of the Association for Computational Linguistics: EMNLP 2024},
  pages={14199--14230},
  year={2024}
}

@inproceedings{tang2023understanding,
  title={Understanding factual errors in summarization: Errors, summarizers, datasets, error detectors},
  author={Tang, Liyan and Goyal, Tanya and Fabbri, Alex and Laban, Philippe and Xu, Jiacheng and Yavuz, Semih and Kry{\'s}ci{\'n}ski, Wojciech and Rousseau, Justin and Durrett, Greg},
  booktitle={Proceedings of the 61st Annual Meeting of the Association for Computational Linguistics (Volume 1: Long Papers)},
  pages={11626--11644},
  year={2023}
}

@inproceedings{mishrafine,
  title={Fine-grained Hallucination Detection and Editing for Language Models},
  author={Mishra, Abhika and Asai, Akari and Balachandran, Vidhisha and Wang, Yizhong and Neubig, Graham and Tsvetkov, Yulia and Hajishirzi, Hannaneh},
  booktitle={First Conference on Language Modeling}
}

@inproceedings{bao2025faithbench,
  title={Faithbench: A diverse hallucination benchmark for summarization by modern llms},
  author={Bao, Forrest and Li, Miaoran and Qu, Renyi and Luo, Ge and Wan, Erana and Tang, Yujia and Fan, Weisi and Tamber, Manveer Singh and Kazi, Suleman and Sourabh, Vivek and others},
  booktitle={Proceedings of the 2025 Conference of the Nations of the Americas Chapter of the Association for Computational Linguistics: Human Language Technologies (Volume 2: Short Papers)},
  pages={448--461},
  year={2025}
}

@inproceedings{xutheory,
  title={A Theory of Usable Information under Computational Constraints},
  author={Xu, Yilun and Zhao, Shengjia and Song, Jiaming and Stewart, Russell and Ermon, Stefano},
  booktitle={International Conference on Learning Representations}
}

@inproceedings{belyi2025luna,
  title={Luna: A lightweight evaluation model to catch language model hallucinations with high accuracy and low cost},
  author={Belyi, Masha and Friel, Robert and Shao, Shuai and Sanyal, Atindriyo},
  booktitle={Proceedings of the 31st international conference on computational linguistics: industry track},
  pages={398--409},
  year={2025}
}

@inproceedings{orgadllms,
  title={LLMs Know More Than They Show: On the Intrinsic Representation of LLM Hallucinations},
  author={Orgad, Hadas and Toker, Michael and Gekhman, Zorik and Reichart, Roi and Szpektor, Idan and Kotek, Hadas and Belinkov, Yonatan},
  booktitle={The Thirteenth International Conference on Learning Representations}
}

@article{lewis2020retrieval,
  title={Retrieval-augmented generation for knowledge-intensive nlp tasks},
  author={Lewis, Patrick and Perez, Ethan and Piktus, Aleksandra and Petroni, Fabio and Karpukhin, Vladimir and Goyal, Naman and K{\"u}ttler, Heinrich and Lewis, Mike and Yih, Wen-tau and Rockt{\"a}schel, Tim and others},
  journal={Advances in neural information processing systems},
  volume={33},
  pages={9459--9474},
  year={2020}
}

@inproceedings{manakul2023selfcheckgpt,
  title={Selfcheckgpt: Zero-resource black-box hallucination detection for generative large language models},
  author={Manakul, Potsawee and Liusie, Adian and Gales, Mark},
  booktitle={Proceedings of the 2023 conference on empirical methods in natural language processing},
  pages={9004--9017},
  year={2023}
}

@article{zheng2023judging,
  title={Judging llm-as-a-judge with mt-bench and chatbot arena},
  author={Zheng, Lianmin and Chiang, Wei-Lin and Sheng, Ying and Zhuang, Siyuan and Wu, Zhanghao and Zhuang, Yonghao and Lin, Zi and Li, Zhuohan and Li, Dacheng and Xing, Eric and others},
  journal={Advances in neural information processing systems},
  volume={36},
  pages={46595--46623},
  year={2023}
}

@inproceedings{tang2024minicheck,
  title={Minicheck: Efficient fact-checking of llms on grounding documents},
  author={Tang, Liyan and Laban, Philippe and Durrett, Greg},
  booktitle={Proceedings of the 2024 Conference on Empirical Methods in Natural Language Processing},
  pages={8818--8847},
  year={2024}
}

@inproceedings{wan2025cognitive,
  title={A cognitive writing perspective for constrained long-form text generation},
  author={Wan, Kaiyang and Mu, Honglin and Hao, Rui and Luo, Haoran and Gu, Tianle and Chen, Xiuying},
  booktitle={Findings of the Association for Computational Linguistics: ACL 2025},
  pages={9832--9844},
  year={2025}
}

@article{gu2024survey,
  title={A survey on llm-as-a-judge},
  author={Gu, Jiawei and Jiang, Xuhui and Shi, Zhichao and Tan, Hexiang and Zhai, Xuehao and Xu, Chengjin and Li, Wei and Shen, Yinghan and Ma, Shengjie and Liu, Honghao and others},
  journal={The Innovation},
  publisher={Elsevier}
}

@inproceedings{houlsby2019parameter,
  title={Parameter-efficient transfer learning for NLP},
  author={Houlsby, Neil and Giurgiu, Andrei and Jastrzebski, Stanislaw and Morrone, Bruna and De Laroussilhe, Quentin and Gesmundo, Andrea and Attariyan, Mona and Gelly, Sylvain},
  booktitle={International conference on machine learning},
  pages={2790--2799},
  year={2019},
  organization={PMLR}
}

@article{gao2023retrieval,
  title={Retrieval-augmented generation for large language models: A survey},
  author={Gao, Yunfan and Xiong, Yun and Gao, Xinyu and Jia, Kangxiang and Pan, Jinliu and Bi, Yuxi and Sun, Jiawei and Wang, Haofen and others}
}

@article{madaan2023self,
  title={Self-refine: Iterative refinement with self-feedback},
  author={Madaan, Aman and Tandon, Niket and Gupta, Prakhar and Hallinan, Skyler and Gao, Luyu and Wiegreffe, Sarah and Alon, Uri and Dziri, Nouha and Prabhumoye, Shrimai and Yang, Yiming and others},
  journal={Advances in neural information processing systems},
  volume={36},
  pages={46534--46594},
  year={2023}
}

@article{radford2019language,
  title={Language models are unsupervised multitask learners},
  author={Radford, Alec and Wu, Jeffrey and Child, Rewon and Luan, David and Amodei, Dario and Sutskever, Ilya and others},
  journal={OpenAI blog},
  volume={1},
  number={8},
  pages={9},
  year={2019}
}

@article{brown2020language,
  title={Language models are few-shot learners},
  author={Brown, Tom and Mann, Benjamin and Ryder, Nick and Subbiah, Melanie and Kaplan, Jared D and Dhariwal, Prafulla and Neelakantan, Arvind and Shyam, Pranav and Sastry, Girish and Askell, Amanda and others},
  journal={Advances in neural information processing systems},
  volume={33},
  pages={1877--1901},
  year={2020}
}

@inproceedings{wolf2020transformers,
  title={Transformers: State-of-the-art natural language processing},
  author={Wolf, Thomas and Debut, Lysandre and Sanh, Victor and Chaumond, Julien and Delangue, Clement and Moi, Anthony and Cistac, Pierric and Rault, Tim and Louf, R{\'e}mi and Funtowicz, Morgan and others},
  booktitle={Proceedings of the 2020 conference on empirical methods in natural language processing: system demonstrations},
  pages={38--45},
  year={2020}
}

@article{fisher1936use,
  title={The use of multiple measurements in taxonomic problems},
  author={Fisher, Ronald A},
  journal={Annals of eugenics},
  volume={7},
  number={2},
  pages={179--188},
  year={1936},
  publisher={Wiley Online Library}
}

@article{alain2016understanding,
  title={Understanding intermediate layers using linear classifier probes},
  author={Alain, Guillaume and Bengio, Yoshua},
  journal={arXiv preprint arXiv:1610.01644},
  year={2016}
}

@article{belinkov2022probing,
  title={Probing classifiers: Promises, shortcomings, and advances},
  author={Belinkov, Yonatan},
  journal={Computational Linguistics},
  volume={48},
  number={1},
  pages={207--219},
  year={2022}
}

@article{williams2010nonnegative,
  title={Nonnegative decomposition of multivariate information},
  author={Williams, Paul L and Beer, Randall D},
  journal={arXiv preprint arXiv:1004.2515},
  year={2010}
}

@article{bertschinger2014quantifying,
  title={Quantifying unique information},
  author={Bertschinger, Nils and Rauh, Johannes and Olbrich, Eckehard and Jost, J{\"u}rgen and Ay, Nihat},
  journal={Entropy},
  volume={16},
  number={4},
  pages={2161--2183},
  year={2014},
  publisher={Molecular Diversity Preservation International (MDPI)}
}

@inproceedings{xin2020deebert,
  title={DeeBERT: Dynamic early exiting for accelerating BERT inference},
  author={Xin, Ji and Tang, Raphael and Lee, Jaejun and Yu, Yaoliang and Lin, Jimmy},
  booktitle={Proceedings of the 58th annual meeting of the association for computational linguistics},
  pages={2246--2251},
  year={2020}
}

@inproceedings{fanreducing,
  title={Reducing Transformer Depth on Demand with Structured Dropout},
  author={Fan, Angela and Grave, Edouard and Joulin, Armand},
  booktitle={International Conference on Learning Representations}
}

@article{tibshirani1993introduction,
  title={An introduction to the bootstrap},
  author={Tibshirani, Robert J and Efron, Bradley},
  journal={Monographs on statistics and applied probability},
  volume={57},
  number={1},
  pages={1--436},
  year={1993}
}

@article{yang2025qwen3,
  title={Qwen3 technical report},
  author={Yang, An and Li, Anfeng and Yang, Baosong and Zhang, Beichen and Hui, Binyuan and Zheng, Bo and Yu, Bowen and Gao, Chang and Huang, Chengen and Lv, Chenxu and others},
  journal={arXiv preprint arXiv:2505.09388},
  year={2025}
}

@article{grattafiori2024llama,
  title={The llama 3 herd of models},
  author={Grattafiori, Aaron and Dubey, Abhimanyu and Jauhri, Abhinav and Pandey, Abhinav and Kadian, Abhishek and Al-Dahle, Ahmad and Letman, Aiesha and Mathur, Akhil and Schelten, Alan and Vaughan, Alex and others},
  journal={arXiv preprint arXiv:2407.21783},
  year={2024}
}

@misc{jiang2023mistral7b,
      title={Mistral 7B}, 
      author={Albert Q. Jiang and Alexandre Sablayrolles and Arthur Mensch and Chris Bamford and Devendra Singh Chaplot and Diego de las Casas and Florian Bressand and Gianna Lengyel and Guillaume Lample and Lucile Saulnier and Lélio Renard Lavaud and Marie-Anne Lachaux and Pierre Stock and Teven Le Scao and Thibaut Lavril and Thomas Wang and Timothée Lacroix and William El Sayed},
      year={2023},
      eprint={2310.06825},
      archivePrefix={arXiv},
      primaryClass={cs.CL},
      url={https://arxiv.org/abs/2310.06825}, 
}

@article{kingma2014adam,
  title={Adam: A method for stochastic optimization},
  author={Kingma, Diederik P and Ba, Jimmy},
  journal={arXiv preprint arXiv:1412.6980},
  year={2014}
}

@inproceedings{guo2017calibration,
  title={On calibration of modern neural networks},
  author={Guo, Chuan and Pleiss, Geoff and Sun, Yu and Weinberger, Kilian Q},
  booktitle={International conference on machine learning},
  pages={1321--1330},
  year={2017},
  organization={PMLR}
}

@inproceedings{wang-etal-2024-factuality,
    title = "Factuality of Large Language Models: A Survey",
    author = "Wang, Yuxia  and
      Wang, Minghan  and
      Manzoor, Muhammad Arslan  and
      Liu, Fei  and
      Georgiev, Georgi Nenkov  and
      Das, Rocktim Jyoti  and
      Nakov, Preslav",
    editor = "Al-Onaizan, Yaser  and
      Bansal, Mohit  and
      Chen, Yun-Nung",
    booktitle = "Proceedings of the 2024 Conference on Empirical Methods in Natural Language Processing",
    month = nov,
    year = "2024",
    address = "Miami, Florida, USA",
    publisher = "Association for Computational Linguistics",
    pages = "19519--19529",
}

@inproceedings{azaria2023internal,
  title={The internal state of an LLM knows when it’s lying},
  author={Azaria, Amos and Mitchell, Tom},
  booktitle={Findings of the Association for Computational Linguistics: EMNLP 2023},
  pages={967--976},
  year={2023}
}

@inproceedings{burnsdiscovering,
  title={Discovering Latent Knowledge in Language Models Without Supervision},
  author={Burns, Collin and Ye, Haotian and Klein, Dan and Steinhardt, Jacob},
  booktitle={The Eleventh International Conference on Learning Representations}
}

@article{kadavath2022language,
  title={Language models (mostly) know what they know},
  author={Kadavath, Saurav and Conerly, Tom and Askell, Amanda and Henighan, Tom and Drain, Dawn and Perez, Ethan and Schiefer, Nicholas and Hatfield-Dodds, Zac and DasSarma, Nova and Tran-Johnson, Eli and others},
  journal={arXiv preprint arXiv:2207.05221},
  year={2022}
}

@article{li2023inference,
  title={Inference-time intervention: Eliciting truthful answers from a language model},
  author={Li, Kenneth and Patel, Oam and Vi{\'e}gas, Fernanda and Pfister, Hanspeter and Wattenberg, Martin},
  journal={Advances in Neural Information Processing Systems},
  volume={36},
  pages={41451--41530},
  year={2023}
}

@inproceedings{tenney2019bert,
  title={BERT rediscovers the classical NLP pipeline},
  author={Tenney, Ian and Das, Dipanjan and Pavlick, Ellie},
  booktitle={Proceedings of the 57th annual meeting of the association for computational linguistics},
  pages={4593--4601},
  year={2019}
}

@inproceedings{hewitt2019structural,
  title={A structural probe for finding syntax in word representations},
  author={Hewitt, John and Manning, Christopher D},
  booktitle={Proceedings of the 2019 Conference of the North American Chapter of the Association for Computational Linguistics: Human Language Technologies, Volume 1 (Long and Short Papers)},
  pages={4129--4138},
  year={2019}
}

@inproceedings{liu2020fastbert,
  title={Fastbert: a self-distilling bert with adaptive inference time},
  author={Liu, Weijie and Zhou, Peng and Wang, Zhiruo and Zhao, Zhe and Deng, Haotang and Ju, Qi},
  booktitle={Proceedings of the 58th annual meeting of the association for computational linguistics},
  pages={6035--6044},
  year={2020}
}

@inproceedings{xin2021berxit,
  title={BERxiT: Early exiting for BERT with better fine-tuning and extension to regression},
  author={Xin, Ji and Tang, Raphael and Yu, Yaoliang and Lin, Jimmy},
  booktitle={Proceedings of the 16th conference of the European chapter of the association for computational linguistics: Main Volume},
  pages={91--104},
  year={2021}
}

@article{zhou2020bert,
  title={Bert loses patience: Fast and robust inference with early exit},
  author={Zhou, Wangchunshu and Xu, Canwen and Ge, Tao and McAuley, Julian and Xu, Ke and Wei, Furu},
  journal={Advances in Neural Information Processing Systems},
  volume={33},
  pages={18330--18341},
  year={2020}
}

@inproceedings{elhoushi2024layerskip,
  title={Layerskip: Enabling early exit inference and self-speculative decoding},
  author={Elhoushi, Mostafa and Shrivastava, Akshat and Liskovich, Diana and Hosmer, Basil and Wasti, Bram and Lai, Liangzhen and Mahmoud, Anas and Acun, Bilge and Agarwal, Saurabh and Roman, Ahmed and others},
  booktitle={Proceedings of the 62nd Annual Meeting of the Association for Computational Linguistics (Volume 1: Long Papers)},
  pages={12622--12642},
  year={2024}
}

@inproceedings{chen2024ee,
  title={EE-LLM: Large-Scale Training and Inference of Early-Exit Large Language Models with 3D Parallelism},
  author={Chen, Yanxi and Pan, Xuchen and Li, Yaliang and Ding, Bolin and Zhou, Jingren},
  booktitle={International Conference on Machine Learning},
  pages={7163--7189},
  year={2024},
  organization={PMLR}
}

@inproceedings{wan2026fano,
  title={A fano-style accuracy upper bound for llm single-pass reasoning in multi-hop qa},
  author={Wan, Kaiyang and Gao, Lang and Mu, Honglin and Nakov, Preslav and Wang, Yuxia and Chen, Xiuying},
  booktitle={International Conference on Learning Representations},
  volume={2026},
  pages={52600--52621},
  year={2026}
}

\appendix

\section{Limitations and Broader Impact}
\label{app:limitations}

\paragraph{Limitations.}
This study focuses on binary hallucination detection and evaluates a representative set of decoder-only LLM backbones. 
Future work could extend the analysis to more fine-grained factuality labels, additional architectures, and other verification tasks. 
Our early-exit rule uses a static cutoff shared by all inputs; combining InfoWindow-guided truncation with input-adaptive exiting may further improve the efficiency--accuracy trade-off.
We also restrict \textsc{s2s} to a single verdict token as a matched control, and the \textsc{cls} advantage we measure is large on hallucination detection and small on QNLI and SST-2 (Section~\ref{sec:conclusion}).

\paragraph{Broader impact.}
Hallucination detection is an important component of trustworthy LLM systems, especially in retrieval-augmented generation and agentic workflows where model outputs are often passed to downstream users or tools. 
By studying how fine-tuning objectives shape detector representations, this work provides evidence for designing stronger and more efficient verification models. 
The early-exit results suggest that hallucination detectors can be made cheaper without sacrificing accuracy, which may make verification more accessible in resource-constrained deployments. 
As with any automated verifier, these detectors should complement rather than replace task-specific evaluation and human oversight in high-stakes applications.

\section{Existing assets and licenses}
\label{app:license}
All datasets, model backbones, and software packages used in this work are existing publicly available research assets. We cite the original sources for these assets and use them in accordance with the access conditions and usage terms specified by their original providers or hosting platforms.

\section{Experimental Setup Details}
\label{app_setup}

This section provides the full experimental protocol underlying the compact description in \S\ref{sec:setup}. We emphasize the details that affect the validity of the objective comparison: model choice, data provenance, split construction, fine-tuning parity, and the separation between analysis and benchmark evaluation.

\paragraph{Models.}
Our main representation analysis uses Qwen3-4B and Qwen3-8B~\citep{yang2025qwen3}, two decoder-only transformers from the same family. Both models have 36 transformer blocks but differ in width, which lets us compare per-layer measurements across scales without changing the layer index. To test whether the main \textsc{cls} versus \textsc{s2s} pattern depends on this model family, we also replicate the headline comparison on Llama-3.2-3B~\citep{grattafiori2024llama} and Mistral-7B-v0.3~\citep{jiang2023mistral7b}. All backbones run in bfloat16 with SDPA attention and a 2048-token training context window.

\paragraph{Data sources.}
We build the fine-tuning set from three binary hallucination-detection corpora: RAGTruth~\citep{niu2024ragtruth}, ClaimVerify~\citep{liu2023evaluating}, and FactCheck-GPT~\citep{wang2024factcheck}, for a total of $22{,}827$ raw training rows. For out-of-distribution evaluation, we use three held-out corpora that are never used during fine-tuning: AggreFact~\citep{tang2023understanding} for summarization factuality, FavaBench~\citep{mishrafine} for fine-grained hallucination types, and FaithBench~\citep{bao2025faithbench} for summarization. All six corpora provide paired training and testing files, except AggreFact, FaithBench, and FavaBench, which contribute test files only. Table~\ref{tab:data-stats} reports the per-file row counts and class balance after preprocessing.

\paragraph{Test split construction: deduplication followed by stratified 50/50 splitting.}
For every test file, we first deduplicate examples by the $(\text{premise}, \text{hypothesis})$ fingerprint. This step removes literal duplicate pairs before any split is created; in particular, we found $7$ duplicate pairs in RAGTruth and $51$ in AggreFact. Without this deduplication, identical content could appear in both halves even when the example indices are disjoint. We then split each deduplicated file into a stratified $50{/}50$ \emph{probe-half} and \emph{eval-half} using a fixed seed, \texttt{SPLIT\_SEED=42}. Since both halves are generated from the same deduplicated seeded permutation, they are disjoint at the content level.

The two halves serve different purposes throughout the paper. The probe-half is used only for the representation analyses in \S\ref{sec:objective-comparison} and \S\ref{sec:plateau}, including probe fitting, $\mathcal{V}$-information estimation, Fisher scores, and PID. The eval-half is used only for benchmark reporting in Table~\ref{tab:main} and Table~\ref{tab:early-exit}. Thus, no probe-half example is used to fine-tune a model or report benchmark performance, and no eval-half example is used to fit an analysis probe.

\paragraph{Training/validation split for fine-tuning.}
Fine-tuning uses only the dedicated training files from RAGTruth, ClaimVerify, and FactCheck-GPT. We concatenate these files, shuffle them with seed $42$, and hold out $5\%$ as an in-training validation slice for logging. This yields approximately $21{,}686$ training rows and $1{,}141$ validation rows. The validation slice is not used for model selection: all reported results use final-step checkpoints. This choice avoids introducing checkpoint-selection differences between the two fine-tuning paradigms.

\paragraph{Length truncation.}
Both paradigms tokenize examples with the model's native chat template and use left padding. The training-time context window is capped at $2{,}048$ tokens using HuggingFace truncation, with \texttt{truncation=True} and \texttt{max\_length=2048}. Hidden-state extraction for the representation analyses uses a $4{,}096$-token window. Under the Qwen3 tokenizer, more than $99\%$ of training examples fit within $2{,}048$ tokens. For the rare overflowing examples, tokenizer truncation removes tokens from the left, preserving the trailing \texttt{Answer:} cue and the readout position used by the \textsc{cls} head.

\paragraph{Per-dataset statistics.}
Table~\ref{tab:data-stats} summarizes the row counts after deduplication and the per-half sizes used in the experiments. The in-distribution probe pool contains $2{,}870$ rows and the OOD probe pool contains $6{,}506$ rows; the corresponding eval-half sizes differ by at most one row because of odd corpus sizes. Class balance varies substantially across corpora, with $p(Y{=}1)$ ranging from $0.20$ on FavaBench to $0.74$ on ClaimVerify. We therefore report macro-$F_1$ rather than accuracy throughout the paper.

\begin{table}[h]
\centering
\small
\setlength{\tabcolsep}{4.5pt}
\caption{\textbf{Per-corpus row counts.} Training corpora (top) feed fine-tuning under a $95{/}5$ train/validation split. Test corpora (bottom) are deduplicated by $(\text{premise},\text{hypothesis})$ fingerprint, then split stratified $50{/}50$ into a probe-half for static and dynamic analyses and an eval-half for benchmark evaluation. $p(Y{=}1)$ is the positive-class rate measured on the deduplicated corpus.}
\label{tab:data-stats}
\resizebox{.98\linewidth}{!}{%
\begin{tabular}{lllrrrrr}
\toprule
\textbf{Corpus} & \textbf{Domain} & \textbf{Role} & \textbf{Raw} & \textbf{Deduped} & \textbf{Probe-half} & \textbf{Eval-half} & $p(Y{=}1)$ \\
\midrule
\multicolumn{8}{l}{\textit{Training corpora ($95\%$ train / $5\%$ in-training validation)}} \\
RAGTruth\textsubscript{train}      & general    & train & 14{,}949 & --- & --- & --- & 0.554 \\
ClaimVerify\textsubscript{train}   & general    & train &  6{,}341 & --- & --- & --- & 0.720 \\
FactCheck-GPT\textsubscript{train} & general    & train &  1{,}537 & --- & --- & --- & 0.173 \\
\textit{Total}                     &            &       & 22{,}827 & --- & --- & --- & 0.487 \\
\midrule
\multicolumn{8}{l}{\textit{In-distribution test corpora (probe-half $\to$ analysis; eval-half $\to$ benchmark)}} \\
RAGTruth\textsubscript{test}       & general    & probe + eval &  2{,}681 &  2{,}674 & 1{,}337 & 1{,}337 & 0.649 \\
ClaimVerify\textsubscript{test}    & general    & probe + eval &  1{,}517 &  1{,}503 &    751 &    752 & 0.741 \\
FactCheck-GPT\textsubscript{test}  & general    & probe + eval &  1{,}566 &  1{,}565 &    782 &    783 & 0.240 \\
\midrule
\multicolumn{8}{l}{\textit{OOD test corpora (probe-half $\to$ analysis; eval-half $\to$ benchmark)}} \\
AggreFact\textsubscript{test}      & summarization & probe + eval &  9{,}421 &  9{,}370 & 4{,}685 & 4{,}685 & 0.592 \\
FaithBench\textsubscript{test}     & summarization & probe + eval &     459 &     459 &    229 &    230 & 0.231 \\
FavaBench\textsubscript{test}      & fine-grained  & probe + eval &  3{,}185 &  3{,}185 & 1{,}592 & 1{,}593 & 0.200 \\
\midrule
\textit{Total (test)}              &              &              & 18{,}829 & 18{,}756 & 9{,}376 & 9{,}380 & --- \\
\bottomrule
\end{tabular}}
\end{table}

\paragraph{Fine-tuning recipe shared by both paradigms.}
Both \textsc{cls} and \textsc{s2s} use the same parameter-efficient fine-tuning recipe. We attach LoRA adapters~\citep{hu2022lora} with rank $r=32$, scaling $\alpha=64$, and dropout $0.05$ to all seven projection matrices in every transformer block: \texttt{q\_proj}, \texttt{k\_proj}, \texttt{v\_proj}, \texttt{o\_proj}, \texttt{gate\_proj}, \texttt{up\_proj}, and \texttt{down\_proj}. The base model weights remain frozen. AdamW optimizes the adapters for $3$ epochs with learning rate $10^{-4}$, weight decay $0.01$, and a linear warmup over the first $3\%$ of steps. Each run uses an effective batch size of $8$ sequences, groups sequences by length to reduce padding, and shuffles training examples with seed $42$.

Each run uses a single H200 GPU. For a fixed backbone, the two paradigms see the same examples in the same order for the same number of optimization steps. The only intended difference is the readout and loss defined in \S\ref{sec:prelim}: \textsc{cls} trains a binary classifier head on the final hidden state, whereas \textsc{s2s} trains the pretrained language-modeling head to generate the verdict token. This matched setup isolates the objective and readout choice as the experimental contrast.

\paragraph{Heads and decoding.}
For \textsc{cls}, we replace HuggingFace's default score layer with a bias-free linear head $w \in \mathbb{R}^{D \times 2}$ applied in fp32 to the hidden state of the last input token at layer $L$. Left padding ensures that this readout position always corresponds to the final column. Predictions use
\[
\hat{Y} = \argmax_y (w^\top H_L)_y .
\]
For \textsc{s2s}, we keep the pretrained language-modeling head and train with \texttt{trl}'s \texttt{SFTTrainer}. Prompt tokens are masked to $-100$, so the loss applies only to the \texttt{Yes}/\texttt{No} verdict. At test time, we greedily decode and parse the first verdict token after the \texttt{Answer:} cue. Both paradigms receive the same chat-formatted NLI prompt, so their inputs are byte-identical up to the readout point.

\paragraph{Prompt template.}
Each $(\mathrm{premise}, \mathrm{hypothesis})$ pair is rendered as a single user-turn chat message with no system prompt:
\begin{verbatim}
Can the hypothesis be inferred from the premise? Answer using Yes or No only.

Premise: {premise}
Hypothesis: {hypothesis}
Answer:
\end{verbatim}
The rendered string is then passed through each model's native chat template. The \textsc{cls} and \textsc{s2s} variants share this rendering byte-for-byte; they differ only after the \texttt{Answer:} cue, where \textsc{cls} reads the hidden state at the trailing token and \textsc{s2s} predicts the next \texttt{Yes}/\texttt{No} token.

\paragraph{Reporting.}
We report macro-$F_1$ on each of the six test corpora and their unweighted average. Every backbone--paradigm cell in Table~\ref{tab:main} and every exit in Table~\ref{tab:early-exit} except L15 is fine-tuned with three seeds (42, 43, 44) under an otherwise identical recipe and evaluated at its final-step checkpoint; we report the mean and population standard deviation and compute gaps and early-exit deltas per seed before averaging. The static and dynamic analyses refit their probes with three independent seeds and report paired bootstrap $95\%$ confidence intervals. As a result, the benchmark tables evaluate held-out examples that are never used for probe fitting, while the layer-wise analysis reports uncertainty for the statistics that support the mechanistic claims.

\section{Extended Methodology: Analysis Tools}
\label{app:tools}

This section gives the formal definitions and implementation details for the representation-analysis tools used in the paper. The common design principle is to keep the data split, readout position, probe protocol, and calibration procedure fixed whenever we compare layers, checkpoints, model scales, or fine-tuning objectives.

\subsection{Visualization Details}
\label{app:visual}

This appendix describes the implementation of Fig.~\ref{fig:geometry}. For each model scale, we construct a two-dimensional coordinate system from the corresponding \textsc{cls}-fine-tuned model and then reuse the same axes for the pretrained \textsc{base}, \textsc{s2s}-fine-tuned, and \textsc{cls}-fine-tuned variants. This shared coordinate system makes the panels comparable: visual differences reflect changes in the hidden states, not changes in the projection chosen for each paradigm.

\paragraph{Projection axes.}
For each scale, we load the trained \textsc{cls} score head $W_{\textsc{cls}}\in\mathbb{R}^{2\times D}$ and define the discriminative axis as
\begin{equation}
    w^{\mathrm{disc}}
    =
    \frac{W_{\textsc{cls}}[1]-W_{\textsc{cls}}[0]}
         {\|W_{\textsc{cls}}[1]-W_{\textsc{cls}}[0]\|_2}.
\end{equation}
This axis is the unit normal of the linear decision hyperplane learned by the \textsc{cls} head. We then extract last-layer hidden states from the same \textsc{cls}-fine-tuned model, remove the component along $w^{\mathrm{disc}}$, and compute the leading principal direction of the residual:
\begin{equation}
    \tilde H
    =
    H
    -
    (H w^{\mathrm{disc}}) {w^{\mathrm{disc}}}^{\top},
    \qquad
    w^\perp
    =
    \mathrm{PC}_1(\tilde H).
\end{equation}
We normalize $w^\perp$ to unit length. In implementation, $w^\perp$ is the top right singular vector of the centered residual matrix. Thus, the horizontal axis is aligned with the trained \textsc{cls} decision direction, while the vertical axis captures the largest remaining direction of variation orthogonal to it.

\paragraph{Shared coordinates across paradigms.}
For a fixed scale, every example from each paradigm is projected onto the same two axes:
\begin{equation}
    z_1(x)=H_L(x)^\top w^{\mathrm{disc}},
    \qquad
    z_2(x)=H_L(x)^\top w^\perp .
\end{equation}
The axes are shared across \textsc{base}, \textsc{s2s}, and \textsc{cls}. This prevents each paradigm from selecting its own favorable projection and makes the three panels a direct geometric comparison. The 4B and 8B models use separately constructed axes because their hidden dimensions differ.

\paragraph{Density rendering.}
For each class and panel, we estimate a two-dimensional Gaussian kernel density over $(z_1,z_2)$ with bandwidth factor $0.30$. The density is evaluated on a $140\times140$ grid and each class density is normalized by its own maximum before contour plotting. This normalization makes the figure suitable for comparing class locations, overlap, and shape, but not for comparing absolute density magnitudes across classes or panels.

\paragraph{Axis limits and figure layout.}
Within each scale, we compute axis limits from the union of projected points across the three variants and apply the same limits to all three panels. Thus, differences among \textsc{base}, \textsc{s2s}, and \textsc{cls} reflect geometry rather than rescaling. The main figure arranges the six panels as
\[
\text{4B: Base, S2S, CLS}
\quad
\text{and}
\quad
\text{8B: Base, S2S, CLS}.
\]
All panels use square aspect ratio, shared class colors, and a shared legend.

\subsection{Direction-aligned Fisher score}
\label{app:geometry}

\paragraph{Goal.}
For each layer $\ell$ of each model, we compute a scalar measure of how separable the two classes are along the direction used by the trained \textsc{cls} classifier. This produces a geometry-aligned diagnostic that is directly tied to Fig.~\ref{fig:geometry}: the visualization shows class geometry in the \textsc{cls} decision coordinate system, and the Fisher score quantifies class separation along that same decision direction.

\paragraph{Inputs.}
We use the held-out probe-half defined in Appendix~\ref{app_setup}, which is disjoint from both the fine-tuning data and the benchmark eval-half. For every model
\[
m \in \{\textsc{base}, \textsc{s2s}, \textsc{cls}\} \times \{\text{4B}, \text{8B}\},
\]
we cache the per-layer block-output hidden states at the final answer position,
\[
H^{(m)} \in \mathbb{R}^{N \times (L+1) \times D},
\]
where $N$ is the number of probe examples, $L=36$ for both Qwen3-4B and Qwen3-8B, and $D$ is the hidden width ($D_{\text{4B}}=2560$, $D_{\text{8B}}=4096$). We discard the embedding output at index $0$ and index transformer blocks as $\ell=1,\dots,L$, matching the layer convention used by the linear probes in Appendix~\ref{app:vinfo}. Labels are $y_i\in\{0,1\}$, with $0=\textsc{faithful}$ and $1=\textsc{hallucinated}$. All arithmetic for this diagnostic is performed in \texttt{float64}.

\paragraph{Step 1: extract the decision normal.}
For each scale, we load the final-step \textsc{cls} adapter checkpoint and read its score head
\[
W_{\mathrm{cls}} \in \mathbb{R}^{2 \times D},
\qquad
W_{\mathrm{cls}} =
\begin{bmatrix}
w_0^\top \\
w_1^\top
\end{bmatrix}.
\]
The logit margin between the hallucinated and faithful classes is the linear functional $h\mapsto (w_1-w_0)^\top h$, so the separating hyperplane is normal to $w_1-w_0$. We define the unit decision normal as
\begin{equation}
w^{\mathrm{disc}}
=
\frac{w_1-w_0}{\|w_1-w_0\|_2}
\in\mathbb{R}^{D}.
\label{eq:wdisc-def}
\end{equation}
This vector is computed once per scale from the trained \textsc{cls} head and then shared across \textsc{base}, \textsc{s2s}, and \textsc{cls}. This is the only learned direction used by the metric.

\paragraph{Step 2: project hidden states.}
For each model and layer, we compute scalar projections onto the shared decision direction:
\begin{equation}
z^{(m)}_{i,\ell}
=
\langle w^{\mathrm{disc}}, H^{(m)}_{i,\ell,\cdot}\rangle,
\qquad
i=1,\dots,N.
\label{eq:proj}
\end{equation}
Because $\|w^{\mathrm{disc}}\|_2=1$, the projection has the scale of a single hidden-coordinate direction. Differences along this axis correspond to changes in the \textsc{cls} logit margin up to the multiplicative factor $\|w_1-w_0\|_2$.

\paragraph{Step 3: compute class statistics in one dimension.}
Let $\mathcal{I}_+=\{i:y_i=1\}$ and $\mathcal{I}_-=\{i:y_i=0\}$. For each model and layer, we compute the projected class means
\[
\mu^{+}_{m,\ell}
=
\frac{1}{|\mathcal{I}_+|}
\sum_{i\in\mathcal{I}_+} z^{(m)}_{i,\ell},
\qquad
\mu^{-}_{m,\ell}
=
\frac{1}{|\mathcal{I}_-|}
\sum_{i\in\mathcal{I}_-} z^{(m)}_{i,\ell},
\]
the signed mean gap
\[
\Delta_{m,\ell}
=
\mu^{+}_{m,\ell}-\mu^{-}_{m,\ell},
\]
and the unbiased projected variances
\[
\sigma^{2,+}_{m,\ell}
=
\frac{1}{|\mathcal{I}_+|-1}
\sum_{i\in\mathcal{I}_+}
\bigl(z^{(m)}_{i,\ell}-\mu^{+}_{m,\ell}\bigr)^2,
\qquad
\sigma^{2,-}_{m,\ell}
=
\frac{1}{|\mathcal{I}_-|-1}
\sum_{i\in\mathcal{I}_-}
\bigl(z^{(m)}_{i,\ell}-\mu^{-}_{m,\ell}\bigr)^2.
\]

\paragraph{Step 4: direction-aligned Fisher score.}
The per-layer score is
\begin{equation}
F_w^{(m)}(\ell)
=
\frac{\Delta_{m,\ell}^2}
     {\sigma^{2,+}_{m,\ell}+\sigma^{2,-}_{m,\ell}}
=
\frac{
\bigl(w^{\mathrm{disc}\,\top}(\mu^{+}_{m,\ell}-\mu^{-}_{m,\ell})\bigr)^2
}{
w^{\mathrm{disc}\,\top}(\Sigma^{+}_{m,\ell}+\Sigma^{-}_{m,\ell})w^{\mathrm{disc}}
}.
\label{eq:fisher-aligned}
\end{equation}
The second expression writes the same quantity in terms of the original $D$-dimensional class means and covariances. In implementation, we use the 1D form in Eq.~\ref{eq:fisher-aligned}, which requires only an $O(NL)$ projection pass and an $O(NL)$ statistics pass. We do not normalize hidden states, smooth across layers, or shrink covariance estimates. If the denominator is exactly zero, which did not occur in our runs, we set the score to zero.

\paragraph{Step 5: best layer.}
For each model, we report the peak layer and peak score:
\[
\ell^{\star}_{m}
=
\operatorname*{arg\,max}_{\ell\in\{1,\dots,L\}} F_w^{(m)}(\ell),
\qquad
F_w^{\star,m}
=
F_w^{(m)}(\ell^{\star}_{m}).
\]
This identifies the layer at which class separation along the \textsc{cls} decision direction is largest.

\paragraph{Why use direction-aligned rather than trace-form Fisher.}
A conventional trace-form Fisher score would be
\[
F_{\mathrm{tr}}^{(m)}(\ell)
=
\frac{\|\mu^{+}_{m,\ell}-\mu^{-}_{m,\ell}\|_2^2}
     {\mathrm{tr}(\Sigma^{+}_{m,\ell}+\Sigma^{-}_{m,\ell})}.
\]
This form aggregates over all hidden dimensions. That aggregation creates two problems for our purpose. First, if \textsc{cls} fine-tuning concentrates class information into a low-dimensional decision subspace while leaving high-variance task-irrelevant directions intact, the trace denominator dilutes the very signal we want to measure. Second, the trace denominator scales with hidden width under roughly isotropic variance, making raw trace-form scores difficult to compare across Qwen3-4B and Qwen3-8B without an additional normalization convention.

The direction-aligned score avoids both issues by measuring separation after projecting to the same 1D decision coordinate. It should therefore be read as a \emph{\textsc{cls}-discoverable} separability diagnostic: it measures separation along the direction that the trained \textsc{cls} head actually uses. This is the appropriate convention for Sec.~\ref{sec:geometry}, whose claim concerns the geometry induced by \textsc{cls} fine-tuning and read out by the \textsc{cls} classifier. Complementary diagnostics in Appendix~\ref{app:vinfo} do not restrict the readout to this single direction.

\paragraph{Cross-paradigm convention.}
For \textsc{base} and \textsc{s2s}, $w^{\mathrm{disc}}$ is borrowed from the same scale's trained \textsc{cls} head rather than relearned. This is essential for interpretation: $F_w^{\textsc{base}}(\ell)$, $F_w^{\textsc{s2s}}(\ell)$, and $F_w^{\textsc{cls}}(\ell)$ are measurements along a common axis in the same representation space. Differences among the curves therefore reflect differences in the hidden states, not differences in the direction chosen by the metric. A paradigm-specific LDA direction, by contrast, would give each paradigm its own most favorable axis and would no longer support a direct geometric comparison.

\paragraph{Numerical and reproducibility notes.}
All projections, means, and variances are computed on CPU in \texttt{float64}. Sample variances use Bessel's correction. The \texttt{score.linear.weight} parameter has shape $(2,D)$, with row $0$ corresponding to the \textsc{faithful} logit and row $1$ corresponding to the \textsc{hallucinated} logit. Therefore $w_1-w_0$ points from \textsc{faithful} toward \textsc{hallucinated}, and positive $\Delta_{m,\ell}$ indicates separation in the same orientation as the trained head. We save the arrays $\{F_w^{(m)}(\ell)\}_{\ell}$, $\{\Delta_{m,\ell}\}_{\ell}$, and $\{\sigma^{2,+}_{m,\ell}+\sigma^{2,-}_{m,\ell}\}_{\ell}$ as \texttt{.npy} files, together with a JSON summary recording the source \textsc{cls} checkpoint, $\ell^\star_m$, $F_w^{\star,m}$, and the metric definition.

\subsection{Layer-wise linear probe and $\mathcal{V}$-information}
\label{app:vinfo}

\paragraph{Goal.}
For each layer $\ell$ of each model, we estimate how much of the binary label $Y\in\{0,1\}$ can be recovered by a linear readout from the last-token hidden state $H_\ell\in\mathbb{R}^D$. We instantiate this quantity with $\mathcal{V}$-information~\citep{xutheory} using a temperature-calibrated linear probe family. The resulting curve $\{I_{\mathcal{V}}(\ell)\}_{\ell=1}^{L}$ is the layer-wise decodability quantity plotted in Fig.~\ref{fig:vinfo} and recomputed over checkpoints in Sec.~\ref{sec:dynamic}.

\paragraph{$\mathcal{V}$-information identity.}
$\mathcal{V}$-information measures the label information accessible to a restricted predictor family $\mathcal{V}\subseteq\{f:\mathbb{R}^D\to\Delta^{|\mathcal{Y}|}\}$:
\begin{equation}
I_{\mathcal{V}}(H_\ell \to Y)
=
H_{\mathcal{V}}(Y)
-
H_{\mathcal{V}}(Y\mid H_\ell),
\label{eq:vinfo-def}
\end{equation}
where
\[
H_{\mathcal{V}}(Y)
=
\inf_{f\in\mathcal{V}}
\mathbb{E}[-\log f(\varnothing)(Y)]
\]
and
\[
H_{\mathcal{V}}(Y\mid H_\ell)
=
\inf_{f\in\mathcal{V}}
\mathbb{E}[-\log f(H_\ell)(Y)].
\]
We take $\mathcal{V}$ to be the family of linear probes with bias,
\[
\mathcal{V}
=
\left\{
h\mapsto \mathrm{softmax}(Wh+b)
:
W\in\mathbb{R}^{2\times D},\ b\in\mathbb{R}^{2}
\right\}.
\]
The bias term ensures that $\mathcal{V}$ contains constant predictors. Thus the unconditional term is the empirical label entropy, $H_{\mathcal{V}}(Y)=H(Y)$, and
\begin{equation}
I_{\mathcal{V}}(H_\ell \to Y)
=
H(Y)
-
\widehat{\mathrm{CE}}_{\mathcal{V}}(\ell),
\label{eq:iv-cefrom}
\end{equation}
where $\widehat{\mathrm{CE}}_{\mathcal{V}}(\ell)$ is the calibrated test cross-entropy of the best linear probe at layer $\ell$. This quantity is bounded above by $H(Y)\le \log 2$ nats for binary labels. In finite samples, raw probe cross-entropy can exceed $H(Y)$ because of overconfidence or optimization noise, so we report the clipped estimate in Eq.~\ref{eq:iv-final} and store the unclipped value as a diagnostic.

\paragraph{Inputs.}
We use the held-out probe-half pooled across the six test corpora, comprising approximately $9{,}376$ examples. This pool is disjoint from both the fine-tuning training corpus and the benchmark eval-half described in Appendix~\ref{app_setup}. We optionally cap the pool at $N=10{,}000$ using stratified sampling with \texttt{seed=42}; in our experiments the cap is non-binding. For every model $m$ in the analysis registry, we cache per-layer last-token hidden states at the final answer position:
\[
H^{(m)} \in \mathbb{R}^{N \times (L+1) \times D},
\qquad
L=36,\quad
D_{\text{4B}}=2560,\quad
D_{\text{8B}}=4096.
\]
We discard the embedding output at index $0$ and index layers as $\ell=1,\dots,L$. We also cache the post-attention residual stream after each block's self-attention and before its MLP, yielding a parallel tensor of shape $(N,L,D)$. Unless otherwise stated, all probe results in the paper use block-output hidden states.

\paragraph{Stratified train/validation/test split.}
A single stratified split with \texttt{seed=42} partitions the probe pool into 70\% train, 15\% validation, and 15\% test. The split indices are saved and reused across probe families, layers, model variants, and checkpoints. This makes $I_{\mathcal{V}}$ values comparable across layers, paradigms, scales, and training steps because all evaluations use the same held-out examples. Stratification preserves the empirical class prior in each split, so $H(Y)$ is a fixed constant for the corresponding probe dataset.

\paragraph{Probe head and training procedure.}
At each layer $\ell$, the probe is
\[
f_\ell(h)=\mathrm{softmax}(W_\ell h+b_\ell),
\qquad
W_\ell\in\mathbb{R}^{2\times D},\quad b_\ell\in\mathbb{R}^{2}.
\]
The head is initialized with PyTorch's default \texttt{nn.Linear} initialization. We train with Adam~\citep{kingma2014adam}, full-batch gradient descent, cross-entropy loss, weight decay $0$, and a budget of $100$ epochs. The learning rate is selected per layer from
\[
\{10^{-4},\ 5\times10^{-4},\ 10^{-3}\}
\]
by validation macro-F$_1$. For each candidate learning rate, we record the epoch with the highest validation F$_1$; we then keep the head state from the candidate whose best validation F$_1$ is largest. Test cross-entropy and per-sample probabilities are computed from this selected head.

We do not apply representation normalization, layer-wise rescaling, or $\ell_2$ regularization. Thus, the probe family matches the unconstrained linear-with-bias family in Eq.~\ref{eq:vinfo-def}. We repeat the full procedure under three probe initialization seeds $\{42,123,456\}$, while keeping the train/validation/test split fixed. The shaded bands in Fig.~\ref{fig:vinfo} report mean and standard deviation over these seeds. For downstream PID and trajectory analyses, we use the median seed according to average F$_1$ over layers.

\paragraph{Temperature calibration.}
Upper-layer probes can become overconfident, producing cross-entropies that are poorly aligned with their classification accuracy. We therefore apply post-hoc temperature scaling~\citep{guo2017calibration}. For each saved layer head, we hold the probe weights fixed and fit a scalar temperature $T_\ell>0$ on validation logits:
\[
T_\ell^\star
=
\operatorname*{arg\,min}_{T>0}
\mathbb{E}_{(h,y)\in\mathrm{val}}
\left[
-\log
\mathrm{softmax}\!\left(
\frac{1}{T} f_\ell^{\mathrm{logit}}(h)
\right)_y
\right].
\]
We optimize this objective with LBFGS and strong-Wolfe line search, then evaluate calibrated cross-entropy on the test split:
\[
\widehat{\mathrm{CE}}_{\mathcal{V}}(\ell)
=
\mathbb{E}_{(h,y)\in\mathrm{test}}
\left[
-\log
\mathrm{softmax}\!\left(
\frac{1}{T_\ell^\star} f_\ell^{\mathrm{logit}}(h)
\right)_y
\right].
\]
The temperature is fit only on validation examples and then applied to test examples; test data never enters either head training or temperature fitting. We feed the calibrated CE into Eq.~\ref{eq:iv-cefrom} and report
\begin{equation}
\widehat{I}_{\mathcal{V}}(\ell)
=
\max\!\left(0,\ H(Y)-\widehat{\mathrm{CE}}_{\mathcal{V}}(\ell)\right).
\label{eq:iv-final}
\end{equation}
The unclipped value is saved separately to track overconfidence and optimization failure cases.

\paragraph{Trajectory variant.}
For Sec.~\ref{sec:dynamic} and Fig.~\ref{fig:iv-trajectory}, we re-extract last-token hidden states at every saved \textsc{cls} checkpoint $t\in\{0,150,300,\dots\}$, using the same examples, split, prompt formatting, and answer-position convention. We refit the per-layer linear probes from scratch at each checkpoint, recalibrate them, and assemble
\[
I_{\mathcal{V}}(\ell,t)\in\mathbb{R}^{L\times T}.
\]
Because the data split is fixed across checkpoints, changes along a row of this matrix reflect changes in the model representation rather than changes in the probe dataset.

\paragraph{Joint and SwiGLU variants used by PID.}
The PID analysis in Sec.~\ref{sec:pid} requires a joint probe
\[
\widehat{I}_{\mathcal{V}}(H_\ell\oplus H_p\to Y)
\]
from a family $\mathcal{V}_{\mathrm{joint}}$ that contains the marginal probe families used on $H_\ell$ and $H_p$. Otherwise, monotonicity comparisons between marginal and joint $\mathcal{V}$-information would be hard to interpret. Our fusion head uses an asymmetric SwiGLU--Linear architecture: a SwiGLU expansion on the candidate-layer branch and a Linear branch on the plateau-layer branch. To match that architecture, we train two additional marginal probe families per layer under the same 70/15/15 split, learning-rate grid, and calibration protocol:
\begin{itemize}
\item a SwiGLU probe
\[
\mathcal{V}_\ell
=
\left\{
h \mapsto
\mathrm{Linear}_{2}
\left(
\mathrm{Dropout}(\mathrm{SwiGLU}(h))
\right)
\right\},
\]
with hidden width matched to the fusion head, used for the candidate-layer marginal $H_\ell$;
\item the Linear-with-bias probe family above, used for the plateau-layer marginal $H_p$.
\end{itemize}
The PID inputs are the calibrated quantities
\[
I_{\mathcal{V}_\ell}(H_\ell;Y),
\qquad
I_{\mathcal{V}_p}(H_p;Y),
\qquad
I_{\mathcal{V}_{\mathrm{joint}}}(H_\ell,H_p;Y),
\]
all evaluated on the same test split. The nonlinear probe families use AdamW, weight decay $5\times10^{-3}$, and $150$ epochs to absorb their larger parameter count; all other choices match the linear-probe protocol.

\paragraph{Reported quantities.}
For each model and layer, the pipeline saves: raw and calibrated test CE, $\widehat{I}_{\mathcal{V}}(\ell)$ in nats, macro-F$_1$, accuracy, per-sample softmax probabilities, the selected learning rate, the selected epoch, the fitted temperature $T_\ell^\star$, and the probe head state. Multi-seed mean and standard deviation are stored alongside the median-seed snapshot used by downstream analyses.

\paragraph{Reproducibility notes.}
Probe training runs on a single GPU in \texttt{float32}; temperature fitting runs on CPU in \texttt{float64}. The 70/15/15 split is keyed by \texttt{seed=42} and is invariant to probe family, probe initialization seed, model checkpoint, and representation choice. This fixed split is what makes the layer-wise, cross-paradigm, cross-scale, and temporal comparisons numerically meaningful.

\subsection{Upper-layer saturation and plateau interchangeability}
\label{app:saturation}

This appendix details the plateau test reported in Sec.~\ref{sec:saturation}. We run this test on the \textsc{cls}-fine-tuned models because the resulting plateau boundary is the input to the early-exit analysis in Sec.~\ref{sec:early-exit}.

\paragraph{Goal.}
We test whether an intermediate layer of a \textsc{cls}-fine-tuned model carries as much linearly decodable task information as the final layer. Operationally, a layer is considered interchangeable with the final layer if a linear probe trained on that layer matches the final-layer probe in macro-F$_1$ within a paired bootstrap confidence interval. The smallest such layer, denoted $L^\star$, marks the start of the upper plateau: above this point, deeper layers no longer improve the linear readout on the same held-out examples.

\paragraph{Information-gain localization for the search interval.}
Before running the bootstrap test, we localize where fine-tuning adds linearly decodable label information relative to the pretrained backbone. For a \textsc{cls}-fine-tuned model $f$ and its pretrained backbone $f_0$, both evaluated with the calibrated linear-probe protocol of Appendix~\ref{app:vinfo}, we define
\begin{equation}
\Delta I_{\mathcal{V}}(\ell)
=
\max\!\left(
\widehat{I}_{\mathcal{V}}^{\,f}(\ell)
-
\widehat{I}_{\mathcal{V}}^{\,f_0}(\ell),
0
\right).
\label{eq:delta-iv}
\end{equation}
We mark a layer as having measurable gain if
\[
\Delta I_{\mathcal{V}}(\ell)
\ge
\tau\cdot \max_{\ell'}\Delta I_{\mathcal{V}}(\ell'),
\qquad
\tau=0.10.
\]
On both Qwen3-4B and Qwen3-8B, this rule identifies a signal-forming region starting around L16, with rapid growth through the upper-middle layers and near-final values in the upper layers. We therefore use the gain region beginning at L16 as the candidate interval for the bootstrap plateau test.

\paragraph{Probe head and training procedure.}
At every candidate layer $\ell$, we train a layer-specific linear probe
\[
f_\ell(h)=\mathrm{softmax}(W_\ell h+b_\ell),
\qquad
W_\ell\in\mathbb{R}^{2\times D},\quad b_\ell\in\mathbb{R}^{2},
\]
using the same linear-with-bias family as Appendix~\ref{app:vinfo}. Training uses full-batch Adam, cross-entropy loss, weight decay $0$, $100$ epochs, and learning-rate selection from
\[
\{10^{-4},\ 5\times10^{-4},\ 10^{-3}\}
\]
by validation macro-F$_1$. We repeat the full training procedure under seeds $\{42,123,456\}$, changing probe initialization but not the data split. The bootstrap confidence interval is computed from the per-sample predictions of the median seed, so the interval reflects sample-level uncertainty. Across-seed variation is reported separately in Fig.~\ref{fig:vinfo}.

\paragraph{Stratified train/validation/test split.}
The plateau test uses the same 70/15/15 split as Appendix~\ref{app:vinfo}, jointly stratified by label and in-domain dataset. Joint stratification ensures that each label--dataset cell is represented in train, validation, and test with deterministic counts. The same split indices are reused for every layer, so macro-F$_1$ differences across layers are evaluated on identical examples.

\paragraph{Macro-F$_1$ on the held-out split.}
We report macro-F$_1$ on the held-out test portion of the in-domain probe split. Macro-F$_1$ averages the per-class F$_1$ scores for \textsc{faithful} and \textsc{hallucinated}, weighting both labels equally. Predictions are obtained by argmax on calibrated softmax probabilities; no per-layer threshold tuning is used.

\paragraph{Paired bootstrap on the F$_1$ difference.}
For each candidate layer $\ell$, we form the macro-F$_1$ gap to the final layer:
\begin{equation}
\Delta_{\mathrm{final}}(\ell)
=
F_1(\ell)-F_1(L_{\max}).
\label{eq:delta-final}
\end{equation}
We compute a 95\% confidence interval by paired bootstrap with $B=1000$ replicates. Each replicate draws one resample index vector over the held-out examples and applies the same vector to both the layer-$\ell$ and final-layer prediction streams. Macro-F$_1$ is recomputed for both layers on the resampled examples, and the bootstrap replicate records the difference. The confidence interval is the empirical $[2.5,97.5]$ percentile range of these paired differences. Pairing preserves sample-level correlation and directly tests whether two probes differ on the same examples.

\paragraph{The $L^\star$ identification rule.}
We define the saturation layer as the earliest layer whose F$_1$ gap to the final layer is statistically indistinguishable from zero:
\begin{equation}
L^\star
=
\min
\left\{
\ell\in\{1,\dots,L_{\max}-1\}
:
0\in
\widehat{\mathrm{CI}}_{0.95}
\bigl(\Delta_{\mathrm{final}}(\ell)\bigr)
\right\}.
\label{eq:lstar-simple}
\end{equation}
We use the symmetric ``CI contains zero'' criterion because either direction of a statistically reliable gap is informative: an intermediate layer that loses to the final layer would not yet be plateau-equivalent, while an intermediate layer that beats the final layer would also contradict a simple monotone-depth interpretation. The body reports $\Delta_{\mathrm{final}}(L^\star)$ and its 95\% CI as the plateau-membership statistic.

\paragraph{Reproducibility notes.}
Probe training runs in single-precision GPU arithmetic; the bootstrap loop runs on CPU in double precision. Bootstrap resampling uses a fixed seed offset by layer index. The 70/15/15 split is keyed by the dataset seed and is invariant to probe initialization seed and model checkpoint. The plateau results use $B=1000$ replicates and the median of the three probe seeds $\{42,123,456\}$ as the per-sample prediction source.

\subsection{Joint probing and partial information decomposition}
\label{app:pid}

This appendix describes the joint-probing and PID analysis reported in Sec.~\ref{sec:pid} and Table~\ref{tab:pid}.

\paragraph{Goal and scope.}
Sec.~\ref{sec:saturation} shows that, from $L^\star$ onward, the upper layers provide interchangeable single-layer readouts for a linear classifier. That result does not determine whether plateau layers contain the same task information or different information that happens to yield the same F$_1$. This distinction matters for deployment. If two layers carry complementary evidence, a fusion readout should improve over either layer alone. If they carry redundant evidence, fusion should add little, and truncating above the plateau onset should be safe. The PID analysis tests this distinction in calibrated information units; the fair OOD fusion test then asks whether the same conclusion holds for downstream F$_1$ under a deployment-style training protocol.

\paragraph{PID formal setup.}
Partial information decomposition~\citep{williams2010nonnegative} decomposes the joint information of two sources $X_1,X_2$ about a target $Y$ into four atoms:
\begin{equation}
I(X_1,X_2;Y)
=
R+U_1+U_2+S,
\label{eq:pid-decomp}
\end{equation}
where $R$ is redundant information, $U_i$ is information unique to source $X_i$, and $S$ is synergistic information available only from the joint observation. The marginal mutual informations satisfy
\begin{align}
I(X_1;Y) &= R+U_1, \label{eq:pid-marg1}\\
I(X_2;Y) &= R+U_2. \label{eq:pid-marg2}
\end{align}
Different PID definitions specify redundancy differently, and therefore produce different atom values. Our goal is not to adjudicate among PID axiomatizations, but to test whether the joint readout extracts information beyond the better single-layer readout.

\paragraph{Min-MI bound used here.}
We use a min-MI decomposition over calibrated $\mathcal{V}$-information quantities. Let $X_1=H_\ell$ be a candidate layer, $X_2=H_p$ be a plateau layer, and $Y$ be the binary label. We define
\begin{align}
\widehat{R}
&=
\min\!\left(
I_{\mathcal{V}_\ell}(H_\ell;Y),
I_{\mathcal{V}_p}(H_p;Y)
\right),
\label{eq:pid-rhat}\\
\widehat{UI}_{\ell}
&=
\max\!\left(
0,\,
I_{\mathcal{V}_\ell}(H_\ell;Y)
-
I_{\mathcal{V}_p}(H_p;Y)
\right),\\
\widehat{UI}_{p}
&=
\max\!\left(
0,\,
I_{\mathcal{V}_p}(H_p;Y)
-
I_{\mathcal{V}_\ell}(H_\ell;Y)
\right),\\
\widehat{\mathcal{S}}
&=
I_{\mathcal{V}_{\mathrm{joint}}}(H_\ell,H_p;Y)
-
I_{\mathcal{V}_\ell}(H_\ell;Y)
-
I_{\mathcal{V}_p}(H_p;Y)
+
\widehat{R}.
\label{eq:pid-shat}
\end{align}
The key diagnostic is whether the joint probe exceeds the better marginal probe:
\begin{equation}
\widehat{\mathcal{S}}>0
\quad\Longleftrightarrow\quad
I_{\mathcal{V}_{\mathrm{joint}}}(H_\ell,H_p;Y)
>
\max\!\left(
I_{\mathcal{V}_\ell}(H_\ell;Y),
I_{\mathcal{V}_p}(H_p;Y)
\right).
\label{eq:s-iff}
\end{equation}
This equivalence is a statement about the calibrated probes themselves and does not depend on choosing a particular exact PID solver. We also report the Practical Sufficiency Gap,
\[
\mathrm{PSG}(\ell)
=
I_{\mathcal{V}_{\mathrm{joint}}}(H_\ell,H_p;Y)
-
I_{\mathcal{V}_p}(H_p;Y)
=
\widehat{UI}_{\ell}+\widehat{\mathcal{S}},
\]
which measures how much the joint readout improves over the plateau layer alone.

\paragraph{Probe-family compatibility.}
The marginal and joint probe families must be compatible: if the joint family cannot express the marginal probes, then comparisons such as
\[
I_{\mathcal{V}_{\mathrm{joint}}}
\ge
\max(I_{\mathcal{V}_\ell}, I_{\mathcal{V}_p})
\]
would not be guaranteed by model-class inclusion. Our joint probe is an asymmetric SwiGLU--Linear fusion head: a SwiGLU branch processes $H_\ell$, a Linear branch processes $H_p$, the two branches are concatenated, and a small fusion MLP produces logits. The matching marginal families are:
\begin{itemize}
\item $\mathcal{V}_{\ell}$: a SwiGLU $\to$ Dropout $\to$ Linear$_2$ probe trained on $H_\ell$ alone;
\item $\mathcal{V}_{p}$: the Linear-with-bias probe trained on $H_p$ alone;
\item $\mathcal{V}_{\mathrm{joint}}$: the asymmetric fusion head trained on $(H_\ell,H_p)$.
\end{itemize}
The marginal probes embed into the joint family through the standard single-branch construction: one branch can be set to a constant contribution while the other branch carries the active marginal readout. All three quantities entering Eqs.~\ref{eq:pid-rhat}--\ref{eq:pid-shat} use the same split and the same temperature-calibration protocol from Appendix~\ref{app:vinfo}.

\paragraph{Per-layer landscape and synergy-optimal layer.}
We compute the PID quantities independently for every candidate layer
\[
\ell\in\{1,\dots,L_{\max}-1\}
\]
against a fixed plateau layer $H_p=H_{L_{\max}}$ by default. This produces per-layer arrays for $\widehat{R}$, $\widehat{UI}_{\ell}$, $\widehat{UI}_p$, $\widehat{\mathcal{S}}$, and $\mathrm{PSG}$. Table~\ref{tab:pid} reports the atoms at the synergy-optimal layer
\begin{equation}
\ell^\star
=
\operatorname*{arg\,max}_{\ell}
\widehat{\mathcal{S}}(\ell).
\label{eq:ell-star}
\end{equation}
This is a conservative summary against the redundancy claim: if any candidate layer contains complementary information relative to the plateau, the synergy-optimal layer is where that evidence should be strongest. We also save the full PID landscape and the synergy-optimal, MI-optimal, and PSG-optimal layer indices.

\paragraph{Plateau-layer robustness.}
By default, the plateau branch uses $L_{\max}$, but the redundancy claim concerns the plateau as a region rather than one final layer. We therefore repeat the joint-probing pipeline while replacing $H_p$ with plateau layers from $L_{\max}-1$ down to $L^\star$. The PID atoms remain within seed-level variation across this sweep: $\widehat{UI}_{\ell}$ stays at zero and $\widehat{\mathcal{S}}$ remains small. This rules out the possibility that the result is an artifact of choosing the final layer as the plateau representative.

\paragraph{Fair OOD fusion test.}
Information-level redundancy does not by itself rule out a practical F$_1$ gain from a higher-capacity fusion model. We therefore run a separate fair OOD fusion test, reported in the right-most column of Table~\ref{tab:pid}.

\textit{Architecture.}
The deployed fusion model takes $[H_\ell;H_p]\in\mathbb{R}^{2D}$ as input. The candidate branch maps $H_\ell$ through a SwiGLU $\to$ Dropout $\to$ Linear $\to$ LayerNorm stack into a 256-dimensional bottleneck. The plateau branch maps $H_p$ through Linear $\to$ LayerNorm into the same bottleneck dimension. A gate
\[
g\in[0,1]^{256}
\]
is computed from
\[
[z_\ell;\ z_p;\ |z_\ell-z_p|;\ z_\ell\odot z_p]\in\mathbb{R}^{1024},
\]
and the fused representation is
\[
z_{\mathrm{fuse}}
=
z_p+g\odot W_{\mathrm{tr}}z_\ell,
\qquad
W_{\mathrm{tr}}\in\mathbb{R}^{256\times256}.
\]
A final Linear$(256\to2)$ layer emits classification logits. Auxiliary single-branch heads are trained jointly for branch supervision but discarded at inference.

\textit{Relation to the analysis-time joint probe.}
The analysis-time joint probe is designed for interpretability: its marginal families are matched to the PID calculation. The deployed fusion head is designed as a stronger engineering baseline, using residual gating and auxiliary supervision. Both use the same SwiGLU--Linear branch decomposition and the same 256-dimensional bottleneck, but they serve different purposes: the former supports the information decomposition, while the latter tests whether fusion helps in deployment.

\textit{Candidate-layer choice.}
For each cell, the deployed fusion model uses the candidate layer with the largest analysis-time synergy, $\ell^\star$. This gives fusion the most favorable layer according to the PID landscape. In Table~\ref{tab:pid}, this corresponds to L18 and L12 for the 4B and 8B \textsc{cls} models, L26 for the \textsc{base} controls at both scales, and L21 and L14 for the 4B and 8B \textsc{s2s} models.

\textit{Training data and evaluation.}
The fusion model is trained on the same in-domain training corpus used by the single-layer baseline: $22{,}827$ examples from RAGTruth, ClaimVerify, and FactCheck-GPT, with the same 95/5 train/validation split. It uses the same LoRA-r32 backbone adaptation, optimizer family, learning rate, and number of epochs. Evaluation uses the six-corpus eval-half: RAGTruth, ClaimVerify, FactCheck-GPT, AggreFact, FaithBench, and FavaBench. We do not use OOD adaptation, dataset re-weighting, or per-corpus threshold tuning. Thus, any positive fusion gain must come from the additional candidate-layer branch rather than from extra access to the evaluation distribution.

\paragraph{Paired-bootstrap significance for $\Delta F_1$.}
For each paradigm--scale cell in Table~\ref{tab:pid}, we align per-sample predictions from the single-layer baseline and deployed fusion model by dataset and example index. The point estimate is the macro-averaged difference over the six evaluation datasets:
\begin{equation}
\widehat{\Delta F_1}
=
\frac{1}{6}
\sum_{d=1}^{6}
\left[
F_1^{\,d}(\mathrm{fusion})
-
F_1^{\,d}(\mathrm{last})
\right].
\label{eq:delta-fair}
\end{equation}
For the 95\% CI, we use a dataset-stratified paired bootstrap with $B=10{,}000$ replicates and seed $42$. Each dataset is resampled independently, preserving equal dataset weight in the final average. Within a dataset, the same resample indices are applied to the fusion and baseline prediction streams. This preserves sample-level correlation while preventing a small dataset such as FaithBench from dominating the aggregate uncertainty. We report the percentile confidence interval and a two-sided bootstrap p-value,
\[
p
=
2\min\left(
\Pr[\Delta_b\le0],
\Pr[\Delta_b\ge0]
\right).
\]
Cells with $p\ge0.05$ are marked as non-significant.

\paragraph{What the fair OOD test rules out.}
A near-zero or negative $\Delta F_1$ rules out the deployment-relevant version of the complementary-information hypothesis: even when fusion receives the most favorable candidate layer and a stronger gated architecture, it does not produce a generalizable advantage over the last-layer readout under matched training and evaluation conditions. Together with $\widehat{UI}_{\ell^\star}=0$ and small $\widehat{\mathcal{S}}(\ell^\star)$, this supports the interpretation that the plateau is redundant for the label information needed by the detector.

\paragraph{Reproducibility notes.}
All V-information quantities in Eqs.~\ref{eq:pid-rhat}--\ref{eq:pid-shat} use the temperature-calibrated estimates from Appendix~\ref{app:vinfo}. The analysis-time joint probes run on a single GPU in \texttt{float32}; PID atom computation runs on CPU in \texttt{float64}. The fair-fusion bootstrap uses \texttt{seed=42} and $B=10{,}000$ replicates. Saved prediction files reproduce the CIs in Table~\ref{tab:pid} bit-for-bit.

\subsection{Training dynamics of the plateau}
\label{app:dynamics}

This appendix describes the per-checkpoint trajectory analysis reported in Sec.~\ref{sec:dynamic} and Fig.~\ref{fig:iv-trajectory}. The analysis motivates the substrate / InfoWindow / upper-plateau partition used by the early-exit rule in Sec.~\ref{sec:early-exit}.

\paragraph{Goal and scope.}
Endpoint analyses show that the upper layers of a trained \textsc{cls} model contain largely redundant linearly decodable label information. They do not reveal how that redundancy emerges during training. Two mechanisms are possible. In the first, the upper layers create the binary signal and earlier layers later align to it. In the second, a middle band forms the signal first, and the upper layers later become redundant carriers of that signal. These mechanisms can produce similar endpoint probe curves, so we inspect the temporal order in which layers acquire decodable label information. The trajectory analysis supports the second mechanism.

\paragraph{Per-checkpoint $\mathcal{V}$-information matrix.}
At every saved \textsc{cls} checkpoint
\[
t\in\{0,t_1,t_2,\dots,T\},
\]
including $t=0$ for the pretrained backbone, we extract last-token hidden states
\[
H_\ell^{(t)}\in\mathbb{R}^{N\times D}
\]
for every layer $\ell\in\{1,\dots,L\}$ on the same in-domain probe pool used by the static analysis. We then refit the calibrated linear-probe estimate from Appendix~\ref{app:vinfo}:
\begin{equation}
I_{\mathcal{V}}(\ell,t)
=
\max\!\left(
0,\,
H(Y)
-
\widehat{\mathrm{CE}}_{\mathcal{V}}\bigl(H_\ell^{(t)}\bigr)
\right).
\label{eq:iv-of-t}
\end{equation}
This produces the layer-by-time matrix
\[
I_{\mathcal{V}}(\cdot,\cdot)\in\mathbb{R}^{L\times T}
\]
visualized in Fig.~\ref{fig:iv-trajectory}. Rows are layers, columns are checkpoints, and color denotes calibrated $\mathcal{V}$-information in nats.

\paragraph{What is held constant across checkpoints.}
Temporal comparisons are meaningful only if changes in $I_{\mathcal{V}}(\ell,t)$ reflect changes in the model representation. We therefore hold the following fixed across checkpoints:
\begin{itemize}
\item \emph{Sample identity.} The same in-domain examples and the same 70/15/15 stratified split are reused at every checkpoint.
\item \emph{Probe family.} The linear-with-bias probe family from Appendix~\ref{app:vinfo} is used throughout.
\item \emph{Optimization protocol.} Each per-layer probe is trained from scratch with the same Adam, full-batch, learning-rate-grid, 100-epoch, validation-F$_1$ selection procedure. We do not warm-start probes from earlier checkpoints.
\item \emph{Calibration.} A fresh scalar temperature is fit for each layer and checkpoint on the validation split and applied to the test split.
\item \emph{Input processing.} Tokenization, chat template, prompt formatting, and answer-position selection are identical to the static run.
\end{itemize}
Under this design, the only intended source of temporal variation is the model checkpoint.

\paragraph{Checkpoint cadence.}
Checkpoints are saved every $150$ optimizer steps, including the pretrained baseline at $t=0$. Across the three-epoch \textsc{cls} schedule with effective batch size $8$, this yields approximately $55$ columns per scale. This cadence is dense enough to resolve the staggered onset in the InfoWindow and the more synchronous saturation in the upper plateau. We keep all intermediate checkpoints rather than selecting only validation-best checkpoints, so the trajectory is not biased toward a particular operating point.

\paragraph{Per-layer trajectory summaries.}
We summarize each layer's trajectory using two quantities.

\textit{Total information gain.}
For each layer,
\begin{equation}
\Delta I_{\mathcal{V}}(\ell)
=
I_{\mathcal{V}}(\ell,T)
-
I_{\mathcal{V}}(\ell,0).
\label{eq:delta-iv-time}
\end{equation}
This measures how much linearly decodable label information the layer gains over fine-tuning. We mark a layer as having non-trivial gain if
\[
\Delta I_{\mathcal{V}}(\ell)
\ge
\tau\cdot
\max_{\ell'}
\Delta I_{\mathcal{V}}(\ell'),
\qquad
\tau=0.10.
\]

\textit{Saturation time.}
For layers that cross the gain threshold, we define
\begin{equation}
t_\ell^{0.9}
=
\min
\left\{
t:
I_{\mathcal{V}}(\ell,t)
\ge
0.9\cdot I_{\mathcal{V}}(\ell,T)
\right\}.
\label{eq:t90}
\end{equation}
Layers below the gain threshold are not assigned a saturation time because their trajectories are dominated by small fluctuations around the pretrained value. The saturation time is reported in optimizer steps.

Together, $\Delta I_{\mathcal{V}}(\ell)$ measures how much a layer learns, while $t_\ell^{0.9}$ measures when it learns it. The combination distinguishes a layer that forms the signal early from a layer that merely carries it later.

\paragraph{Three regimes from the trajectory matrix.}
The trajectory matrix yields a stable three-regime partition across Qwen3-4B and Qwen3-8B.

\begin{description}
\item[{\textnormal{(R1)} \emph{Lower substrate} ($\ell\in[\mathrm{L}1,\mathrm{L}15]$).}]
These layers remain below the gain threshold. Their $I_{\mathcal{V}}$ trajectories stay close to their pretrained values, and no $t_\ell^{0.9}$ is assigned. We interpret this region as a lower substrate: fine-tuning reaches these layers, but it does not substantially increase their linear label-decodability.

\item[{\textnormal{(R2)} \emph{InfoWindow} ($\ell\in[\mathrm{L}16,\mathrm{L}19]$).}]
These are the first layers to exceed the gain threshold. Their saturation times follow a depth-staggered pattern: lower InfoWindow layers reach $t_\ell^{0.9}$ before upper InfoWindow layers. This staggered onset suggests that the binary signal is progressively constructed across this narrow band rather than appearing simultaneously across the network.

\item[{\textnormal{(R3)} \emph{Synchronous upper plateau} ($\ell\in[\mathrm{L}20,L_{\max}]$).}]
Upper layers also acquire label information, but their saturation times cluster tightly:
\begin{equation}
\max_{\ell\in\mathrm{plateau}} t_\ell^{0.9}
-
\min_{\ell\in\mathrm{plateau}} t_\ell^{0.9}
\le
\tau_{\mathrm{step}},
\qquad
\tau_{\mathrm{step}}=1000.
\label{eq:plateau-sync}
\end{equation}
Moreover, these saturation times occur after the InfoWindow has already crossed its own saturation threshold. Thus, the upper plateau is the last region to align to the task signal, not the first.
\end{description}

\paragraph{What the temporal order rules out.}
If the upper plateau were the source of the binary signal, we would expect its saturation times to precede those of the InfoWindow. Instead, the InfoWindow saturates first and does so in a depth-staggered manner, while the upper plateau saturates later and more synchronously. This temporal pattern supports the interpretation that the InfoWindow forms the task signal and the upper plateau later carries a redundant copy. Combined with the endpoint redundancy evidence from Sec.~\ref{sec:pid}, the trajectory analysis identifies the upper plateau as a late carrier rather than the locus of signal formation.

\paragraph{Robustness of the partition.}
The three-regime split is stable under reasonable perturbations:
\begin{itemize}
\item \emph{Probe seed.} Repeating the trajectory under probe seeds $\{42,123,456\}$ shifts individual $t_\ell^{0.9}$ values by at most one checkpoint for most layers and does not change the regime boundaries.
\item \emph{Gain threshold.} Sweeping $\tau\in[0.05,0.20]$ moves the upper InfoWindow boundary by at most one layer and keeps the lower boundary at L16.
\item \emph{Synchrony threshold.} Tightening $\tau_{\mathrm{step}}$ from $1000$ to $500$ optimizer steps still leaves most plateau layers in the synchrony band; loosening it to $2000$ does not change the ordering that the InfoWindow saturates before the plateau.
\item \emph{Calibration.} Using uncalibrated CE changes the numerical scale of some upper-layer values but preserves the ordering of the regimes. We report calibrated values because they give a cleaner information-scale diagnostic.
\end{itemize}

\paragraph{Connection to the early-exit rule.}
The trajectory partition provides the dynamic basis for Sec.~\ref{sec:early-exit}. The early-exit rule keeps the lower substrate and InfoWindow, where the model constructs the binary signal, and exits at or just above the plateau onset, where additional layers mainly act as redundant carriers. The trajectory analysis supplies what endpoint metrics alone cannot: evidence that removing the synchronous plateau removes late carriers rather than the source of the task signal.

\paragraph{Reproducibility notes.}
The trajectory matrix uses the same dataset seed, probe initialization seeds $\{42,123,456\}$, learning-rate grid, 100-epoch budget, and LBFGS calibration protocol as Appendix~\ref{app:vinfo}. Probe training runs in \texttt{float32} on GPU; calibration runs in \texttt{float64} on CPU. Saturation times are computed from the median seed's trajectory, with multi-seed mean and standard deviation saved for reference. The pretrained column at $t=0$ is obtained by applying the same probe protocol to the unmodified pretrained backbone, so Eq.~\ref{eq:delta-iv-time} matches the cross-model gain definition in Appendix~\ref{app:saturation} for the \textsc{cls} trajectory.

\section{Early-Exit Experimental Details}
\label{app:early_exit}

This appendix gives the full design and implementation details for the early-exit experiments reported in Sec.~\ref{sec:early-exit} and Table~\ref{tab:early-exit}. The goal is to make the early-exit numbers reproducible without relying on codebase-specific assumptions.

\paragraph{Goal and scope.}
For each backbone--exit pair $(m,\ell_{\mathrm{exit}})$, we train a fresh \textsc{cls} adapter whose classification head reads from the layer-$\ell_{\mathrm{exit}}$ residual stream instead of the final layer $L$. All other ingredients are matched to the full-depth \textsc{cls} baseline: data, prompt, tokenizer, optimizer, LoRA configuration, classification-head architecture, training schedule, and evaluation protocol. Thus, the early-exit comparison isolates the effect of moving the readout depth.

The early-exit configurations evaluated in Table~\ref{tab:early-exit} are:
\begin{itemize}
\item \textit{Qwen3-4B} ($L=36$): $\ell_{\mathrm{exit}}\in\{15,20,25\}$, covering a pre-InfoWindow exit, the first plateau-aligned exit, and a deeper InfoWindow-plus-margin exit.
\item \textit{Qwen3-8B} ($L=36$): $\ell_{\mathrm{exit}}\in\{15,20,25\}$, using the same regime-defined exits as Qwen3-4B.
\item \textit{Llama-3.2-3B} ($L=28$): $\ell_{\mathrm{exit}}\in\{17,20\}$, corresponding to aggressive $\mathrm{round}(0.60L)$ and conservative $\lceil0.70L\rceil$ retained-depth rules.
\item \textit{Mistral-7B-v0.3} ($L=32$): $\ell_{\mathrm{exit}}\in\{19,23\}$, using the same aggressive and conservative retained-depth rules.
\end{itemize}
This produces ten early-exit fine-tuning runs in total. Each run yields one adapter and one classifier head, evaluated on the same six-corpus eval-half as the full-depth baseline.

\paragraph{Forward pass and readout.}
Given a tokenized input of length $T$, the backbone produces residual-stream tensors
\[
H_0,H_1,\dots,H_L\in\mathbb{R}^{T\times D},
\]
where $H_0$ is the embedding output and $H_\ell$ for $\ell\ge1$ is the output of transformer block $\ell$. With left padding, the final input position is always the last sequence index. The early-exit representation is therefore
\begin{equation}
\mathbf{h}_{\mathrm{cls}}(x)
=
H_{\ell_{\mathrm{exit}}}(x)[T-1,:]
\in\mathbb{R}^{D}.
\label{eq:ee-pool}
\end{equation}
A bias-free linear head then produces the binary logits:
\begin{equation}
\mathrm{logits}(x)
=
W_{\textsc{cls}}\mathbf{h}_{\mathrm{cls}}(x),
\qquad
W_{\textsc{cls}}\in\mathbb{R}^{2\times D}.
\label{eq:ee-head}
\end{equation}
This head has the same architecture as the full-depth \textsc{cls} head in Appendix~\ref{app_setup}: no bias term, fp32 weights, and HuggingFace-default initialization.

\paragraph{Training-time gradient flow.}
During fine-tuning, we keep the full backbone instantiated for implementation simplicity. The forward pass still computes layers above $\ell_{\mathrm{exit}}$, but the loss depends only on $H_{\ell_{\mathrm{exit}}}$. By the chain rule, LoRA parameters in blocks above $\ell_{\mathrm{exit}}$ receive zero gradient. In practice, this means that adapters are inserted in all blocks for configuration parity, but adapters above the exit layer are not updated. When distributed training is used, \texttt{find\_unused\_parameters=True} prevents errors from these unused parameters. This implementation choice does not change the learned early-exit adapter; a deployment implementation can physically truncate the backbone above $\ell_{\mathrm{exit}}$.

\paragraph{Compute-saving accounting.}
The ``Saved'' column of Table~\ref{tab:early-exit} is the fraction of removed transformer blocks, $(L-\ell_{\mathrm{exit}})/L$.
During fine-tuning and benchmark evaluation we keep the full forward pass for implementation simplicity; the loss depends only on $H_{\ell_{\mathrm{exit}}}$, so the blocks above the exit receive no gradient and the trained adapter is the same as with a truncated stack.
The wall-clock saving of the truncated stack is measured separately in Appendix~\ref{app:latency} and is within a few percent of the layer fraction.

\paragraph{Exit selection uses the probe half only.}
Three quantities determine the Qwen3 exits: the InfoWindow (L16--L19) and plateau onset (L20) from the trajectory analysis of Appendix~\ref{app:dynamics}, and the readout-equivalence layer $L^\star$ (L22/L23) from Appendix~\ref{app:saturation}.
All three are computed on the probe half, which is disjoint from the eval half (Appendix~\ref{app_setup}).
The eval half enters only the final numbers of Table~\ref{tab:early-exit}; no benchmark result influenced the choice of L15, L20, or L25, and the Llama and Mistral candidates were fixed by the two depth ratios before any result on those models was observed.

\paragraph{Backbone configurations.}
Table~\ref{tab:ee-backbones} lists the constants needed to reproduce the early-exit runs. Here $L$ is the number of transformer blocks and $D$ is the residual-stream width. The classifier head has shape $2\times D$.

\begin{table}[h]
\centering
\caption{\textbf{Per-backbone constants for early-exit runs.} $L$ is the number of transformer blocks; $D$ is the hidden width; ``Exits'' lists the $\ell_{\mathrm{exit}}$ values evaluated in Table~\ref{tab:early-exit}.}
\label{tab:ee-backbones}
\setlength{\tabcolsep}{8pt}
\renewcommand{\arraystretch}{1.2}
\begin{tabular}{@{} l c c l @{}}
\toprule
\textbf{Backbone} & $\bm{L}$ & $\bm{D}$ & \textbf{Exits $\ell_{\mathrm{exit}}$} \\
\midrule
Qwen3-4B            & 36 & 2560 & 15, 20, 25 \\
Qwen3-8B            & 36 & 4096 & 15, 20, 25 \\
Llama-3.2-3B        & 28 & 3072 & 17, 20 \\
Mistral-7B-v0.3     & 32 & 4096 & 19, 23 \\
\bottomrule
\end{tabular}
\end{table}

\paragraph{LoRA configuration.}
Early-exit runs use the same LoRA configuration as the full-depth \textsc{cls} baseline in Appendix~\ref{app_setup}: rank $r=32$, scaling $\alpha=64$, dropout $0.05$, no bias adaptation, and LoRA modules attached to all seven projection matrices in every transformer block:
\[
\texttt{q\_proj},\ 
\texttt{k\_proj},\ 
\texttt{v\_proj},\ 
\texttt{o\_proj},\ 
\texttt{gate\_proj},\ 
\texttt{up\_proj},\ 
\texttt{down\_proj}.
\]
The base backbone weights remain frozen. The classification head $W_{\textsc{cls}}$ is the only non-LoRA trainable component and is registered through PEFT's \texttt{modules\_to\_save}, so its weights are saved with the adapter.

\paragraph{Score-head precision.}
The classification head runs in fp32 with autocast disabled, while the backbone forward pass runs in bfloat16. This choice follows the full-depth \textsc{cls} setup and improves early optimization stability. The randomly initialized score head starts with small weights but reads from residual streams that may contain large-coordinate outliers, especially in Mistral-style backbones. Keeping the readout in fp32 reduces quantization error in the logits without changing the backbone precision.

\paragraph{Optimization hyperparameters.}
Table~\ref{tab:ee-hparams} gives the optimization hyperparameters. Qwen3 and Llama-3.2-3B use the default recipe. Mistral-7B-v0.3 uses the backbone-specific override described below.

\begin{table}[h]
\centering
\caption{\textbf{Optimization hyperparameters for early-exit runs.} ``Default'' covers Qwen3-4B, Qwen3-8B, and Llama-3.2-3B. The Mistral column changes only the listed optimization entries. The effective batch size is per-device batch size $\times$ gradient-accumulation steps $\times$ world size; all runs use one GPU, so the effective batch is $2\times4\times1=8$.}
\label{tab:ee-hparams}
\setlength{\tabcolsep}{8pt}
\renewcommand{\arraystretch}{1.2}
\resizebox{.98\linewidth}{!}{%
\begin{tabular}{@{} l c c @{}}
\toprule
\textbf{Hyperparameter} & \textbf{Default (Qwen3, Llama)} & \textbf{Mistral-7B-v0.3} \\
\midrule
Optimizer                     & AdamW & AdamW \\
Learning rate                 & $1\times10^{-4}$ & $\bm{6\times10^{-5}}$ \\
Weight decay                  & $0.01$ & $\bm{0.1}$ \\
Warmup ratio (linear)         & $0.03$ & $\bm{0.05}$ \\
LR scheduler                  & linear decay & linear decay \\
Number of epochs              & $3$ & $3$ \\
Per-device train batch        & $2$ & $2$ \\
Gradient accumulation steps   & $4$ & $4$ \\
Effective batch (1 GPU)       & $8$ & $8$ \\
Per-device eval batch         & $4$ & $4$ \\
Backbone precision            & bfloat16 & bfloat16 \\
Score-head precision          & fp32, autocast disabled & fp32, autocast disabled \\
Max sequence length (train)   & $2048$ & $2048$ \\
Truncation policy             & left truncation of overlong inputs & left truncation of overlong inputs \\
Padding side                  & left & left \\
Group sequences by length     & yes & yes \\
Random seed                   & $42$ (reseeds $43$, $44$) & $42$ (reseeds $43$, $44$) \\
Save-total-limit              & unlimited & $3$ \\
\bottomrule
\end{tabular}}
\end{table}

\paragraph{Backbone-specific override for Mistral.}
Mistral-7B-v0.3 required a more conservative optimization recipe. With the default learning rate $10^{-4}$, weight decay $0.01$, and $3\%$ warmup, the randomly initialized fp32 score head produced extremely large initial losses and unstable optimization. This behavior is consistent with the known presence of large-magnitude activation outliers in some decoder-only backbones. We therefore use a lower learning rate, longer warmup, and stronger weight decay for all Mistral runs:
\[
\mathrm{lr}=6\times10^{-5},
\qquad
\mathrm{weight\ decay}=0.1,
\qquad
\mathrm{warmup}=5\%.
\]
The override is applied uniformly to Mistral full-depth \textsc{cls}, full-depth \textsc{s2s}, and both Mistral early-exit runs. Therefore, Mistral early-exit results remain directly comparable to the Mistral full-depth baseline within Table~\ref{tab:early-exit}. No other backbone uses a special recipe.

\paragraph{Data, prompt, and tokenization.}
The early-exit runs use the same training corpus, split, prompt, and tokenization as the full-depth \textsc{cls} baseline.
\begin{itemize}
\item \emph{Training data.} We use $22{,}827$ examples from RAGTruth\textsubscript{train}, ClaimVerify\textsubscript{train}, and FactCheck-GPT\textsubscript{train}. The files are concatenated, shuffled with \texttt{seed=42}, and split into approximately $21{,}686$ training rows and $1{,}141$ in-training validation rows under the same $95/5$ split used by the baseline.
\item \emph{Prompt template.} Each example is rendered as a single user-turn chat message:
\begin{verbatim}
Can the hypothesis be inferred from the premise?
Answer using Yes or No only.

Premise: {premise}
Hypothesis: {hypothesis}
Answer:
\end{verbatim}
The rendered string is passed through the native chat template of each backbone.
\item \emph{Tokenization.} We use each backbone's native tokenizer with left padding and a maximum training length of $2{,}048$ tokens. Overlong inputs are truncated from the left, so the trailing \texttt{Answer:} cue and the final-token readout position in Eq.~\ref{eq:ee-pool} are preserved. The data collator pads each batch to the longest sequence in the batch.
\end{itemize}
The early-exit setting changes only the residual-stream layer used for classification; it does not change which rows or tokens the model sees.

\paragraph{Schedule and checkpointing.}
With approximately $21{,}686$ training rows and an effective batch size of $8$, three epochs correspond to approximately $8{,}133$ optimizer steps. We save checkpoints and run intermediate evaluation every $150$ optimizer steps, giving about $55$ snapshots per run. The HuggingFace trainer is configured with validation macro-F$_1$ for best-checkpoint tracking. However, to match the full-depth \textsc{cls} reporting protocol in Sec.~\ref{sec:setup}, Table~\ref{tab:early-exit} reports the final-step checkpoint rather than the validation-best checkpoint. Intermediate checkpoints are retained for trajectory analyses and do not enter the early-exit benchmark table. For Mistral, we keep only the three most recent checkpoints on disk to control storage; this does not alter the final-step adapter.

\paragraph{Evaluation.}
Evaluation follows the same readout as training. For each input, we left-pad, run the backbone with hidden states enabled, extract $H_{\ell_{\mathrm{exit}}}[T-1,:]$, apply the trained head in Eq.~\ref{eq:ee-head}, and take the argmax of the two logits as the predicted label. We report macro-F$_1$ on the eval-half of each of the six test corpora: RAGTruth, ClaimVerify, FactCheck-GPT, AggreFact, FaithBench, and FavaBench. Eval-half sizes are listed in Appendix~\ref{app_setup}. We do not apply per-corpus threshold tuning, calibration, dataset reweighting, or post-processing.

\paragraph{Hardware.}
Each early-exit run uses a single NVIDIA H200 GPU with 141\,GB HBM3e. The backbone runs in bfloat16. SDPA attention is used by default; FlashAttention-2 is used when available for the corresponding backbone and otherwise the run falls back to SDPA without changing any other configuration. Latency in Table~\ref{tab:latency} is measured separately on the physically truncated stack (Appendix~\ref{app:latency}).

\paragraph{Reproducibility notes.}
The original runs use \texttt{seed=42} for data shuffling, the train/validation split, PyTorch and NumPy initialization, and dataloader worker seeding; the reseeded runs behind Tables~\ref{tab:main} and~\ref{tab:early-exit} use seeds 43 and 44 for the same quantities and are otherwise identical. Layer indexing follows Appendix~\ref{app:vinfo}: $\ell=1,\dots,L$ denotes post-block residual streams, while $\ell=0$ is the embedding output and is never used as an exit. Tables~\ref{tab:ee-backbones} and~\ref{tab:ee-hparams} list the backbone dimensions and optimizer settings needed to reproduce the runs. A reimplementation with the same backbone IDs, data files, split seed, prompt template, and hyperparameters should reproduce Table~\ref{tab:early-exit} up to standard GPU nondeterminism in attention and matrix-multiplication kernels.

% ======================================================================
%  New appendix section collecting every result added after review.
%  Insert with \input{snippets/app_rebuttal.tex} before \newpage / checklist.
%  Tables are \input from snippets/, figures from figs/ (paths relative to main.tex).
% ======================================================================

\section{Robustness and Generality Analyses}
\label{app:robustness}

This appendix tests the results of Sections~\ref{sec:setup} and~\ref{sec:early-exit} against the choices they could depend on: the fine-tuning seed, the \textsc{s2s} readout, the composition of the test corpora, the backbone, and the task.
Every run keeps the data, optimizer, LoRA configuration, and schedule of Appendix~\ref{app_setup} and changes one variable at a time.

\subsection{Seed variability}
\label{app:seeds}

Tables~\ref{tab:per-seed-main} and~\ref{tab:per-seed-exit} list the individual runs behind the means in Tables~\ref{tab:main} and~\ref{tab:early-exit}.
The in-distribution gap is positive in 11 of 12 runs and its standard deviation never exceeds 1.1 points.
The out-of-distribution gap varies by an order of magnitude more (standard deviation 10.1 on Llama-3.2-3B and 12.1 on Mistral-7B-v0.3), and the variation comes from single runs.
The seed-44 \textsc{cls} model of Mistral-7B-v0.3 collapses toward the positive class on FavaBench and averages 35.6 on the three OOD corpora, against 63.5 and 53.1 for the other two seeds; the seed-44 \textsc{cls} model of Llama-3.2-3B does the same (42.4).
Both collapses are threshold effects on an imbalanced corpus (Appendix~\ref{app:fava}) and leave the in-distribution scores of the same runs intact (77.2 and 80.2).
The same two runs produce the large standard deviations of the Mistral and Llama exits in Table~\ref{tab:early-exit}, because each delta is taken against the same-seed full-depth model.

\begin{table}[htbp]
\centering
\caption{\textbf{Individual full-depth runs.} Macro-F$_1$ (\%) and paired differences in percentage points. Within each group, $\Delta$ is \textsc{cls}$-$\textsc{s2s}.}
\label{tab:per-seed-main}
\setlength{\tabcolsep}{4pt}
\renewcommand{\arraystretch}{1.1}
\resizebox{\linewidth}{!}{%
\begin{tabular}{@{}llrrrrrrrrr@{}}
\toprule
Backbone & Seed & S2S ID & CLS ID & $\Delta$ ID & S2S OOD & CLS OOD & $\Delta$ OOD & S2S All & CLS All & $\Delta$ All \\
\midrule
Qwen3-4B & 42 & 78.09 & 80.24 & 2.14 & 52.43 & 71.27 & 18.84 & 65.26 & 75.75 & 10.49 \\
Qwen3-4B & 43 & 78.85 & 82.36 & 3.51 & 47.34 & 70.75 & 23.41 & 63.10 & 76.56 & 13.46 \\
Qwen3-4B & 44 & 76.52 & 80.40 & 3.87 & 50.54 & 70.38 & 19.84 & 63.53 & 75.39 & 11.86 \\
\midrule
Qwen3-8B & 42 & 77.97 & 80.33 & 2.36 & 66.68 & 74.07 & 7.39 & 72.33 & 77.20 & 4.87 \\
Qwen3-8B & 43 & 79.02 & 81.47 & 2.46 & 60.22 & 68.56 & 8.34 & 69.62 & 75.02 & 5.40 \\
Qwen3-8B & 44 & 78.12 & 82.10 & 3.98 & 66.54 & 69.34 & 2.80 & 72.33 & 75.72 & 3.39 \\
\midrule
Llama-3.2-3B & 42 & 74.57 & 78.19 & 3.61 & 45.83 & 60.15 & 14.32 & 60.20 & 69.17 & 8.97 \\
Llama-3.2-3B & 43 & 76.47 & 79.04 & 2.58 & 48.74 & 54.36 & 5.62 & 62.60 & 66.70 & 4.10 \\
Llama-3.2-3B & 44 & 75.29 & 80.22 & 4.92 & 52.61 & 42.41 & -10.20 & 63.95 & 61.31 & -2.64 \\
\midrule
Mistral-7B-v0.3 & 42 & 77.89 & 79.91 & 2.02 & 57.42 & 63.45 & 6.03 & 67.65 & 71.68 & 4.02 \\
Mistral-7B-v0.3 & 43 & 78.14 & 79.82 & 1.68 & 53.06 & 53.06 & 0.00 & 65.60 & 66.44 & 0.84 \\
Mistral-7B-v0.3 & 44 & 77.71 & 77.17 & -0.54 & 57.72 & 35.59 & -22.13 & 67.72 & 56.38 & -11.34 \\
\bottomrule
\end{tabular}}
\end{table}

\begin{table}[htbp]
\centering
\caption{\textbf{Individual early-exit runs.} Macro-F$_1$ (\%) and differences from the same-seed full-depth \textsc{cls} model. Exit depths are held fixed across seeds.}
\label{tab:per-seed-exit}
\setlength{\tabcolsep}{4pt}
\renewcommand{\arraystretch}{1.1}
% \resizebox{\linewidth}{!}{%
\begin{tabular}{@{}lllrrrrrr@{}}
\toprule
Backbone & Exit & Seed & ID & OOD & All & $\Delta$ ID & $\Delta$ OOD & $\Delta$ All \\
\midrule
Qwen3-4B & L15 & 42 & 79.52 & 52.19 & 65.85 & -0.72 & -19.08 & -9.90 \\
Qwen3-4B & L20 & 42 & 81.14 & 67.05 & 74.09 & 0.91 & -4.22 & -1.66 \\
Qwen3-4B & L20 & 43 & 80.17 & 50.31 & 65.24 & -2.19 & -20.44 & -11.32 \\
Qwen3-4B & L20 & 44 & 80.72 & 68.76 & 74.74 & 0.33 & -1.63 & -0.65 \\
Qwen3-4B & L25 & 42 & 81.40 & 72.11 & 76.76 & 1.17 & 0.84 & 1.00 \\
Qwen3-4B & L25 & 43 & 81.01 & 67.42 & 74.22 & -1.35 & -3.33 & -2.34 \\
Qwen3-4B & L25 & 44 & 80.90 & 69.37 & 75.14 & 0.51 & -1.01 & -0.25 \\
\midrule
Qwen3-8B & L15 & 42 & 78.57 & 47.12 & 62.84 & -1.76 & -26.96 & -14.36 \\
Qwen3-8B & L20 & 42 & 79.99 & 69.39 & 74.69 & -0.33 & -4.68 & -2.51 \\
Qwen3-8B & L20 & 43 & 80.53 & 68.75 & 74.64 & -0.94 & 0.19 & -0.37 \\
Qwen3-8B & L20 & 44 & 80.66 & 71.16 & 75.91 & -1.44 & 1.82 & 0.19 \\
Qwen3-8B & L25 & 42 & 80.68 & 73.00 & 76.84 & 0.35 & -1.07 & -0.36 \\
Qwen3-8B & L25 & 43 & 80.76 & 73.75 & 77.25 & -0.71 & 5.19 & 2.24 \\
Qwen3-8B & L25 & 44 & 81.31 & 71.38 & 76.34 & -0.79 & 2.04 & 0.62 \\
\midrule
Llama-3.2-3B & L17 & 42 & 80.20 & 61.71 & 70.96 & 2.01 & 1.57 & 1.79 \\
Llama-3.2-3B & L17 & 43 & 80.21 & 52.00 & 66.11 & 1.17 & -2.36 & -0.60 \\
Llama-3.2-3B & L17 & 44 & 79.69 & 50.89 & 65.29 & -0.53 & 8.48 & 3.98 \\
Llama-3.2-3B & L20 & 42 & 79.94 & 45.63 & 62.79 & 1.75 & -14.51 & -6.38 \\
Llama-3.2-3B & L20 & 43 & 79.08 & 50.93 & 65.01 & 0.04 & -3.43 & -1.70 \\
Llama-3.2-3B & L20 & 44 & 80.06 & 48.50 & 64.28 & -0.16 & 6.09 & 2.96 \\
\midrule
Mistral-7B-v0.3 & L19 & 42 & 77.88 & 64.38 & 71.13 & -2.03 & 0.94 & -0.55 \\
Mistral-7B-v0.3 & L19 & 43 & 80.18 & 57.28 & 68.73 & 0.37 & 4.22 & 2.29 \\
Mistral-7B-v0.3 & L19 & 44 & 75.80 & 49.05 & 62.42 & -1.37 & 13.45 & 6.04 \\
Mistral-7B-v0.3 & L23 & 42 & 79.18 & 66.32 & 72.75 & -0.73 & 2.87 & 1.07 \\
Mistral-7B-v0.3 & L23 & 43 & 79.06 & 57.52 & 68.29 & -0.75 & 4.46 & 1.85 \\
Mistral-7B-v0.3 & L23 & 44 & 79.50 & 61.07 & 70.29 & 2.33 & 25.48 & 13.90 \\
\bottomrule
\end{tabular}
\end{table}

\subsection{Stronger \textsc{s2s} readouts}
\label{app:strong-s2s}

We take the fine-tuned \textsc{s2s} adapters of Table~\ref{tab:main} (seed 42) and replace single-token greedy decoding with three stronger readouts, keeping the adapters fixed.
\textbf{(a) Likelihood-ratio scoring.}
One forward pass records the last-position logit of every verdict string that tokenizes to a single token (\texttt{Yes}/\texttt{No}, \texttt{yes}/\texttt{no}, \texttt{YES}/\texttt{NO}, \texttt{True}/\texttt{False}, \texttt{true}/\texttt{false}, \texttt{Correct}/\texttt{Incorrect}, \texttt{Right}/\texttt{Wrong}, with and without a leading space).
The score is the log of the total probability mass on the positive strings minus the log of the mass on the negative strings (a log-sum-exp over each set), and an example is predicted positive when the score exceeds zero.
The string set is fixed in advance and not selected on the eval half.
\textbf{(b) Calibration.}
Temperature scaling is monotone and changes no argmax decision, so we report it only for completeness.
Label-shift calibration fits one decision threshold per corpus on the probe half and applies it to the eval half; we apply the same per-corpus fitting to the \textsc{cls} logit margin so that both paradigms receive the same information.
\textbf{(c) LM-head adaptation.}
We add the unembedding matrix to the trainable parameters (LoRA on all projections plus a full update of \texttt{lm\_head}) and retrain with the recipe of Appendix~\ref{app_setup}, which frees the \texttt{Yes}/\texttt{No} direction that single-token \textsc{s2s} otherwise inherits from pretraining.
For Qwen3-4B, whose unembedding is tied to the input embedding, the update unties the two matrices.
This variant is read out by generation and verdict parsing, as in Table~\ref{tab:main}.

Table~\ref{tab:strong-s2s} reports six-corpus averages.
Likelihood scoring moves \textsc{s2s} by less than one point.
Label-shift calibration raises \textsc{s2s} by 5 to 14 points but raises \textsc{cls} by 2 to 7 points as well.
The adapted LM head changes \textsc{s2s} by at most 3 points (a second Qwen3-8B seed scores 67.9, further below \textsc{cls}).
\textsc{cls} leads under every readout, by 2.2 to 5.3 points when both paradigms are calibrated and by 4.0 to 10.4 points when neither is.

\begin{table}[h]
\centering
\caption{\textbf{Alternative \textsc{s2s} readouts and matched threshold calibration.}
Six-corpus mean macro-F$_1$ (\%), using fine-tuning seed 42. The verbalizer ensemble is fixed, without selection on the eval half. Per-corpus thresholds are fitted separately for each paradigm on labeled probe-half examples from that corpus. LM-head adaptation uses generation and verdict parsing.}
\label{tab:strong-s2s}
\setlength{\tabcolsep}{6pt}
\renewcommand{\arraystretch}{1.15}
\resizebox{0.9\linewidth}{!}{%
\begin{tabular}{@{} l cccc @{}}
\toprule
\textbf{Variant} & \textbf{Qwen3-4B} & \textbf{Qwen3-8B} & \textbf{Llama-3.2-3B} & \textbf{Mistral-7B} \\
\midrule
\textsc{s2s}, generation (paper protocol) & 65.2 & 72.3 & 60.2 & 67.7 \\
(a) \textsc{s2s}, Fixed verbalizer-ensemble LLR & 65.3 & 71.6 & 59.5 & 67.7 \\
(c) \textsc{s2s}, Generation + trainable LM head & 68.0 & 71.6 & 61.7 & 69.5 \\
(b) \textsc{s2s}, LLR + per-corpus label-shift calibration & 74.0 & 76.9 & 73.5 & 75.5 \\
\midrule
\textbf{\textsc{cls}, fixed $0.5$ threshold (no tuning)} & \textbf{75.7} & \textbf{77.2} & \textbf{69.2} & \textbf{71.7} \\
\textbf{\textsc{cls}, same label-shift calibration as (b)} & \textbf{79.3} & \textbf{79.5} & \textbf{76.6} & \textbf{77.7} \\
\midrule
$\Delta$ uncalibrated: \textsc{cls}@0.5 $-$ (a) & \textcolor{gainpos}{+10.4} & \textcolor{gainpos}{+5.6} & \textcolor{gainpos}{+9.7} & \textcolor{gainpos}{+4.0} \\
$\Delta$ calibrated: \textsc{cls}-LS $-$ (b) & \textcolor{gainpos}{+5.3} & \textcolor{gainpos}{+2.5} & \textcolor{gainpos}{+3.1} & \textcolor{gainpos}{+2.2} \\
\bottomrule
\end{tabular}}
\end{table}

\subsection{FavaBench diagnosis and significance tests}
\label{app:fava}

FavaBench is the most imbalanced test corpus, with a positive rate of 0.20, and contributes the largest single-corpus gaps in Table~\ref{tab:main}.
Three explanations are possible for the low \textsc{s2s} scores there: verdicts that fail to parse, collapse to one class, and a miscalibrated threshold.
Parsing is not the cause.
Every fine-tuned model emits a valid verdict on 100\% of FavaBench examples (Table~\ref{tab:fava}).
The predicted positive rate of \textsc{s2s} ranges from 31\% to 89\% against a true rate of 20\%, so the loss is a threshold effect on an imbalanced corpus.
The same effect reaches \textsc{cls}: at the fixed 0.5 threshold, one \textsc{cls} seed each of Llama-3.2-3B and Mistral-7B-v0.3 collapses toward the positive class on FavaBench (Appendix~\ref{app:seeds}).
We therefore base the headline comparison on the in-distribution gap, on the calibrated comparison of Appendix~\ref{app:strong-s2s}, and on the bootstrap below, and we report the fixed-threshold OOD average as what it is, a quantity that either paradigm can lose on a single seed.

\textbf{Paired bootstrap.}
For each backbone we resample eval-half examples with replacement, independently within each corpus and paired between the \textsc{cls} and \textsc{s2s} prediction streams ($10{,}000$ replicates), recompute each corpus's macro-F$_1$ and the average gap, and report the 95\% percentile interval and the two-sided percentile $p$-value, once over all six corpora and once excluding FavaBench (Table~\ref{tab:fava}).
Excluding FavaBench shrinks the seed-42 gap from 4.0 to 10.5 points to 2.2 to 3.6 points, and every interval stays above zero.

\begin{table}[h]
\centering
\caption{\textbf{FavaBench predictions and sensitivity to excluding the corpus.}
All results use fine-tuning seed 42. Left: invalid-verdict rate and predicted class proportions on FavaBench, whose positive-label rate is 20\%. Right: paired \textsc{cls}$-$\textsc{s2s} differences in mean macro-F$_1$ points, with 95\% percentile intervals from 10,000 within-corpus bootstrap resamples. These intervals describe example-sampling uncertainty for fixed models.}
\label{tab:fava}
\setlength{\tabcolsep}{5pt}
\renewcommand{\arraystretch}{1.15}
\resizebox{\linewidth}{!}{%
\begin{tabular}{@{} l cc c !{\color{rulegray}\vrule} cc cc @{}}
\toprule
& \multicolumn{3}{c}{\textbf{\textsc{s2s} on FavaBench}} & \multicolumn{2}{c}{\textbf{Gap, all six}} & \multicolumn{2}{c}{\textbf{Gap, excl.\ FavaBench}} \\
\cmidrule(lr){2-4} \cmidrule(lr){5-6} \cmidrule(lr){7-8}
\textbf{Backbone} & parse fail & pred.\ neg / pos & verdict & $\Delta$ [95\% CI] & $p$ & $\Delta$ [95\% CI] & $p$ \\
\midrule
\textit{Qwen3-4B} & 0.0\% & 33\% / 67\% & over-positive & +10.5 [+9.1, +11.8] & $<$0.001 & \textbf{+3.3} [+1.8, +4.9] & $<$0.001 \\
\textit{Qwen3-8B} & 0.0\% & 69\% / 31\% & miscalibrated & +4.9 [+3.5, +6.2] & $<$0.001 & \textbf{+3.6} [+2.0, +5.1] & $<$0.001 \\
\textit{Llama-3.2-3B} & 0.0\% & 11\% / 89\% & collapse to positive & +9.0 [+7.6, +10.3] & $<$0.001 & \textbf{+2.9} [+1.3, +4.4] & $<$0.001 \\
\textit{Mistral-7B-v0.3} & 0.0\% & 49\% / 51\% & miscalibrated & +4.0 [+2.7, +5.3] & $<$0.001 & \textbf{+2.2} [+0.7, +3.7] & 0.004 \\
\bottomrule
\end{tabular}}
\end{table}

\subsection{Per-layer probing of the transfer backbones and the readout-swap control}
\label{app:cross-model}

\begin{figure}[t]
  \centering
  \includegraphics[width=\linewidth]{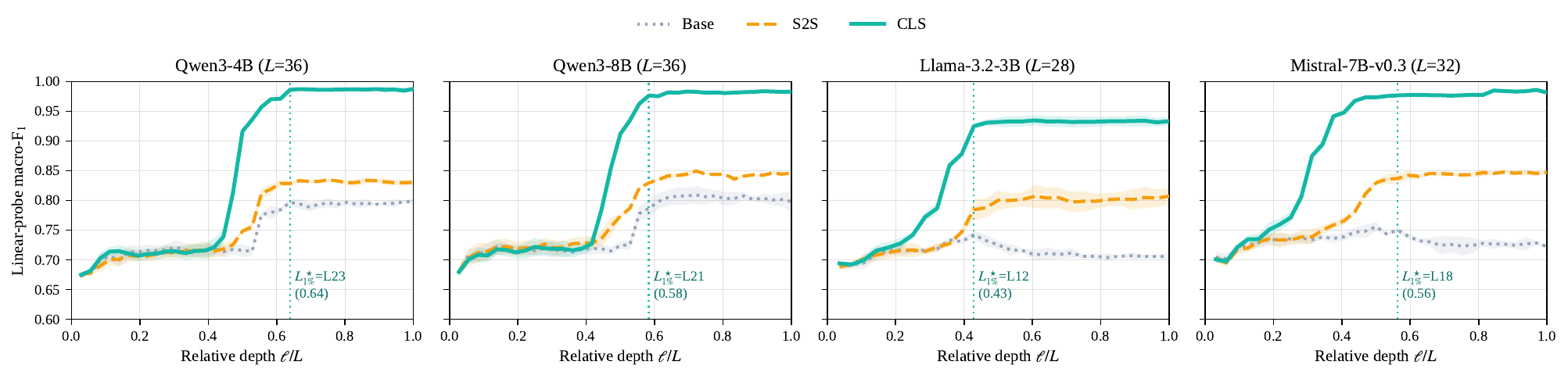}
  \caption{\textbf{Per-layer linear-probe F$_1$ on all four backbones.}
  \textsc{base}, \textsc{s2s}, and \textsc{cls} probes on the in-domain probe half (mean of three probe seeds, bands show one standard deviation across probe seeds), plotted against relative depth $\ell/L$.
  Every backbone shows the flat pretrained substrate, a sharp rise, and a plateau, with \textsc{cls} above \textsc{s2s} above \textsc{base}; the dotted line marks the \textsc{cls} saturation layer $L^\star_{1\%}$.}
  \label{fig:probe-four}
\end{figure}

Section~\ref{sec:early-exit} fixes the Llama-3.2-3B and Mistral-7B-v0.3 exits from a depth fraction before any analysis of those models.
Here we run the per-layer linear probing of Appendix~\ref{app:vinfo} on both, unchanged: the same probe family, the same 70/15/15 split of the probe half, and the same three probe seeds, for the \textsc{base}, \textsc{s2s}, and \textsc{cls} models of Table~\ref{tab:main}.
Figure~\ref{fig:probe-four} plots probe F$_1$ against relative depth for all four backbones and Table~\ref{tab:lstar-four} lists the best-layer scores.

The curves share one shape.
Probe F$_1$ stays near the pretrained level through the lower 25\% to 45\% of depth, rises within a few layers, and plateaus, with \textsc{cls} above \textsc{s2s} above \textsc{base} on the plateau for every backbone.
As an inference-only summary of where the rise ends we report $L^\star_{1\%}$, the earliest layer within 0.01 F$_1$ of the best layer.
It is layer 12 of 28 for Llama-3.2-3B (0.43 of depth), 18 of 32 for Mistral-7B-v0.3 (0.56), and 23 and 21 of 36 for Qwen3-4B and 8B (0.64 and 0.58).
$L^\star_{1\%}$ is a descriptive proxy for the bootstrap saturation layer $L^\star$ of Section~\ref{sec:saturation}, which we computed only for Qwen3; on Qwen3 the two differ by one to two layers.
All four transfer exits of Table~\ref{tab:early-exit} lie above $L^\star_{1\%}$, which is why both the 0.60 and the 0.70 ratio land on the plateau for these two models.

\textbf{Readout-swap control.}
Higher probe decodability under \textsc{cls} could in principle reflect only that \textsc{cls} trains a linear head on the hidden state directly.
\textsc{s2s}, however, is also a linear objective on the same final state: its verdict is the sign of $(e_{\texttt{Yes}}-e_{\texttt{No}})^{\top}h$, a boundary whose direction is fixed by the unembedding.
To separate the representation from the trained head, we discard both heads, freeze each backbone, and fit the same fresh linear probe, layer by layer, on the features of the \textsc{s2s} and \textsc{cls} models.
The last column of Table~\ref{tab:lstar-four} reports the difference at the best layer: the \textsc{s2s} representation trails by 0.153, 0.134, 0.128, and 0.138 F$_1$ on Qwen3-4B, Qwen3-8B, Llama-3.2-3B, and Mistral-7B-v0.3.
A fresh head does not recover the gap, so the gap sits in the representation.

\begin{table}[h]
\centering
\caption{\textbf{Linear-probe accuracy and saturation depth across backbones.}
Scores are the maximum over layers of mean probe macro-F$_1$ across three probe seeds, using the original fine-tuning runs. $L^\star_{1\%}$ is the earliest layer within 0.01 of that maximum. The final column subtracts the best-layer scores of separately fitted probes with the same architecture and training protocol. These are descriptive probe comparisons, not early-exit selection results.}
\label{tab:lstar-four}
\setlength{\tabcolsep}{6pt}
\renewcommand{\arraystretch}{1.15}
\resizebox{0.95\linewidth}{!}{%
\begin{tabular}{@{} l c ccc c c !{\color{rulegray}\vrule} c @{}}
\toprule
\textbf{Backbone} & $\bm{L}$ & \textbf{\textsc{base}} & \textbf{\textsc{s2s}} & \textbf{\textsc{cls}} & $\bm{L^\star_{1\%}}$ & $\bm{L^\star_{1\%}/L}$ & \textbf{\textsc{cls}$-$\textsc{s2s} (same probe family)} \\
\midrule
\textit{Qwen3-4B} & 36 & 0.798 & 0.835 & \textbf{0.988} & L23 & 0.64 & \textcolor{gainpos}{+0.153} \\
\textit{Qwen3-8B} & 36 & 0.809 & 0.850 & \textbf{0.984} & L21 & 0.58 & \textcolor{gainpos}{+0.134} \\
\textit{Llama-3.2-3B} & 28 & 0.742 & 0.807 & \textbf{0.935} & L12 & 0.43 & \textcolor{gainpos}{+0.128} \\
\textit{Mistral-7B-v0.3} & 32 & 0.756 & 0.848 & \textbf{0.986} & L18 & 0.56 & \textcolor{gainpos}{+0.138} \\
\bottomrule
\end{tabular}}
\end{table}

\subsection{Replication of the layer structure on QNLI and SST-2}
\label{app:second-task}

\begin{figure}[t]
  \centering
  \includegraphics[width=\linewidth]{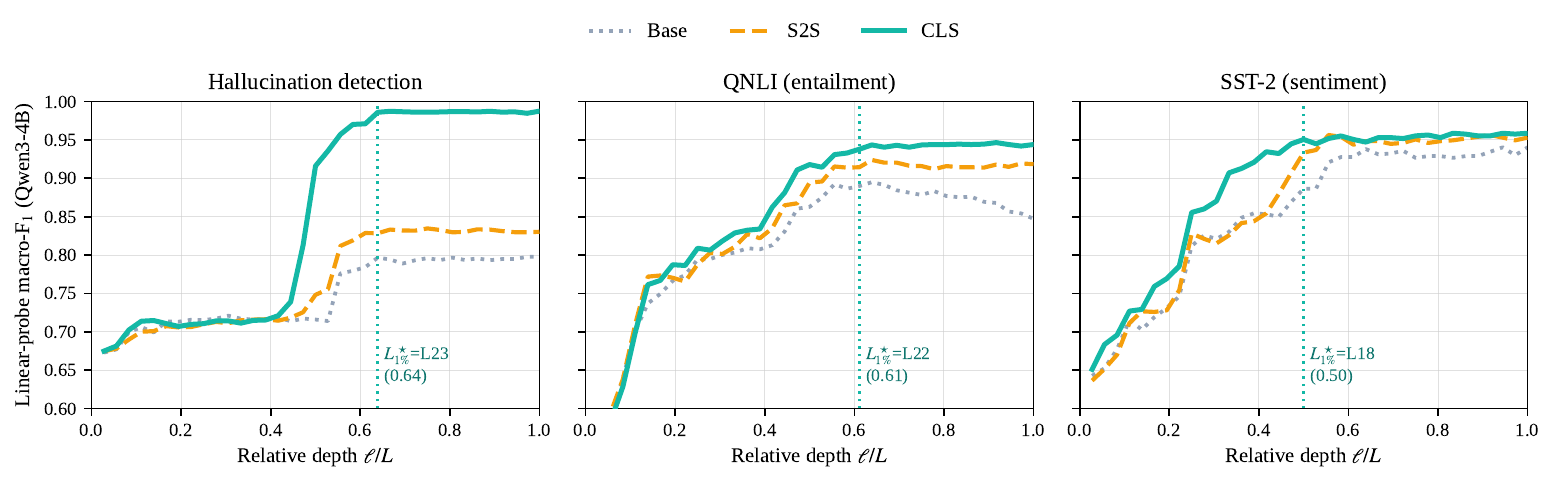}
  \caption{\textbf{The layer structure replicates on QNLI and SST-2.}
  Per-layer probe F$_1$ for Qwen3-4B fine-tuned on hallucination detection, QNLI, and SST-2 with the same \textsc{cls}/\textsc{s2s} recipe. The shape and the location of $L^\star_{1\%}$ are shared; the \textsc{cls}$-$\textsc{s2s} gap shrinks on the two easier tasks because the pretrained backbone already decodes them well.}
  \label{fig:probe-tasks}
\end{figure}

To test whether the structure is specific to hallucination detection, we fine-tune Qwen3-4B with the unchanged \textsc{cls} and \textsc{s2s} recipes on two further binary tasks.
QNLI uses its sentence as context and its question as claim; SST-2 uses the review as context and the fixed claim ``The review expresses positive sentiment.''
Each task uses $22{,}827$ balanced training examples to match the hallucination training set, and we hold out $5{,}463$ QNLI and $872$ SST-2 test examples.
We then fit a logistic-regression probe per layer on $3{,}000$ training examples and score macro-F$_1$ on up to $2{,}000$ held-out examples for the \textsc{base}, \textsc{s2s}, and \textsc{cls} models (one fine-tuning seed, one probe seed; Table~\ref{tab:second-task}, Figure~\ref{fig:probe-tasks}).

Both tasks reproduce the substrate, the sharp rise, and the plateau, and $L^\star_{1\%}$ falls at 0.61 (QNLI) and 0.50 (SST-2) of depth, inside the 0.43 to 0.64 band of the four hallucination detectors.
The \textsc{cls}$-$\textsc{s2s} gap at the best layer, however, is 0.023 on QNLI and 0.002 on SST-2, against 0.13 to 0.16 on hallucination detection.
The pretrained backbone already reaches 0.895 and 0.940 probe F$_1$ on the two tasks, so fine-tuning has little to reorganize; on hallucination detection it starts from 0.74 to 0.81.
The shape of the curve is task-general, and the size of the \textsc{cls} advantage tracks how much the task requires the representation to change.

\begin{table}[h]
\centering
\caption{\textbf{Best-layer probe scores and saturation depths on three tasks.}
QNLI and SST-2 use a single fine-tuning seed and the logistic-regression protocol described in Appendix~\ref{app:second-task}; hallucination detection uses the three-probe-seed protocol in Appendix~\ref{app:cross-model}. Differences are computed before rounding. $L^\star_{1\%}$ summarizes early saturation and does not test statistical equivalence.}
\label{tab:second-task}
\setlength{\tabcolsep}{6pt}
\renewcommand{\arraystretch}{1.15}
\resizebox{0.85\linewidth}{!}{%
\begin{tabular}{@{} l ccc c c c @{}}
\toprule
\textbf{Task (Qwen3-4B)} & \textbf{\textsc{base}} & \textbf{\textsc{s2s}} & \textbf{\textsc{cls}} & \textbf{\textsc{cls}$-$\textsc{s2s}} & $\bm{L^\star_{1\%}}$(\textsc{cls}) & $\bm{L^\star_{1\%}/L}$ \\
\midrule
Hallucination detection & 0.798 & 0.835 & \textbf{0.988} & +0.153 & L23 & 0.64 \\
QNLI (entailment) & 0.895 & 0.924 & \textbf{0.947} & +0.023 & L22 & 0.61 \\
SST-2 (sentiment) & 0.940 & 0.956 & \textbf{0.959} & +0.002 & L18 & 0.50 \\
\bottomrule
\end{tabular}}
\end{table}

\subsection{Measured latency of truncated backbones}
\label{app:latency}

Table~\ref{tab:latency} reports per-example forward latency of the \textsc{cls} models with the transformer blocks above the exit physically removed before the adapter is loaded, at batch size 1 and sequence length $2{,}048$ in bfloat16, averaged after five warm-up iterations, on an NVIDIA H200 and an RTX A6000.
Removing 31\% of the blocks (L25) cuts latency by $1.42$ to $1.44\times$ and removing 44\% (L20) by $1.76$ to $1.79\times$.
Batched throughput at batch size 16 rises accordingly, from 4.8 to 6.9 to 8.6 examples per second for Qwen3-4B on the A6000 and from 14.6 to 20.9 to 26.1 for Qwen3-8B on the H200.
The speedups fall a few percent short of the layer fraction because the cost of the embedding and the readout does not shrink with depth.

\begin{table}[h]
\centering
\caption{\textbf{Forward-pass latency of physically truncated backbones.}
Mean time per forward pass at batch size 1 and sequence length 2,048, using bf16 and SDPA. Models use pretrained backbone weights and a two-class head, without task adapters. Parentheses give speedup over full depth. These measurements isolate the effect of reducing depth; they are not end-to-end detector timings.}
\label{tab:latency}
\setlength{\tabcolsep}{7pt}
\renewcommand{\arraystretch}{1.15}
\resizebox{\linewidth}{!}{%
\begin{tabular}{@{} l l c c c @{}}
\toprule
\textbf{GPU} & \textbf{Model} & \textbf{Full depth} & \textbf{L25 ($-31\%$ blocks)} & \textbf{L20 ($-44\%$ blocks)} \\
\midrule
H200 (datacenter) & Qwen3-4B & 47.4 ms & 33.0 ms (\textbf{1.44$\times$}) & 26.5 ms (\textbf{1.79$\times$}) \\
H200 (datacenter) & Qwen3-8B & 71.8 ms & 50.1 ms (\textbf{1.43$\times$}) & 40.1 ms (\textbf{1.79$\times$}) \\
RTX A6000 (constrained) & Qwen3-4B & 220.6 ms & 155.8 ms (\textbf{1.42$\times$}) & 125.0 ms (\textbf{1.76$\times$}) \\
\bottomrule
\end{tabular}}
\end{table}

\newpage
\section*{NeurIPS Paper Checklist}

\begin{enumerate}

\item {\bf Claims}
    \item[] Answer: \answerYes{}
    \item[] Justification: The abstract and introduction clearly state the paper's main empirical and analytical claims, including the performance gap between \textsc{cls} and \textsc{s2s}, the objective-dependent changes in representation geometry, the emergence of an InfoWindow during training, and the resulting early-exit rule. Each claim is evaluated through controlled experiments and layer-wise analyses in Sections~\ref{sec:setup}--\ref{sec:early-exit}.

\item {\bf Limitations}
    \item[] Answer: \answerYes{}
    \item[] Justification: The paper discusses limitations and future directions in Section~\ref{sec:conclusion} and Appendix~\ref{app:limitations}. These discussions clarify the scope of the present study and suggest that the observed representation-level patterns may offer useful insights for future research on hallucination detection and efficient model verification.

\item {\bf Theory assumptions and proofs}
    \item[] Answer: \answerNA{}
    \item[] Justification: The paper does not present formal theoretical results or theorems. The mathematical expressions are used to define training objectives, diagnostic quantities, and empirical estimators for the representation and early-exit analyses.

\item {\bf Open access to data and code}
    \item[] Answer: \answerNo{}
    \item[] Justification: The study builds on publicly available datasets and open-weight model backbones. The appendices provide detailed documentation of the experimental pipeline, including data processing, model configurations, training hyperparameters, probing protocols, bootstrap procedures, and evaluation settings, to support reproducibility and future research.

\item {\bf Experimental setting/details}
    \item[] Answer: \answerYes{}
    \item[] Justification: The paper provides the main details needed to reproduce the experimental results, including datasets, splits, prompts, model backbones, LoRA configurations, optimization hyperparameters, probing protocols, bootstrap procedures, and early-exit settings. These details are reported in Appendices~\ref{app_setup}--\ref{app:early_exit}.

\item {\bf Experiment statistical significance}
    \item[] Answer: \answerYes{}
    \item[] Justification: The paper reports uncertainty estimates for the key layer-wise conclusions. In particular, paired bootstrap confidence intervals are used for plateau interchangeability and fusion analyses in Sections~\ref{sec:saturation}--\ref{sec:pid} and Appendices~\ref{app:saturation}--\ref{app:pid}.

\item {\bf Experiments compute resources}
    \item[] Answer: \answerYes{}
    \item[] Justification: The paper reports the main compute resources used for training and evaluation, including single NVIDIA H200 GPUs, bfloat16 execution, training steps, batch sizes, and approximate wall-clock time for early-exit runs in Appendix~\ref{app:early_exit}. Additional training, probing, and representation-analysis details are provided in Appendices~\ref{app_setup}--\ref{app:dynamics}.
    
\item {\bf Code of ethics}
    \item[] Answer: \answerYes{}
    \item[] Justification: The research uses public datasets and open-weight model backbones to study hallucination detection and efficient fine-tuning. The work is intended to improve understanding of model verification behavior and does not introduce new data collection, human-subject experiments, or high-risk deployment artifacts.

\item {\bf Broader impacts}
    \item[] Answer: \answerYes{}
    \item[] Justification: Appendix~\ref{app:limitations} discusses the broader implications of efficient hallucination detection for more trustworthy RAG and agentic systems. The paper also emphasizes that automated verifiers should complement, rather than replace, task-specific evaluation, careful deployment testing, and human oversight where appropriate.
    
\item {\bf Safeguards}
    \item[] Answer: \answerNA{}
    \item[] Justification: The paper does not introduce or release a new high-risk generative model, scraped dataset, or dual-use foundation model. It analyzes hallucination-detection fine-tuning using existing public datasets and open-weight model backbones.

\item {\bf Licenses for existing assets}
    \item[] Answer: \answerYes{}
    \item[] Justification: The paper cites the original sources for all datasets, model backbones, and software packages used in the study, and describes their roles in the experimental setup. Appendix~\ref{app:license} further states that these existing assets are publicly available research resources and that their use follows the access conditions and usage terms specified by the original providers or hosting platforms.

\item {\bf New assets}
    \item[] Answer: \answerNA{}
    \item[] Justification: The study does not introduce a new dataset, benchmark, model checkpoint, or software package requiring separate asset documentation. All experiments are conducted with existing research datasets and model backbones.

\item {\bf Crowdsourcing and research with human subjects}
    \item[] Answer: \answerNA{}
    \item[] Justification: The paper does not involve new crowdsourcing experiments or new research with human subjects. All experiments use existing datasets.

\item {\bf Institutional review board (IRB) approvals or equivalent for research with human subjects}
    \item[] Answer: \answerNA{}
    \item[] Justification: The paper does not involve new human-subject experiments, crowdsourcing, participant recruitment, or participant data collection.

\item {\bf Declaration of LLM usage}
    \item[] Answer: \answerYes{}
    \item[] Justification: LLMs are central to the research methodology. The paper fine-tunes decoder-only LLM backbones under \textsc{s2s} and \textsc{cls} objectives, analyzes their layer-wise representations, and evaluates early-exit variants. The models, training setup, and evaluation protocol are described in Sections~\ref{sec:prelim}--\ref{sec:early-exit} and Appendices~\ref{app_setup}--\ref{app:early_exit}.

\end{enumerate}

\end{document}